\RequirePackage{fix-cm}
\documentclass[smallextended]{svjour3}       
\smartqed  
\usepackage{graphicx}
\usepackage{amsfonts}
\usepackage{amsmath}
\usepackage{amssymb}
\usepackage{caption}
\usepackage{subcaption}
\usepackage{booktabs}
\usepackage{adjustbox}
\usepackage{colortbl}
\usepackage{multirow,tabularx}
\usepackage{tikz}
\usepackage{soul}
\usepackage{bm}
\usepackage{tcolorbox}
\usepackage{longtable}
\usepackage{xcolor}
\usepackage{hyperref}
\usepackage[linesnumbered,ruled,vlined]{algorithm2e}

\SetKwInput{KwInput}{Input}
\SetKwInput{KwOutput}{Output}
\SetKwInput{KwRequired}{Required}
\definecolor{Gray}{gray}{0.9}
\newtheorem{defn}[theorem]{Definition} %

\newcommand{\nd}{\vspace{1mm}\noindent}%

\newcolumntype{g}{>{\columncolor{Gray}}c}

\newcommand{\Gabriel}[1]{\textcolor{red}{{\it [Gabriel says: #1]}}}

\begin{document}

\title{Detection and Evaluation of bias-inducing Features in Machine learning 
}


\author{Moses Openja         
       \and Gabriel Laberge \and  Foutse Khomh  
}


\institute{M. Openja \at
              Polytechnique Montreal \\
              Tel.: +(1)438-505 5297\\
              \email{openja.moses@polymtl.ca}           
\and
           G. Laberge \at
               Polytechnique Montreal \\
               \email{gabriel.laberg@polymtl.ca}
           \and
           F. Khomh \at
               Polytechnique Montreal \\
               \email{foutse.khomh@polymtl.ca}
}

\date{Accepted: October 18, 2023}

\maketitle

\begin{abstract}

The cause-to-effect analysis can help us decompose all the likely causes of a problem, such as an undesirable business situation or unintended harm to the individual(s). This implies that we can identify how the problems 
are inherited, rank the causes to help prioritize fixes, simplify a complex problem and visualize them. 
In the context of machine learning (ML), one can use cause-to-effect analysis to understand the reason for the biased behavior of the system. For example, we can 
examine the root causes of biases by checking each feature for a potential cause of bias in the model. To approach this, one can apply small changes to a given feature or a pair of features in the data, following some guidelines and observing how it impacts the decision made by the model (i.e., model prediction). Therefore, 
we can use cause-to-effect analysis to identify 
the potential bias-inducing features, even when these features are originally are unknown. 
This is important since most current methods require a pre-identification of sensitive features for bias assessment and can actually miss other relevant bias-inducing features, which is why systematic identification of such features is necessary. Moreover, it often occurs that to achieve an equitable outcome, one has to take into account sensitive features in the model decision. Therefore, it should be up to the domain experts to decide based on their knowledge of the context of a decision whether bias induced by specific features is acceptable or not. In this study, we propose an approach for systematically identifying all bias-inducing features of a model to help support the decision-making of domain experts. Our technique is based on the idea of swapping the values of the features and computing the divergences in the distribution of the model prediction using different distance functions. We evaluated our technique using four well-known datasets to showcase how our contribution can help  
spearhead the standard procedure when developing, testing, maintaining, and deploying fair/equitable machine learning systems. 



\end{abstract}

\section{Introduction}

The use of machine learning (ML) systems is permeating every aspect of our life, such as healthcare (e.g., the presence of heart disease~\cite{detrano1989international,aha1988instance}), autonomous systems, education, banking (e.g., loan approval, or marketing campaigns~\cite{moro2014data}), recruitment, and court justice to  assess the likelihood 
that a defendant recommits a crime. These systems are trained on data that are usually biased towards some features, leading to the biased behavior of the resulting model. For example, the COMPAS (Correctional Offender Management Profiling for Alternative Sanctions) algorithm for scoring defendants was found to be biased in the sense of having different False Positive Rates between white and black sub-populations. The Gender Shades project~\cite{buolamwini2018gender}, commercial facial recognition systems, is found to be biased towards the dark skinned women, or gender bias in the Google neural machine translation models~\cite{kuczmarski2018reducing}, among others. The above examples highlight the importance of quantifying whether machine learning systems exhibit biased behavior that impacts some individuals. 

In the recent years, the researchers have been proposing methods such as~\cite{alelyani2021detection,chakraborty2021bias,perera2022search,kamiran2012:decision,hardt2016equality,pleiss2017:fairness,corbett2017algorithmic} to test if a model is making a fairer decision (and mitigating unfair decision). For instance, enforcing that the true positive or false positive rates, or both~\cite{hardt2016equality} are similar for different populations. Likewise, the error rate (i.e., misclassification error) parity~\cite{hardt2016equality} requires that the error rates are the same across all groups or ensure conditional statistical equality~\cite{corbett2017algorithmic}.  These methods can work as long as the sensitive feature is known; however, they may not work for the case where the sensitive features are unknown. Moreover, even when these sensitive features for bias assessment are known upfront, we can still miss other relevant bias-inducing features, which is why it's necessary to systematically identify such features. Additionally, as pointed out in~\cite{corbett2018measure}, one often has to take into account sensitive features in the model decision to achieve an equitable outcome. For example, consider the case of statistical discrimination~\cite{RePEc:pri:indrel:30a,phelps1972statistical} in economy where auto insurers (i.e., representing the domain expert) want to account for gender differences in accident rates and maximize profit by deciding to charge a premium to male drivers. Similarly, in the context of the dominant legal doctrine of discrimination which focuses on the motivations of a decision maker, such as equal protection law established by the U.S. Constitution’s Fourteenth Amendment forbids the action of discriminatory purposes by the government representatives. For instance, in promoting diversity in college admissions, some race-conscious affirmative action programs are legally permissible to further the interest of the government~\cite{2016fisher}. This can be clearly interpreted as anti-classification in the current legal standards where protected characteristics are explicitly necessary for risk assessment algorithms to achieve an equitable outcome. 
Finally, the calibration, by conditioning the model decision to be independent of the sensitive features given the context of a decision. 
Therefore, it should be up to the domain experts to decide whether bias induced by specific features is acceptable or not based on their knowledge of the context of a decision and some ethical implications. To support the context-specific decisions about what constitutes an acceptable bias or not, and hence a fair/equitable outcome, it is better to know all bias-inducing features. For example, age, race, gender, religion, etc., are considered sensitive features by law and correspond to some ethical implications~\cite{yapo2018ethical,barbosa2019rehumanized}.


The cause-to-effect analysis can help us decompose all the potential causes of a problem. This implies that one can investigate all the possible causes until the root cause. 
The advantage of this is to propose a fix based on the cause ranking, decompose or simply a complex problem, visualize the different general causes, and where to put more focus. In the context of machine learning, cause-to-effect analysis has been used~\cite{freedman2005specifying,holland1986statistics,holland2003causation,pearl2000models,blank2004measuring} to understand the reasons for the biased behavior of machine learning systems. 

In this study, we propose an approach for systematically identifying all bias-inducing features of a model to help support the decision-making of domain experts. Specifically, we propose a novel single feature swapping and double feature swapping functions to help estimate the direct and indirect impact of each feature on the model prediction. The single feature swapping function modifies the values of the single feature keeping other features unchanged, and the function's output is used to estimate the direct impact of the feature on the model prediction. The double features swapping function will alter the values of pairs consisting of the feature and all the mediating variables (following the temporal priority ordering~\cite{sep-causation-probabilistic}) used to estimate the total natural impact of the feature on the model prediction. 
The impact of swapping the feature's values on the model predictions is studied by computing the statistical difference in the distribution of the model prediction before and after swapping data, using four different distance functions. Finally, we validate our technique by performing multiple empirical experiments using four well-known datasets to demonstrate the usefulness of our contributions, based on the following two main research objectives:

\begin{itemize}
    \item \textbf{Identify features that potentially introduce bias to the model:}
    The first objective was to demonstrate how our proposed swapping functions can be used to identify the features 
    that directly and indirectly introduce 
    bias to the model.
     \item \textbf{The important features to the model:}
     Given these identified bias-inducing features, we wanted to assert whether or not they are important to the model. 
     By feature importance, we determine the relative importance of each feature in the dataset on the model performance;--- by assessing the impact of each feature on the model prediction without necessarily comparing the impact across the sub-group (coarse-grained), as for the case of bias assessment (fine-grained)~\cite{shin2019role}. We also refer the reader to~\cite{bhattacharya2022applied}, for further reading about the difference. 
     To this end, we contrast our proposed techniques with the SHAP Values (an acronym from Shapley Additive exPlanations), a model explainable method to explain the individual prediction based on game theoretically optimal Shapley values. Notably, we want to demonstrate that treating the concepts of feature importance and bias inducing features as separate is essential in making informed decisions about what features can be 
     most relevant or least relevant, when building a fairer ML model. Providing insights into what features are most relevant or least relevant to the model prediction and are least bias-inducing as interpreted by the domain experts will help the domain expert choose the features that improve both the predictive performance and fairer model.  
\end{itemize}

Through the above two research objectives, we demonstrate empirically that the features potentially bias-inducing to the model and are less important to the model can be removed to improve the fairness of the machine learning model. We believe that is the first step to making an informed decision by the domain experts through the systematic identification of all bias-inducing features. Moreover, This will help visualize the cause of bias in the model, systematic features selection, prioritizing the fixes,  and contributing to the standard procedure when developing, testing, deploying, and maintaining fairer machine learning systems. 

\paragraph{}\textbf{Paper organization.} Section \ref{sec:contribution} provides details about the contribution of our study. In Section~\ref{sec:notation} we introduced the main notations used in this paper followed by the concept of a counterfactual approach to causal inference on which our method is based. Moreover, we introduce the problem definition and the proposed solutions addressing the problems. Section~\ref{sec:experiment} provides the empirical study to validate our proposed techniques following two main research questions.  
Section \ref{sec:results} presents the results of our experiments on identifying the bias-inducing features and the important features of the model. Section~\ref{sec:evaluation} evaluate our framework--- demonstrating that, treating bias-inducing and feature importance as separate is essential when building a fairer ML model. 
 Section \ref{sec:discussion} contains the discussion on how our framework can be used to diagnose and fix an unfair model in a given context. as well as the possible threat to the validity of this work. Section~\ref{sec:related-works} details the related works. Section~\ref{sec:threat} discuss the possible threat to validity, and finally Section \ref{sec:conclusion} concludes the paper.


\section{Contribution of this Study}\label{sec:contribution}

In summary, the following are the main contributions of our study:
\begin{itemize}
    \item First, we investigate ML model bias following the counterfactual approach to causal inference. Some of the concepts of the counterfactual approach to causal inference are discussed in this paper. 
    
    \item Next, a bias detection technique is proposed based on a novel single feature and double features swapping function to detect bias features in the dataset and ML model.
    
    \item Thirdly, an evaluation method is proposed based on divergence measurement to evaluate the impact of biased features on the ML model.

    \item We validate our 
    proposed technique by performing multiple experiments using four different well-known datasets.
    
    \item We showed with the help of the state-of-the-art model explainability tool that the potential bias-inducing features that are less important to the ML model can be removed from the dataset to improve the fairness. We have shown empirically that treating the two entities (bias-inducing and feature importance) as separate is essential in making an informed decision. 
    \item Our study is the first step in helping domain experts make an informed decision by following a systematic identification of bias-inducing features. The domain experts can use our approach to visualize the cause of bias in the ML model, systematic features selection, prioritizing the fixes, and therefore helping contribute to the standard procedure when developing, maintaining, and deploying fairer ML systems.
    
    
\end{itemize}


\section{Notation and Definition}\label{sec:notation}

 We employ a dataset $D_{\text{test}}$ consisting of independent and identically distributed 
 (i.i.d) samples of random variable $\left\{(X_i, Y_i)\right\}_{i=1}^n$ from a joint distribution 
 $\rho(X,Y)$ with domain $\mathcal{X}\times\mathcal{Y}$, where $\mathcal{X}\subseteq \mathbb{R}^m$, i.e. 
 our input space is a vector with $m$ features. We let $X_{ij}$ be the $j$th feature of the $i$th instance. 
 For example, $X_i$ can be the record of $i$th loan applicant and $X_{ij}$ can either be the age, gender, 
 race, or credit history of this applicant.
We also let $\hat{Y}$ be a model prediction i.e. $f(X) \to \hat{Y}$, where $f$ can either be a 
classification or regression model. We can store all measurements of the input in the matrix 
$\bm{X}\in \mathbb{R}^{n\times m}$ by using the convention that $X_{ij}$ is the $i$th row and $j$th 
column of $\bm{X}$.
We will sometimes represent the $i$th row of $\bm{X}$ as $[X_{i1},X_{i2},...,X_{ij},...,X_{im}]$. We will also think of each feature $j$ as being binarized meaning $X_{ij}$ can only take
one of two values ($X_{ij}\in (C_{1j}, C_{2j})\,\,\forall i$), where one category is considered more protected than the other.
Furthermore, we assume that there exists a subset of features $\mathcal{S}\subseteq \{1, 2, \ldots, m\}$ that are considered sensitive meaning that the $j$th column of $\bm{X}$ with $j\in \mathcal{S}$ potentially introduce bias in the model. For the target variable, we will be using $\bm{Y}$ to denote vector containing the true label (i.e., class label), i.e., $\bm{Y} = [Y_1, Y_2, . . . , Y_i, . . . , Y_n]^T$. 


\subsection{Problem Definition}
We want to identify the potentially biased features, evaluate the bias and use the detected features to evaluate the level of bias in a machine learning model or a predictor following some evaluation metrics. Before formally defining our problem statement, we will briefly review the concept  `counterfactual approach to causal inference', on which our method is based.

\subsubsection{Counterfactual Approach to Causal Inference}\label{subsection:Counterfactual}
This section will briefly review the approach of using counterfactual to causal inference and probabilistic causation, on which our method is based. For the detailed discussion on this topic we refer the readers to the literature~\cite{freedman2005specifying,holland2003causation,pearl2000models,blank2004measuring,barocas2017fairness,sep-causation-probabilistic,suppes1970theory,simon1977causal,Theories-of-causal-ordering,johnson2020causal}.

Researchers~\cite{freedman2005specifying,holland2003causation,pearl2000models,blank2004measuring} have used the counterfactual approach to causal inference  when trying to identify the presence or absence of bias between an observing feature (e.g.,  a male gender) and model outcome (e.g., loan approval). To this end, the researchers try to understand the outcome of the given predictor when the alternative value of the observing feature is used (e.g., female, for the gender feature). In other words, they answer the counterfactual question: What would have happened to the outcome of this system if the applicant was a female instead of a male? Answering this question is fundamental to be able to conclude that there is a causal relationship between some specific feature and discrimination, which, in turn, is necessary to conclude that discriminatory behaviors or processes contributed to an observed differential outcome. While this question might not be directly answered, observing the causal relationship between the feature under investigation and discrimination can be ascertained. When understanding causal relationships, one can alter the value or turn on/off the value of the feature under investigation leading to multiple outcomes observed for a single profile, hence answering the above counterfactual question with certainty. Indeed, altering the value of the given feature and monitoring the system outcome is a potential interpretation of causality. However, it's nearly impossible to measure causality in the real world systematically; instead, one can draw causal inferences.

The counterfactual approach to causal inference has seen a number of research studies~\cite{freedman2005specifying,holland1986statistics,holland2003causation,pearl2000models,blank2004measuring} formalizing the assumptions and the deductive process needed to draw cause-and-effect inferences from statistical data.

In the following, we will use the Directed Acyclic Graph (DAG) to describe the cause-and-effect concept given the observing features and the statistical models. 
In DAG, each node represents a different feature, and the graph needs to include unobserved variables that influence observable features. Directed edges between nodes illustrate cause-and-effect relationships between variables. By definition, paths following the directed edges in an acyclic graph cannot lead from a node back to itself (i.e., there is no loop), as it is assumed that a variable cannot be a cause at the same time effect of another variable. 
In the formal statistical theory of DAGs presented by Pearl, Judea~\cite{pearl2000models}, the absence of an edge in the graph corresponds to conditional independence of the variables corresponding to the nodes, given all the other variables represented in the graph.

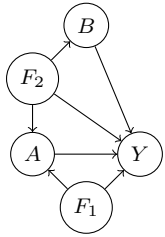
\begin{figure}[!h]
\centering
\begin{tikzpicture}[main/.style = {draw, circle}] 

\node[main] (1) {$A$}; 

\node[main] (3) [above of=1]
{$F_2$};

\node[main] (4) [above right of=3]{$B$}; 

\node[main] (2)  [below right of=1]{ $F_1$};
\node[main] (5) [above right of=2]
{$Y$};

\draw[->] (1) -- (5);
\draw[->] (2) -- (1);
\draw[->] (3) -- (1);
\draw[->] (3) -- (4);
\draw[->] (2) -- (5);
\draw[->] (3) -- (5);
\draw[->] (4) -- (5);
\end{tikzpicture} \caption{The Directed Acyclic Graph (DAG) illustrating the concept of cause-to-effect between the basic four variables case.}\label{fig:DAG-4-variables}
\end{figure}

We can decompose the total cause-to-effect in the DAG by analyzing the path-specific components, which is also referred to as the mediation analysis~\cite{mackinnon2007mediation}. We demonstrate the above concept using the four variables shown by the DAG in Figure~\ref{fig:DAG-4-variables}, although the idea of counterfactual and casual inference 
extends to more complex structures.


Consider the case where we are interested in understanding the impact of the two features $F_1$ and $F_2$. In Figure~\ref{fig:DAG-4-variables}, we can reach $Y$ from $F_1$ by following two different paths.  
A direct 
path (i.e., no intermediate nodes) and indirectly reaching $Y$ through variable $A$, also called the `mediator'. In this case, the conditional expectation $E[Y | F_1 = f_1]$ is the addition of both paths. We can estimate the total effect of the action $F_1 := f_1$ on $Y$ by eliminating the confounding path following the do-operator $E[Y | \text{do}(F_1 := f_1)]$. However, the total effect still lumps the direct and the indirect impact following the two pathways. 

Given that the direct effect does not require any counterfactuals, we can easily estimate the direct impact  of $F_1$ feature above by keeping the mediator variable fixed at state $A:= a$ and studying the divergence of the distribution in the outcome when $F_1:= f_1$ compared to when an alternative treatment $F_1:=\lnot f_1$, 
using the do-operator. If the conditional distributions $\rho(Y|\text{do}(F_1=f_1, A=a))$  and $\rho(Y|\text{do}(F_1=\lnot f_1, A=a))$ are known, then the direct impact will be:

\begin{equation}\label{eqn:CDI}
     \Im_{\mathbb{E}} \big[\,\,\rho(Y|\text{do}(F_1=f_1, A=a))\,\,,\,\, \rho(Y|\text{do}(F_1=\lnot f_1, A=a))\,\,\big] 
\end{equation}

$\Im_{\mathbb{E}}$ in Equation (\ref{eqn:CDI}) denote the expectation of some distance measure quantifying the conditional distributions of the outcome  when do-operations is performed on the values of feature $F_1$, keeping the mediator variable fixed at $A:=a$. Similarly we can compute the direct impact of $F_2$ on Y following similar process, but in this case, we keep both two mediator variables (i.e., $A$ and $B$) of $F_2$ at their respective fixed states, as follows:

\begin{equation}\label{eqn:CDI2}
     \Im_{\mathbb{E}} \big[\,\rho(Y|\text{do}(F_2=f_2, A=a,B=b))\,,\, \rho(Y|\text{do}(F_2=\lnot f_2, A=a, B=b))\,\big] 
\end{equation}

Throughout this paper, we will refer to the direct impact determined using the Equation (\ref{eqn:CDI}) or Equation (\ref{eqn:CDI2}) as ``Controlled Direct Impact'', because they require setting the other variables in some state.

\begin{defn}
    \label{defn:CDI} (Controlled Direct Impact~\cite{berzuini2012causality,rubin1974estimating}) Is the measure of contrast between counterfactual outcomes of the exposure values with alternative values of the exposure (e.g., $F_1:=f_1$ and $F_1:=\lnot f_1$, where $f_1 \neq \lnot f_1$), when the mediator (s) were kept to their respective fixed value. In this study, we will estimate the Controlled Direct Impact using Equation (\ref{eqn:CDI}) or Equation (\ref{eqn:CDI2}). 
\end{defn}

However, the major drawback of directly computing the impact of the features following the above formulation is that it generally does not produce valid counterfactual. Considering that $A$ and $B$ are relevant variables influencing the outcome, computing counterfactual 
with respect to $F_1$ or $F_2$ would require adjusting even the downstream variables $A$ and $B$. We can estimate the impact of the Features $F_1$ and $F_2$ on the outcome $Y$, by adjusting variables $A$ and $B$, as follows:

\begin{equation}\label{eqn:NDI}
\begin{aligned}
    \Im_{\mathbb{E}} \left(\rho(Y|\text{do}(F_1=f_1, A=_{(a,\lnot f_1)})), \rho(Y|\text{do}(F_1=\lnot f_1, A=_{(a,\lnot f_1)}))\right) 
\end{aligned}
\end{equation}
and:
\begin{equation}\label{eqn:NII}
\begin{aligned}
    \Im_{\mathbb{E}} \left(\rho(Y|\text{do}(F_1=\lnot f_1, A=_{(a,f_1)})), \rho(Y|\text{do}(F_1=\lnot f_1, A=_{(a,\lnot f_1)}))\right) 
\end{aligned}
\end{equation}
where the $\rho(Y|\text{do}(F_1=f_1, A=_{(a,\lnot f_1)}))$ in Equation (\ref{eqn:NDI}) denote the probability distribution that the outcome $Y$ would obtain had $F_1$ been set to $f_1$ and had $A$ been set to the value $A$ would've assumed had $F_1$ been set to the alternative value: $F_1 := \lnot f_1$. On the other hand, for Equation (\ref{eqn:NII}), the property $\rho(Y|\text{do}(F_1=\lnot f_1, A=_{(a,f_1)}))$ is the probability distribution of 
the outcome $Y$ when the $F_1$ is kept at the alternative $F_1:=\lnot f_1$ while changing the value of the mediator variable $A$ to the value it would have attained had $F_1:=f_1$ was used. 
The above formulations of impact analysis is referred to as `Natural Impact', where by Equation (\ref{eqn:NII}) and Equation (\ref{eqn:NDI}) can be used to estimate the natural direct and indirect impact of feature $F_1$ on the outcome $Y$.

\begin{defn}\label{defn:NI}
(Natural Impact~\cite{berzuini2012causality,rubin1974estimating}) is the causal effect of the exposure defining the hypothetical contrast between the outcomes that would be observed simultaneously in the same individual in the presence and the absence of the  
exposure. When we add up the two Natural Impacts, i.e., the natural direct effects in Equation~(\ref{eqn:NDI}) and natural indirect effects in Equation~(\ref{eqn:NII}), the resulting value is the total effect~\cite{robins1992identifiability,Pearl2001Direct-and-indirect}.
\end{defn}

So far, we have discussed how the cause-to-effect can be estimated using the counterfactual approach to causal inference involving the exposure to the mediator variables and how it compares to the traditional method. It is worth noting that the technical possibilities of the counterfactuals go beyond the scope of this study discussed above. Intuitively, through counterfactuals we can compute all sorts of effects specific to a pathway. The challenge with this approach is that we may be unsure if there is exposure-mediator interaction and the direction of the interaction. In general, what constitutes relevant features is very crucial when selecting and analyzing casual to effect.  As Pearl and Mackenzie noted in their book~\cite{pearl2018book}, the biggest mistake one can make is mistaking a mediator for a confounder in causal inference and can result in most outrageous error as the latter invites adjustment; the former forbids it.
 
We can exploit the criteria defined in the theories of probabilistic causation~\cite{sep-causation-probabilistic}, to address the challenges of causal inference. Notably, Suppes~\cite{suppes1970theory} explanations provides a better understanding of probabilistic causation by introducing the notion called `prima facie cause'. The definition is based on two principles: (i) \emph{temporal priority} stating that any effect must happen after the cause  (temporal priority), and (ii) \emph{probability raising}, which states that the cause should raise the probability of observing the effect.

\begin{defn}
\label{defn:Probabilistic-causation} (Probabilistic causation~\cite{sep-causation-probabilistic}) Whenever two events involving the cause $e_1$ and effect $e_2$, occurring at times $t_{e1}$ and $t_{e2}$, respectively, under the mild knock down assumptions that their probability ranges:  $0<\rho(e_1), \rho(e_2) < 1$, the event $e_1$ is a prima facie cause of the event $e_2$ if it occurs before the effect and the cause raises
the probability of the effect, such that:
\end{defn}

\begin{equation}\label{eqn:probability-raising}
    t_{e1} < t_{e2} \text{ and } \rho(e_1|e_2) > \rho(e_1|\lnot e_2)
\end{equation}

In the Equation (\ref{eqn:probability-raising}), the \emph{ temporal priority} corresponds to the first condition $t_{e1} < t_{e2}$, where as the second condition: $\rho(e_1|e_2) > \rho(e_1|\lnot e_2)$ is referred to as \emph{probability raising}~\cite{suppes1970theory}. As shown in Equation (\ref{eqn:probability-raising}), to sufficiently claim that event $e_1$ is a cause of event $e_2$ while also satisfying the prima facie cause, the above conditions will require us to know the time information of each event. 
However, because the information about the time is not always available and it is almost impossible to determine this automatically, but the user may have the information about part of the feature. 
We build on the idea that users may have partial or full knowledge about the timing order of the features in the data.  For instance, the features like race and gender may be considered antecedents to most of the features like job, and income, given the time of data collection. Therefore the user may specify that the feature race should be considered in first in the temporal ordering compared to the rest of the features whose timing may be unknown.
To this end, we introduce a notion of full or partial temporal ordering that combines the full or partial temporal ordering manually specified by the user together with the idea of positive statistical dependence, introduced in~\cite{loohuis2014inferring}. The positive statistical ordering translates the timing of the two events as the rate of occurrence (frequency). i.e., for e1 to occur before e2, then:
\begin{equation}\label{eqn:probability-frequency}
p(e_1)>p(e_2) \iff \frac{\rho(e_2|e_1)}{\rho(e_2|\lnot e_1)} >  \frac{\rho(e_1|e_2)}{\rho(e_1|\lnot e_2)}
\end{equation}

I.e., the positive statistical dependence states that for the Equation (\ref{eqn:probability-raising}) to hold between the two events $e_1$ and $e_2$ for which $e_1$  raises the probability of $e_2$ more than $e_2$ raises the probability of $e_1$ (i.e., $\rho(e_1|e_2) > \rho(e_1|\lnot e_2)$), if and only
if $e_1$ is observed more frequently than $e_2$. 

Therefore we require that the ordering specified by the user must satisfy the positive statistical dependence, but in case the ordering is not manually specified as input by the user (or only part of the ordering is used), then only the positive statistical is taken into account. The idea of using manual input for the temporal ordering caters to the case where we have the domain knowledge to capture the timing semantic between the features that may not be reflected as the frequency of occurrence. On one hand, we believe that integrating the domain knowledge is essential, while on the other hand, we do not require the complete knowledge about the time information, for our framework to work. The idea of using partial knowledge about the domain was used before in~\cite{frye2020asymmetric}.


\subsubsection{Divergence measurement Metrics}\label{subsec:distance-functions}

As evidenced by Equations \ref{eqn:CDI}-\ref{eqn:NII}, the causal effects require a notion of divergence (distance) $\Im(\rho_1, \rho_2)\geq 0$ between two probability distributions $\rho_1$ and $\rho_2$ over the output space $\mathcal{Y}$. We now review some possible choices of divergence metrics that will be considered in this study.
\begin{itemize}
\item \textbf{Hellinger Distance:} is defined as
\[
\Im(\rho_1, \rho_2) := D_{HL}(\rho_1||\rho_2) = \frac{1}{\sqrt{2}} \sqrt{\sum_{y\in \mathcal{Y}}\left( \sqrt{\rho_1(y)}-\sqrt{\rho_2(y)}\right)^2}
\]

It is noted that, the value of  Hellinger distance is affected to the greater by the same absolute difference of $\rho_1(y)$ and $\rho_2(y)$ when the value of $f(x)\to \hat{y}$ is small. The Hellinger distance is a measure of sine of the angle between the Hilbert vectors representing the two square-roots $\sqrt{\rho_1(y)}$ and $\sqrt{\rho_2(y)}$ in which each square-roots density is a point on the unit sphere in the Hilbert space.

\item \textbf{Total variation distance:} is defined as

\[
\Im(\rho_1, \rho_2) :=D_{TV}(\rho_1||\rho_2) = \frac{1}{2} \sum_{y\in \mathcal{Y}}\left| \rho_1(y) - \rho_2(y)\right|
\]

The total variation distance is closely related to the Hellinger distance as both distances represent the same topology of the space of probability measures. However, the use of Hilbert spaces (inner products) properties in computing the Hellinger distance gives the Hellinger distance some technical advantages compared to the total variation distance.




\item \textbf{Wasserstein distance:} assuming
$\mathcal{Y}$ is one-dimensional, the distance is
defined as



\[
\Im(\rho_1, \rho_2) :=D_{W}(\rho_1||\rho_2) = \int_{\mathcal{Y}} |F_1(y) - F_2(y)|dy,
\]
where $F_1$ and $F_2$ are the Cumulative Distribution Function (CDF) of the distributions $\rho_1$ and $\rho_2$ respectively. The Wasserstein distance is also known as the earth mover’s distance, because it can be interpreted as the minimum amount of ``work'' required to transform  one distribution into the other. 
\item \textbf{Kullback-Leibler  (KL) divergence:}
is defined as
\[
\Im(\rho_1, \rho_2) :=D_{KL}(\rho_1||\rho_2) =\sum_{y\in \mathcal{Y}} \rho_1(y) \log \left(\frac{\rho_1(y)}{\rho_2(y)}\right).
\]
We note that this divergence measure is not necessarily symmetric \textit{i.e.} $D_{KL}(\rho_1||\rho_2)\neq D_{KL}(\rho_2||\rho_1)$ in general. For this study, we will instead use the modified version called Jensen-Shannon (JS) divergence.
\item  \textbf{Jensen-Shannon (JS) divergence:} is an extension of the KL divergence which aims
at being symmetric and having finite values



\begin{equation}\label{eqn:divergence-ex-js}
    \Im(\rho_1, \rho_2) :=D_{JS}(\rho_1, \rho_2) = \frac{1}{2}D_{KL}(\rho_1||\rho_m) + \frac{1}{2}D_{KL}(\rho_2||\rho_m)
\end{equation}
Where $\rho_m = \frac{\rho_1 + \rho_2}{2}$ is called the mixture distribution.
\end{itemize}

When the two distributions are identical in the Equation (\ref{eqn:divergence-ex-js}), the divergence measure will equate to zero, indicating that the feature is not a potential bias feature. On the other hand, the feature is a potential bias feature if the divergence is greater than zero. Next, we will define the bias machine learning model.





\subsubsection{Data Swapping}\label{subsec:data-swapping}
Having defined measures of divergence $\Im(\cdot,\cdot)$,
we now want to estimate the hypothetical quantities such as $\rho(Y|\text{do}(F_1= f_1, A=a)), \rho(Y|\text{do}(F_1=\lnot f_1, A=a))$ from the actual data and the machine learning model prediction. This estimation will allow us to quantify the impact of each feature on the model prediction, i.e., Controlled Direct Impact following Definition~(\ref{defn:CDI}) and Natural impact following Definition~(\ref{defn:NI}) above, using the data swapping methods.
Willenborg and De Waal~\cite{willenborg2012elements} provide the formal definition of data swapping for $2k$ items in terms of $k$ item swap. They defined data swapping of two data points $i$ and $j$ selected from the same dataset $X$ and interchanging the values of the variable being swapped for these two data points. Visually, we can say that Data swapping involves ``switching the values of columns for two pairs of rows''. To be more precise, we want to condition the random input $\bm{X}$ and the model outcome $Y$ before and after swapping the features in $\bm{X}$. In this case, we are interested in how much the distributions of $Y$ change when we force $X_{ij}$ to take a specific value. This method of forcing the random variable $X_{ij}$ to take a certain value is what we call intervention. With this in mind, we will soon define a``Swapping function'' to evaluate and quantify the bias in the machine learning model. But first of all, we must introduce some key concepts such as the swap ratio and the maximum distortion.

\begin{defn}
    \label{defn:swap-ratio} (Swap Ratio) The swap ratio denoted as $r$, is the percentage number of the records randomly selected from $\bm{X}$ whose features will be swapped.
\end{defn}

The hyper parameter $r$ is manually defined by the user such as $0.1$, or $0.2$ indicating that $10\%$ or $20\%$, respectively, of the random sampled data points are selected for swapping. By default, $0.5$ is used as the standard swap ratio indicating $50\%$ of the data points are selected for swapping

\begin{defn}
    \label{defn:maximum-distortion} (Maximum Distortion) Denoted as $d_\text{max}$ is the maximum allowed statistical distance to quantify how much the data point should change from the original. The idea is that the hypothetical input data should be considered valid only if it is within the accepted range compared to the actual input.
\end{defn}

To compare the distortion with the maximum distortion, we must first quantify the distance between the original and the swapped data points. This quality described in terms of a distortion measure will allow us to know how much the changed data points deviate from the original. Because the input data is likely to contain both the categorical and continuous numerical data, we defined a distance measure for mixed variable data by combining the square Euclidean distance for numeric variables and a simple matching distance for categorical variables, as:

\begin{equation}\label{eqn:distortion}
    d(u,v) = d_{C}(u,v) + d_{N}(u,v)  
\end{equation}

In Equation (\ref{eqn:distortion}), $d_{N}(u,v)$ is the distortion between numeric variables, while the property $d_{C}(u,v)$ is the distortion measure between categorical variables. We however noted that the choice for the distance metric above and the maximum allowed distortion $d_\text{max}$ should be guided by the social implication, with the help of domain experts and stockholder as well the legal consultation such as the 80\% rule.  In our case, we used the Hamming distance to compute the distance for categorical features and for numeric data, we defined a distance measure in the form 
of the ratio of the individual distance measures before $u$ and after swapping $v$, as follows:

\[ d_{C}(u,v) =
  \begin{cases}
    1,       & \quad \text{if } u\neq v\\
    0  & \quad \text{if } u=v
  \end{cases}
\]

\[ 
d_N(u,v) = \|u - v\|
\]
%


It is clear from the Equation (\ref{eqn:distortion}) that the distortion value equal $d_x(u,v)=1$ if only a single binary variable is swapped, and $d_x(u,v)=2$ if the two binary variables are swapped. For this case, the swapped data points will not be subjected to the maximum distortion $d_\text{max}$. The maximum allowed distortion is only used to check for the case where the variable(s) swapped is a continuous variable or a mixture of both continuous and discrete categorical variables.

\begin{defn}
    \label{defn:feature-partitioning} (Feature Values Partitioning) This is a function that splits the values of the given feature, following some binning technique so that the values will belong to only one of the two groups. 
\end{defn}


    Initially all features are assumed to follow a binary categorical format. If there exist continuous or discrete features in $X$, a binning technique is applied to the feature values to ensure the values belong to one of the categories. Specifically, if we denote the different values of the features that we want to partition by $u$. We partition $u$ by the medium point into two categories. 
    Formerly, if we let $C_M = \{C_1, C_2\}$ to be a $M=2$ partition of $u$, then:
    
    \begin{equation}\label{eqn:Bin}
   C_m =  \begin{cases}
    [u_{(1)},  u_{(1)+\delta}     & C_1, \quad \text{if } m = 1\\
    [u_{(1)(M-1)\delta},  u_{(n)}] & C_2,\quad \text{if } m=M=2
  \end{cases}
\end{equation}
      Where $\delta$ is the bin width given by: $\delta = \frac{u_{(n)}-u_{(1)}}{M}$. 
   In Equation (\ref{eqn:Bin}) $\{u_{(i)}\}_{i=1}^n$ denote the ordering of the statistics where the smallest value in the set is $u_{(1)}$ and $u_{(n)}$ is the largest value of the feature, and therefore the $k^{th}$ smallest value in the set is $u_{(k)}$.
 
    We must point out that, even if the bin is of equal width, the number of samples in each category is likely to be non-equal. 
    Also, it is worth pointing out that the actual values of the $x$ were not replaced; instead, this step was only used to help when choosing a random value from a different category during the swapping process. Hence the probability distribution of the data is reserved before the data swapping process. For example, for the dataset containing age feature values between $17$ to $70$ years, the first category may contain all the ages from $17$ to $30$, while $C_2$ are group of age values from $31$, i.e., $C_m=\{\leq 30, >30\}$. A data point with an age value, say 20, can be swapped with any value of age randomly chosen from the category $> 30$. Further details about the swapping process are provided with Definition (\ref{defn:Alternation-function}).


\begin{defn}
    \label{defn:Alternation-function} (Swapping function) is a function that switches between the values of the feature under investigation so that the values of that feature are swapped each time, under some set of constraints.
\end{defn}

    The swapping function denoted as $\varphi$ takes  as input the dataset $\bm{X}$ (i.e., test dataset, for our case), the index of the feature $j$ to swap, the set of indices of instances $I\subset \{1,2,\ldots, n\}$ to swap, and the maximum allowed distorsion $d_\text{max}$. This function returns an alternative dataset $\bm{X}'$ i.e.
    \begin{equation}
        \bm{X}'=\varphi(\bm{X}, j, I, d_\text{max}),
    \end{equation}
    such that
    \begin{equation}
        X'_{ik} = \left\{
         \begin{array}{ll}
           \lnot X'_{ij} \,\,\,\,\,&\text{if } i\in I' \,\,\text{and}\,\,k=j\\\
           \,\,X_{ik}  \,\,\,\,\,&\text{Otherwise}
         \end{array}
       \right.,
    \end{equation}
   where $I'\subseteq I$ is a subset that is chosen in order to ensure that:
    \begin{equation}
        d(\bm{X}, \bm{X}')\leq d_\text{max}.
    \end{equation}
    Simply put, the given feature under investigation $X_{ij}$ is changed with the alternative value: $\lnot X_{ij}$. The alternative value is determined by first identifying the category (i.e., $C_1$ or $C_2$) $X_{ij}$ belongs following the data partitioning technique, see Definition~(\ref{defn:feature-partitioning}, and Equation (\ref{eqn:Bin})). If $X_{ij}$  belongs to $C_1$ then the new value $\lnot X_{ij}$ is randomly chosen from $C_2$, and vice versa. For the gender feature, for instance, this implies that male values become female and vice versa. In some situations, the pairs of data points for swapping need to meet some condition to be  considered swapping candidate. For example, we may allow the switching of the value of the feature only if the distortion of distribution between the original and the post-swapped data point (as measured using some distance function) is not more than the maximum allowed threshold $d_\text{max}$; see Definition (\ref{defn:maximum-distortion}). For instance, if switching a capital-gain feature of  2,000 with a value of 0 can result in a larger difference, as measured using Equation (\ref{eqn:distortion}), then the swap is not allowed.

    \paragraph{$\bullet$ \textbf{Single Feature Swapping Function}:} The single swapping function denoted as $\varphi(\bm{X}, j, I, d_\text{max})$ aims at identifying the Controlled Direct Impact of the feature on the machine learning model following Definition~(\ref{defn:CDI}). 
    
    

\begin{algorithm}[!ht]
\DontPrintSemicolon
  
  \KwInput{$\bm{X}, j, I, d_\text{max}$} 
  \KwOutput{$\bm{X}'$}
  $n, m = \bm{X}\!.\texttt{shape}$
  
  $\bm{X}' = \texttt{zeros}(n, m)$
  
  \For{$i \gets \texttt{range}(0, n)$}
  {
  
  $\bm{X}'[i, :] = \bm{X}[i, :]$
  
  \If{$i\in I$}
  {
   
   \If{$ d(X_{ij}, \lnot X_{ij}) \leq d_\text{max}$}{
    
    $\bm{X}'[i, j] = \lnot X_{ij}$
    
   }
  }
  }
  
  \Return{$\bm{X}'$}

\caption{Single Feature Swapping function}
\label{algo:single-swap-swap}
\end{algorithm}

    We present the summary of Single feature swapping function in Algorithm~(\ref{algo:single-swap-swap}). 
The function takes as input the dataset $\bm{X}$, the feature column to be swapped $j$, the random swap indices $I$, and the maximum distortion $d_\text{max}$. For every candidate data point chosen for swapping (Line 4-5), the distortion is computed to ensure it is within the allowed range in Line 6. 
Without the loss of generality, $X_{ij}$ and the entire dataset $\bm{X}$ is assumed to follow the binary categorical format. The initial step begins by determining the categories of the feature's values $X_{ij}$ with Equation (\ref{eqn:Bin}). 
The output of this function will be used to estimate the Controlled Direct Impact by computing the statistical distance of the distribution in the prediction of $\bm{X}'$ and $\bm{X}$.
    
    

\paragraph{$\bullet$ \textbf{Double Features Swapping Function}:} The double features swapping function 
aims to quantify the Total Natural Impact of each feature in $\bm{X}$ on the machine learning model following Definition~(\ref{defn:NI}). The function takes as arguments (input) the dataset $\bm{X}$ (i.e., test dataset, for our case), the swap ratio $r$, temporal priority ordering $t_o:F\to N$(i.e., manually or automatically determined; please refer to discussion for Definition (\ref{defn:Probabilistic-causation})), and the maximum allowed distortion $d_\text{max}$ and returns the alternative dataset, which is the dataset with alternative values of a feature at column $j$ and all values of the mediating values at column $m$ switched, keeping the rest of variables unchanged and the swapped rows are subjected to the maximum allowed distortion  and the swap ratio $r$. We must note that, before effecting the double swapping, we first check to ensure that the pairs consisting of the feature and a given mediating variable must satisfy the probabilistic causation; refer to Definition~(\ref{defn:Probabilistic-causation}). To this end, we allow the user to provide the partial or full temporal ordering of the features as input and we constrain the rest of the feature using the Equation (\ref{eqn:probability-frequency}). 
For the candidate pairs satisfying the probability causality defined by the temporal priority ordering $t_o:X_j\to X_m$ (indicating that $X_j$ occurred before $X_m$) and Equation (\ref{eqn:probability-frequency}), the following two scenarios are considered when swapping the values of the features and the mediating variable: 

\begin{itemize}
    \item The first scenario includes investigating the case $\bm{do}(X_{j}=x_j, X_m=(x_m, \lnot x_j))$ vs $\bm{do}(X_j=\lnot x_j, X_m=(x_m, \lnot x_j))$ is about changing only the mediating variable vs changing the values for both the mediating variable and the feature to the alternative values as; $X_j:=\lnot x_j, X_m:=\lnot x_m$, for some selected data points $I$. When we use Algorithm (\ref{algo:single-swap-swap}), this implies calling the Algorithm twice, first with the mediator index $m$ and the selected indices $I$, and get the set of swapped datapoints $\bm{X}'$. Next, we will use the swapping Algorithm (\ref{algo:single-swap-swap}) to swap the feature values at index $j$ for the input $\bm{X}'$. Hence we refer to this swapping function as double features swapping function. The output of the double feature swapping function will be the tuple $(\bm{X}', \bm{X}'')$.  
    Formally, we define the double features swapping of data $\bm{X}$ described above as:
    
    %
    \begin{equation}\label{eqn:double-swap-swap1}
    \begin{aligned}
        \bm{X}' &= \varphi(\bm{X}, m, I, d_\text{max})\\
        \bm{X}'' &= \varphi(\bm{X}', j, I, d_\text{max})
    \end{aligned}
    \end{equation}
    
    
    The difference between the distributions of model prediction of $\bm{X}'$ and $\bm{X}''$, in Equation~(\ref{eqn:double-swap-swap1}) will be used to determine the hypothetical Natural Direct Impact of the feature $X_j$.
    
    \item The second scenario includes studying the case $\bm{do}(X_j=\lnot x_j, X_m=(x_m, x_j))$ vs. $\bm{do}(X_j=\lnot x_j, X_m=(x_m, \lnot x_j))$. In this case the property $(X_j=x_j, X_m=(x_m, \lnot x_j))$ includes changing only the value of feature without changing the mediating values for some selected data points $I$, while the property $\bm{do}(X_j=\lnot x_j, X_m=(x_m, \lnot x_j))$ corresponds to changing both the feature values and the mediating values in $I$ to the alternative values. Formally, we can define the second scenario of double features swapping of data $\bm{X}$ above as:
    
    
    \begin{equation}\label{eqn:double-swap-swap2}
    \begin{aligned}
        \bm{X}' &= \varphi(\bm{X}, j, I, d_\text{max})\\
        \bm{X}'' &= \varphi(\bm{X}', m, I, d_\text{max})
    \end{aligned}
    \end{equation}
    
    
    The outcome of this function will be used as input to the model when estimating the Natural Indirect Impact of the above treatment on the model prediction
    
\end{itemize}



\subsubsection{Evaluate the Potential Bias in Machine learning}

As mentioned earlier, our study aims to identify the potentially bias-inducing features and evaluate the level of bias in a machine learning model following some evaluation metrics. 
We wish to identify all the columns $j$ of $\bm{X}$
that may introduce bias to the machine learning model using some evaluation method. 

\begin{defn}
\label{defn:Potential-bias-inducing-Feature} (Potential bias-inducing Feature (PBF)~\cite{alelyani2021detection,richiardi2013mediation}) Given a swap function $\varphi$, the feature $j\in \{1,2,\ldots d\}$ is said to be a potentially bias-inducing feature when the random variables $\hat{Y}=f(X)$ and $\hat{Y}'=f (\,\varphi(X, j)\,)$ have a large divergence.
\end{defn}

In the Definition (\ref{defn:Potential-bias-inducing-Feature}), $\varphi(X, j)$ denotes the swapping of input feature $j$ in
$X$ under some set of constraints 
using the swapping function $\varphi(\cdot \cdot)$; refer to Definition (\ref{defn:Potential-bias-inducing-Feature}) and Section~\ref{subsec:data-swapping}, for the definition of the swapping function, and the different swapping functions proposed in this study. 
$\hat{Y}'$ is the model prediction on the swapped input $f (\,\varphi(X,j)\,)$.  

\nd Using the loan application as an example, $X$ denotes the dataset of the $n$ loan applicants, and $\hat{y}$ is the model's prediction of whether the applicant should be approved for a loan or denied a load. $x_j$ can represent the gender feature, and $\hat{y}'$ represents the model prediction after swapping the gender value from male to female of the $n$ application.

Next, we want to find the divergence in the original model outcome distribution and the model prediction distribution due to data swapping inputs. The idea here is that we want to quantify the potential impact of each feature as being bias-inducing if it satisfy the property $\Im(\,\rho(Y), \rho(\hat{Y}')\,) = \delta$,   
for some some divergence function $\Im(\cdot,\cdot)$, such as Jensen-Shannon divergence~\cite{jensen-shannon:2004}. 
In this case, the larger value of $\delta$ estimates the larger impact of that feature on the model prediction and is ranked as more bias-inducing. In this study, we aim to estimate the impact of the potential biased feature on the model’s prediction using both the single feature swapping function and the double features swapping functions described above. In this study, we consider four different divergence measures (i.e., Hellinger distance, Total variation distance, Jensen-Shannon divergence, and Wasserstein distance, detailed in Subsection~\ref{subsec:distance-functions} above) to compute the statistical distance between the original data and the post-swapped data. Note that, for the double feature swapping, we will compare the statistical distance between the model prediction of the input data $\bm{X}'$ and $\bm{X}''$ for both scenarios (i.e., Equation~\ref{eqn:double-swap-swap1}, and Equation~\ref{eqn:double-swap-swap2}). On the other hand, for the single feature swapping function, we will be computing the statistical distance between predictions of $\bm{X}$ and $\bm{X}'$. 

\begin{defn}
\label{defn:Bias-Machine-Learning-Model} 
(Bias Machine Learning Model) A machine learning is considered biased if it depends on one or more potential bias feature (PBF), either directly or indirectly through some mediating variable (s), given the actual model outcome~\cite{hardt2016equality,agarwal2018reductions}
\end{defn}

We want to quantify the PBF that directly or indirectly impacts the outcome of the machine learning model and evaluate the level of bias induced in the model. To determine if a given feature is a PBF, we compute the statistical difference of the distributions of the model prediction before $\hat{Y}$ and after changing the value of the PBF $\hat{Y}'$, following the set of steps as described above. The larger divergence indicates the greater difference between the distributions of the model outcomes, which in turn suggests the high impact of the feature on the model prediction and likewise
the presence of bias.

\section{Experimental Evaluation}\label{sec:experiment}

The goal of this study is to detect the potentially biased features and empirically evaluate the bias in machine learning using the proposed framework described in Section~\ref{sec:contribution}. In the sections, we used the proposed framework to detect the biased features and evaluate the machine learning using the real-world dataset. Specifically, we want to evaluate our technique based on the following two research questions:

\begin{itemize}

    
    
    \item[\textbf{RQ1}: ] \textbf{Can we identify which features potentially introduce bias to the model? 
    }
    
    This research question aim to examines the features that directly and indirect introduces bias to the model  using our proposed bias detection techniques.
     \item[\textbf{RQ2}: ] \textbf{Given these identified bias-inducing features, 
     can we assert whether or not they are important to the model?}
    
    This research question aims to examine if the bias-inducing or least bias-inducing features are indeed important features to the model. Using the SHAP value, we want to demonstrate how our proposed swapping functions can be integrated with the state-of-the-art model interpretability tool to help explain the most relevant and the least bias-inducing features. The insights that will be derived from answering this question can help the domain experts choose the features that improve predictive performance while making the models the least biased. 
    
\end{itemize}

By answering the above research question, we demonstrate the potential application of our techniques to detect bias in various datasets and machine learning models that are considered biased. We hope that our results will help design explainable and more reliable machine learning models. 

\subsection{Methodology}

For empirical evaluation of bias in the features given the dataset and machine learning model. We first applied a cross-validation (CV) sampling technique to split the dataset into the training $D_{train}$ and the test set $D_{test}$ into $k$-folds, where each fold contains training and testing samples. This well-known technique will allow our method to generalize to different variability in the input data reflecting real-world situations. In this study, we used $k = 10$ folds where the training fold consists of $90\%$ of the data, while the remaining $10\%$ is used for testing. Specifically, the following are the steps we followed: 
\begin{itemize}
   \item Determine and remove the highly correlated features from the dataset. 
   Previous work\cite{tolosi-bioinformatics:2011} have shown that measuring the feature relevance corresponding to the highly correlated features can lead to incorrect model interpretation and misleading feature ranking; (e.g., in the classical model like logistic regression).
   
    \item Split the dataset into training ($D_{train}$) and test dataset ($D_{test}$) using a $10$-fold cross-validation sampling technique, where $90\%$ is taken as the train set, and $10\%$ is used for testing the model.
    \item Train the machine learning model using the dataset $D_{train}$.   
    
    \item Use the trained model to predict all the data points in the  $D_{test}$ as $\hat{Y}$.
    
    \item Apply data swapping on the test set using both single features swapping and double features swapping functions. 
    
    \item Make the model prediction using single feature swapped test dataset as $\hat{Y}'_S$, and similarly predict using the double features swapped dataset as $\hat{Y}'_D$.
    
    \item Using the distance functions (i.e., discussed in Section~\ref{subsec:distance-functions}), evaluate the divergence of the distributions $\hat{Y}$ and $\hat{Y}'_S$, and independently evaluate the divergence between the distributions of  $\hat{Y}$ and $\hat{Y}'_D$, for each features. For the double swapping function, the Natural Direct Impact and the Natural Indirect Impact are computed separately from the input derived from the two scenarios presented in Equation (\ref{eqn:NDI})and Equation (\ref{eqn:NII}), respectively, then the total is reported.
    
    \item The features that return the higher divergences value indicate the larger the bias of that feature on the machine learning model.
\end{itemize}

\subsubsection{Baseline Model}
The proposed framework is flexible, with both classification and regression models as the backbone, including neural networks, logistic regression, and probabilistic classifiers. The work can be extended to other traditional classifiers such as Support Vector Machine (SVM) by simply changing the evaluation metrics e.g., using the propensity score instead of the distance measure.

\subsubsection{SHAP Value}\label{subsec:shap-value}

SHAP (SHapley Additive exPlanations)~\cite{lundberg2017unified} is a state-of-the-art explanability method based on the famous Shapley values from game-theory. Such a technique provides a function 
$\bm{\phi}:\mathcal{X}\rightarrow \mathbb{R}^m$ that, for any input
$x\in\mathcal{X}$, provides a $m$-vector $\bm{\phi}(x)$ whose
components will sum up to
\begin{equation}
    \sum_{j=1}^m \phi_j(x) = f(x) - \mathbb{E}[f(X)].
\end{equation}
That is the difference between the prediction at a specific $x$ and the average prediction is shared among the different features. The vector $\bm{\phi}(x)$ is called the feature attribution and each score $\phi_j(x)$ is meant to convey how
much feature $j$ has contributed to the model output at $x$.
Still, these scores are local and must be aggregated in order
to provide a global sensitivity measure $\bm{\Phi}\in \mathbb{R}^m$. To do so, we average the magnitude of the feature attributions 
\begin{equation}
    \Phi_i = \mathbb{E}[\,|\phi_i(X)|\,],
\end{equation}
which is the default approach in the SHAP Python library~\cite{lundberg2017unified}.

\subsubsection{Dataset}\label{subsec:experiment-dataset}

\begin{itemize}
    
    \item \textbf{Student:} This dataset contains the student performance of the two secondary schools in Portugal. The features include student demographic information, social and school related features. The target variable is final year grade. Specifically the datasets are about the performance in two distinct subjects: Mathematics (mat) and Portuguese language (por). The feature G3 (final year grade) is known to have strong correlation with features G2 (second period grades) and G1 (first period grades). 
    
    \item \textbf{Cleveland Heart Health~\cite{janos_steinbrunn_pfisterer_detrano_1998}:} This dataset contains information about Patients such as the age, sex, ca (number of major vessels), thalach (maximum heart rate), among others from the Cleveland database. The target variable refers to the presence of heart disease in the patient ranging from 0, indicating no presence of the heart disease, to 4 (present).
    
    \item \textbf{COMPAS Recidivism:} This dataset contains over $10,000$ criminal defendants' information used by the COMPAS (Correctional Offender Management Profiling for Alternative Sanctions) algorithm for scoring defendants in Broward County, Florida, for a period of two years. COMPAS is a popular recidivism algorithm used by judges, probation, and parole officers across Florida US State for scoring a criminal defendant’s likelihood of recidivism (reoffending). This database is known to be biased in the sense of having different False Positive Rates between white and black sub-populations \cite{chouldechova2017fair}.
    
    \item \textbf{Bank dataset:}  This dataset is about  marketing campaigns based on phone calls of a Portuguese bank. The goal is to predict whether the client subscribe (yes/no) to the term deposit, used in the Decision Support Systems (DSS)~\cite{moro2014data}.
\end{itemize}




\section{Results}\label{sec:results}
This Section details the experimental results evaluating our proposed approach on four different datasets to answer the two research questions.

\subsection{\textbf{RQ1: Can we identify which features potentially introduce bias to the model?}}\label{subsec:RQ1}
This section reports the results of the experiments using the proposed swapping functions to detect the features that potentially introduce bias to machine learning, answering our \textbf{RQ1}. As described earlier, our swapping functions consist of single feature swapping to estimate the direct impact of each feature on the model prediction under controlled conditions (i.e., Controlled Direct Impact), and the double features swapping function to estimate the features that naturally impact the model prediction through mediating variables (i.e., Total Natural Impacts). 

We studied the impact of swapping each feature using the swapping functions (i.e., single feature swapping functions and double features swapping function) for swapped percentages ($10\%, 30\%, 50\%, \text{ and } 70\%$) of the test dataset, and constrained the swapped input not distorted more than $20\%$ (i.e., $d_{\text{max}} \leq0.2$), motivated by `$80\%$ rule'. The results presented bellow are categorised basing on the dataset used for the experiments, i.e., in the order: \emph{Student dataset}, \emph{Cleveland Heart dataset}, \emph{COMPAS Recidivism dataset}, and \emph{Bank dataset}:

\subsubsection{\textbf{Student Dataset}}
\begin{figure}
     \centering
     \begin{subfigure}[Single feature swapping]{\textwidth}
         \centering
         \includegraphics[width=\textwidth]{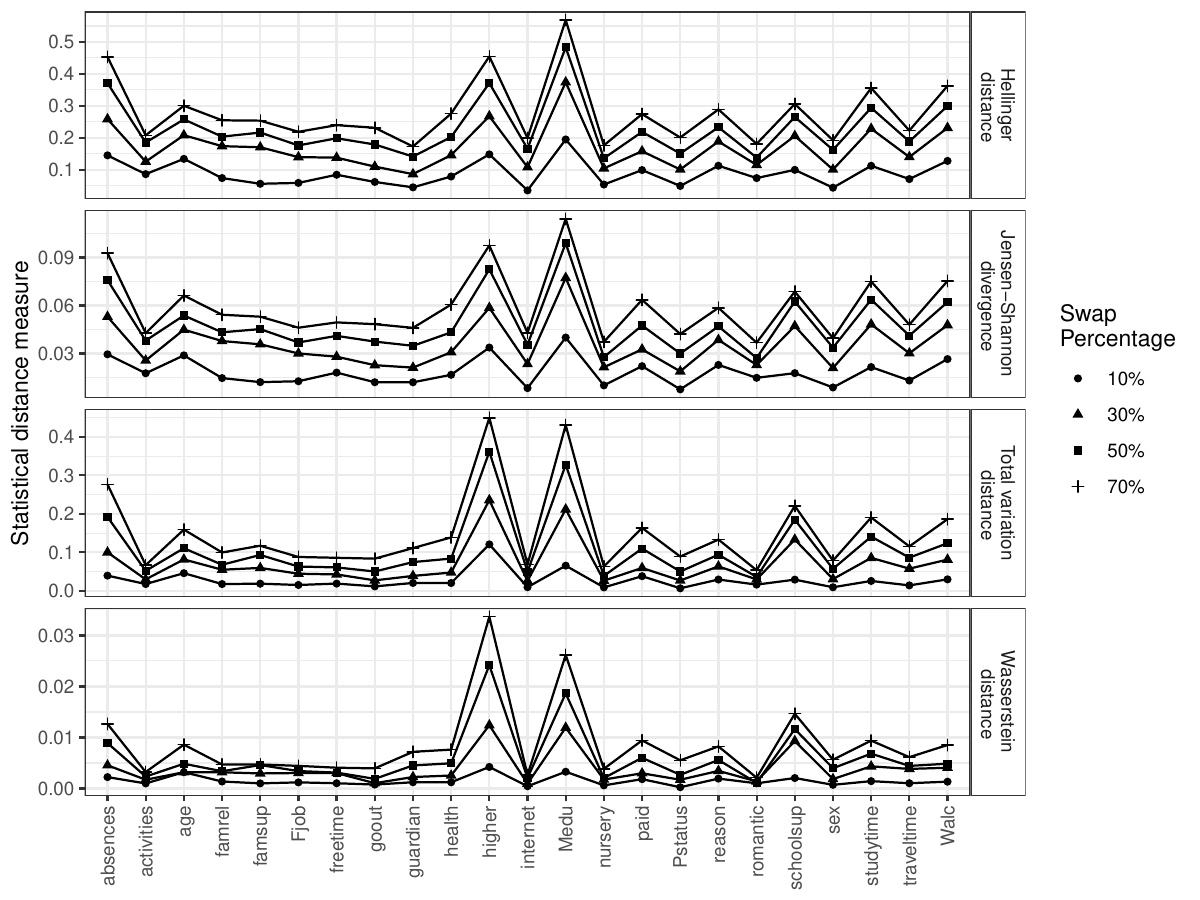}
         \caption{Results of single feature swapping function, showing the direct impact of each feature on the model prediction}
         \label{fig:single-swap-student-data}
     \end{subfigure}
     \hfill
     \begin{subfigure}[Double features swapping]{\textwidth}
         \centering
         \includegraphics[width=\textwidth]{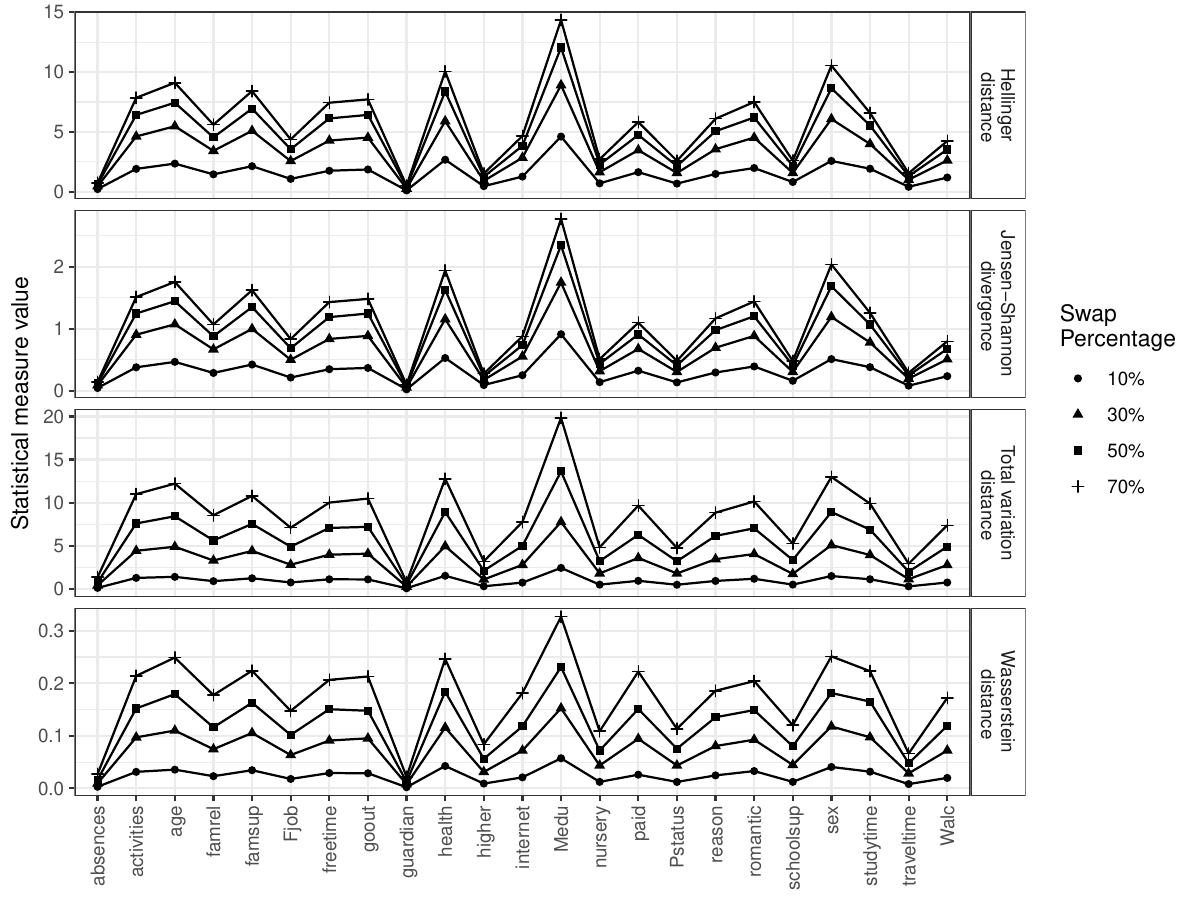}
        
         \caption{The experimental results of the double features swapping function demonstrating the total effect of each feature on the model prediction. For the double features swapping of the student performance dataset, we used the temporal priority (refer to Definition~\ref{defn:Probabilistic-causation}) ordering of: 
         $sex\to age \to activities\to Medu\to health\to famsup\to freetime\to goout\to romantic\to Fjob\to reason\to famrel\to studytime\to paid\to internet\to Walc\to nursery\to Pstatus\to schoolsup\to traveltime\to higher\to guardian\to absences$}
         \label{fig:double-swap-student-data}
     \end{subfigure}
     
        \caption{Distance measure showing the experimental results of the  single feature swapping function in Figure~\ref{fig:single-swap-student-data}, and double features swapping function in Figure~\ref{fig:double-swap-student-data} for the Student's Performance dataset, using the swap percentage of ($10\%, 30\%, 50\%$, and $70\%$)}
        \label{fig:swap-student-data}
\end{figure}

Figure~\ref{fig:swap-student-data} shows the experimental results of our swapping functions comparing the direct impact of each feature on the model prediction and the Total Natural Impact of each feature on the model prediction for the  Student performance dataset.

First, In Figure~\ref{fig:single-swap-student-data} we report the experimental results of the single feature swapping function showing how each feature directly impacts the model prediction when other variables/ features are kept unchanged (Controlled Direct Impact), evaluated using the four different divergence measures: Hellinger, Jensen-Shanon divergence, Total variation distance, and Wasserstein distance.


As seen in Figure~\ref{fig:swap-student-data}, for all four divergence measures, we can observe that the values increase with the increase in the swap percentage. Overall we can see that there is a consistency in the order of feature ranking when values of a single feature are swapped, keeping other features unchanged. As detailed earlier in Section~\ref{subsec:experiment-dataset}, the student dataset contains information about student achievement in secondary education in Portuguese schools where the goal is to predict the student grades. 
on average, in Figures~\ref{fig:single-swap-student-data}, the minimum value of the statistical distance are observed when the feature `nursary', `Pstatus', `Fjob' and `sex' is swapped and the statistical distance is maximum when features `higher', `Medu', or `absences' is swapped. 
The feature `higher' is the binary variable indicating whether a student wants to take higher education ($1$ or yes) or not ($0$ or no), `Medu' is an acronym for mother's education, while `absences' is the numeric feature indicating the number of school absences ranges from 0 to 93. These results indicate that the features `higher', `Medu', or `absences' plays significant role to the model prediction hence highly impact the model compared to the most frequent minimizers features such as: attended nursery school (`nursary'), parent's cohabitation status (`Pstatus'), `romantic' (in a romantic relationship) and `sex'. 

In Figure~\ref{fig:double-swap-student-data} 
shows the results when we perform a double features swapping to investigate the impact of each feature through the mediating variables. 
The temporal priority ordering and the probability raising for the reported results in Figure~\ref{fig:double-swap-student-data}, followed the order: $sex\to age \to activities\to Medu\to health\to famsup\to freetime\to goout\to romantic\to Fjob\to reason\to famrel\to studytime\to paid\to internet\to Walc\to nursery\to Pstatus\to schoolsup\to traveltime\to higher\to guardian\to absences$, where the first two order was defined manually (i.e., $sex\to age$), while the rest of the order followed the condition in Equation(\ref{eqn:probability-frequency}). The results reported in Figure~\ref{fig:double-swap-student-data} is the total sum of the pairwise swapping of each feature and its respective mediating variables. For instance for the feature sex, will be $(sex, age)+(sex,Medu)+(sex,activities)+,.., +(sex,absences)$, refer to Table~\ref{tab:double-swap-student}. According the result in Figure~\ref{fig:double-swap-student-data}, we can observe that the feature `Medu' still indicates the maximum statistical distance, and `age' or `health' are the close second maximal statistical distance. The student sex feature, which originally indicated a minimal impact when a single feature swapping function, now shows a higher impact when the sex and all its mediating variables were swapped using the double features swapping function.

\begin{table}[h]
\Large 
\caption{Summary of natural impact of pairwise  feature (column `Feature') and mediator (column from `age' to `studytime') on the model prediction for \textbf{Student performance dataset}, with temporal priority ordering: $sex\to age \to activities\to Medu\to health\to famsup\to freetime\to goout\to romantic\to Fjob\to reason\to famrel\to studytime\to paid\to internet\to Walc\to nursery\to Pstatus\to schoolsup\to traveltime\to higher\to guardian\to absences$. The maximum values along the rows are highlighted in \textbf{\textit{\large bold face}}. For a clear visualization, some variables are omitted
}.  
     \label{tab:double-swap-student}
        \centering
        \resizebox{\textwidth}{!}{\begin{tabular}{m{2cm} l r  r r m{1.5cm} m{1.5cm} m{1.2cm} m{1.2cm} m{1.2cm} m{1cm} }
        
        \rowcolor{gray!15}
        \textbf{Measure}&\textbf{Feature}&\textbf{age}&\textbf{Medu}&\textbf{health}&\textbf{fam-sup}&\textbf{free-time}&\textbf{study time}&\textbf{school-sup}&\textbf{higher}&\textbf{absences}\\ \midrule
        \midrule

\multirow{7}{*}{\rotatebox[origin=c]{90}{\parbox[c]{3cm}{\centering \textbf{Hellinger distance}}}}&sex&0.34&0.4&0.45&\textbf{\textit{0.56}}&0.38&\textbf{0.47}&0.42&0.18&0.37\\
&age&-&0.36&\textbf{0.39}&\textbf{\textit{0.53}}&0.29&0.37&0.37&0.18&0.33\\
&activities&-&0.28&\textbf{0.38}&\textbf{\textit{0.48}}&0.28&0.35&0.33&0.16&0.3\\
&Medu&-&-&0.66&\textbf{\textit{0.77}}&0.57&\textbf{0.69}&0.49&0.37&0.6\\
&health&-&-&-&\textbf{\textit{0.63}}&0.42&\textbf{0.53}&0.43&0.25&0.47\\
&freetime&-&-&-&-&-&\textbf{0.44}&0.36&0.18&\textbf{0.44}\\
\bottomrule

        \rowcolor{gray!15}
        \textbf{Measure}&\textbf{Feature}&\textbf{age}&\textbf{Medu}&\textbf{health}&\textbf{fam-sup}&\textbf{free-time}&\textbf{study time}&\textbf{school-sup}&\textbf{higher}&\textbf{absences}\\
        \midrule
        \midrule

\multirow{7}{*}{\rotatebox[origin=c]{90}{\parbox[c]{3cm}{\centering \textbf{Jensen-Shannon divergence}}}}&sex&0.06&0.08&0.09&\textbf{\textit{0.11}}&0.08&0.09&0.08&0.04&0.07\\
&age&-&0.06&0.07&\textbf{\textit{0.11}}&0.05&\textbf{0.08}&0.07&0.03&0.07\\
&activities&-&0.06&0.07&\textbf{\textit{0.1}}&0.06&0.07&0.07&0.03&0.06\\
&Medu&-&-&\textbf{0.13}&\textbf{\textit{0.15}}&0.11&\textbf{0.13}&0.1&0.07&0.12\\
&health&-&-&-&\textbf{\textit{0.12}}&0.08&\textbf{0.1}&0.09&0.05&\textbf{0.1}\\
&freetime&-&-&-&-&-&\textbf{0.08}&0.07&0.04&\textbf{\textit{0.09}}\\
\bottomrule

        \rowcolor{gray!15}
        \textbf{Measure}&\textbf{Feature}&\textbf{age}&\textbf{Medu}&\textbf{health}&\textbf{fam-sup}&\textbf{free-time}&\textbf{study time}&\textbf{school-sup}&\textbf{higher}&\textbf{absences}\\
        \midrule
        \midrule

\multirow{7}{*}{\rotatebox[origin=c]{90}{\parbox[c]{3cm}{\centering \textbf{Total variation distance}}}}&sex&0.16&0.21&\textbf{0.31}&\textbf{\textit{0.52}}&0.18&0.3&0.26&0.1&0.2\\
&age&-&0.2&\textbf{0.3}&\textbf{\textit{0.52}}&0.15&0.25&0.25&0.09&0.17\\
&activities&-&0.11&\textbf{0.27}&\textbf{\textit{0.47}}&0.15&0.25&0.21&0.08&0.15\\
&Medu&-&-&\textbf{0.53}&\textbf{\textit{0.73}}&0.37&0.51&0.37&0.26&0.47\\
&health&-&-&-&\textbf{\textit{0.59}}&0.24&\textbf{0.36}&0.27&0.15&\textbf{0.36}\\
&freetime&-&-&-&-&-&\textbf{0.31}&0.24&0.09&\textbf{\textit{0.33}}\\
\bottomrule

        \rowcolor{gray!15}
        \textbf{Measure}&\textbf{Feature}&\textbf{age}&\textbf{Medu}&\textbf{health}&\textbf{fam-sup}&\textbf{free-time}&\textbf{study time}&\textbf{school-sup}&\textbf{higher}&\textbf{absences}\\
        \midrule
        \midrule

\multirow{7}{*}{\rotatebox[origin=c]{90}{\parbox[c]{3cm}{\centering \textbf{Wasserstein distance}}}}&sex&0&0&0.01&\textbf{\textit{0.02}}&0&0.01&0.01&0&0\\
&age&-&0.01&0.01&\textbf{\textit{0.02}}&0&0.01&0.01&0&0\\
&activities&-&0&0.01&\textbf{\textit{0.02}}&0&0.01&0.01&0&0\\
&Medu&-&-&\textbf{0.02}&\textbf{\textit{0.03}}&0&\textbf{0.02}&0.01&0&\textbf{0.02}\\
&health&-&-&-&\textbf{\textit{0.02}}&0&0.01&0.01&0&0.01\\
&freetime&-&-&-&-&-&0.01&0.01&0&0.01\\

\bottomrule

       \end{tabular}}
    \end{table}

Table~\ref{tab:double-swap-student} detailed breakdown of the results for double features swapping function, estimating the natural impact of each feature on predicting the student performance, through the mediating variables (i.e., column header). The results shown in Table~\ref{tab:double-swap-student} are only for the $50\%$ swap percentage. Each features (i.e., the second column named `Features') were checked against each mediating variable (column header starting from `age' to `absences'), and the values shown are the sum of estimated natural direct and indirect impact. The results in Table~\ref{tab:double-swap-student} can help us to better understand how the features through the mediating variables contribute to the natural impact on the model prediction, using our proposed double features swapping function. We note that, for clear visualization, some variables are omitted in Table~\ref{tab:double-swap-student}. According to Table~\ref{tab:double-swap-student}, the variables which dominantly shows the maximal causes of mediation in the model prediction and the pairwise (feature, mediator) are: `famsup', `study time', `schoolsup' and `health'. 

\begin{tcolorbox}
Concluding from our results of swapping values of each feature (keeping other features unchanged) of the student data, we can say that the features `higher', `Medu', or `absences' plays a significant role in predicting the student's performance. 

Moreover, the features which are highly impacted by the mediating variables and the model prediction of student performance are frequently the features such as `age', `sex', `health', `Medu', `activities'. 
\end{tcolorbox}

\subsubsection{Cleveland Heart Dataset}

This Subsection detailed the empirical evaluation of our proposed swapping functions on the Cleveland Heart dataset. For the Cleveland Heart dataset, the features for training and testing the machine learning model include: `sex', `age',  `thalach', `ca', `thal', `exang', `cp', `trestbps', `restecg', `fbs', `oldpeak', `chol'. The descriptions of each features can be found in~\cite{janos_steinbrunn_pfisterer_detrano_1998}. Similar to the Student dataset, the rows containing missing values are dropped and correlated features were removed during the preprocessing step. Also, the continuous features were converted to categorical using the feature value partitioning technique in Definition (\ref{defn:feature-partitioning}) The target variable is a binary class referring to the presence or absence of heart disease in the patient.

\begin{figure}
     \centering
     \begin{subfigure}[Single feature swapping]{\textwidth}
         \centering
         \includegraphics[width=\textwidth]{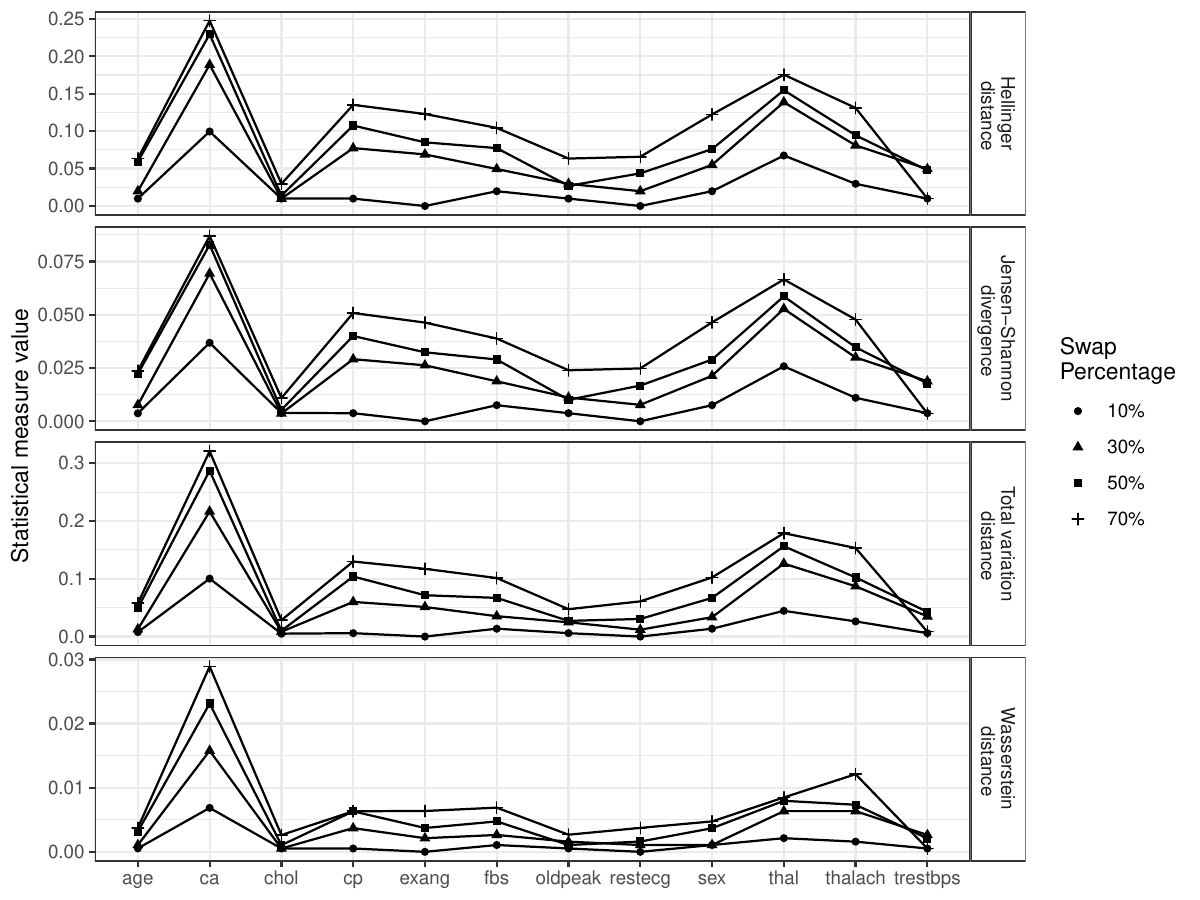}
         \caption{Single feature swapping results, indicating the Controlled Direct Impact of each feature in predicting the presence of heart disease in the patients}
         \label{fig:single-feature-cleveran-heart-data}
     \end{subfigure}
     
     \hfill
     
     \begin{subfigure}[Double features swapping]{\textwidth}
         \centering
         \includegraphics[width=\textwidth]{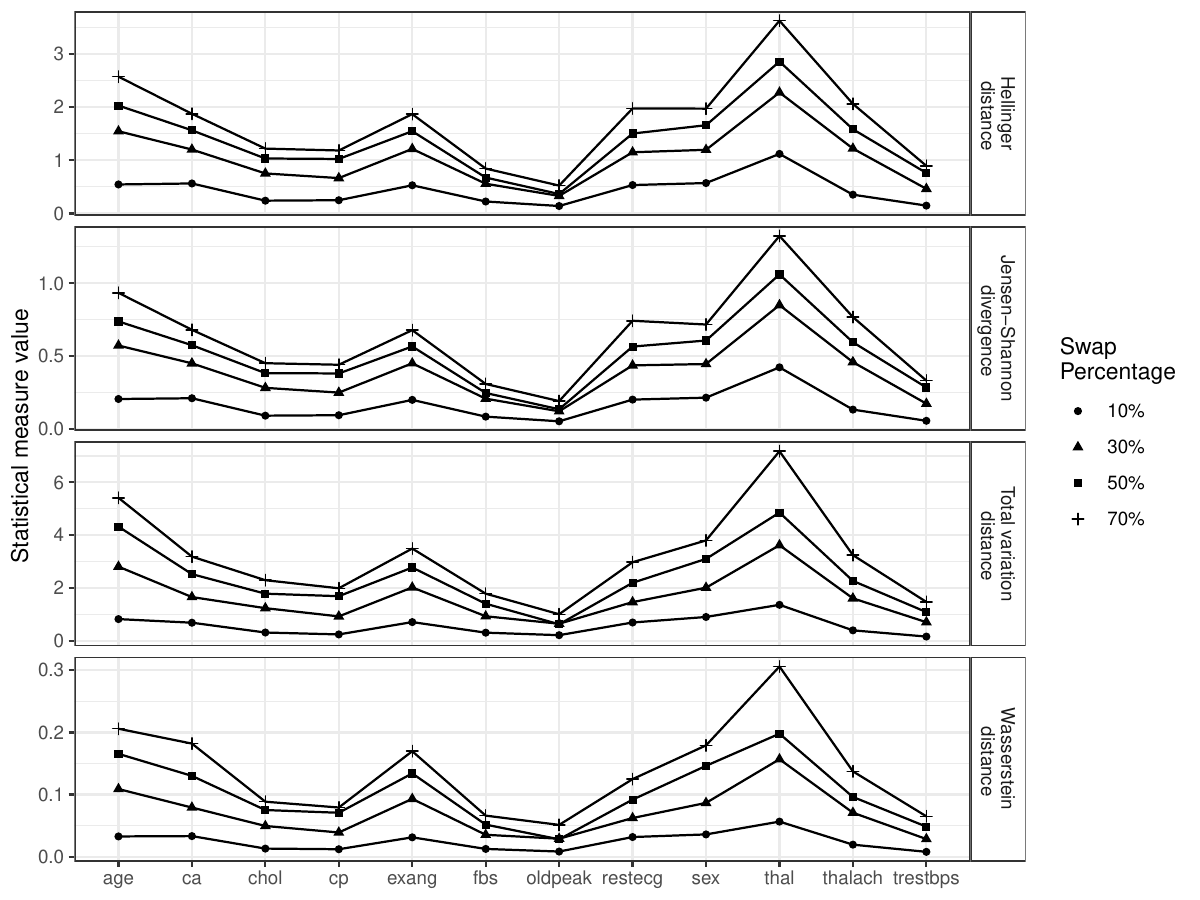}
         \caption{The results of the double features swapping functions to show the Total Natural Impact of each feature on the model prediction. We used the temporal priority (refer to Definition~\ref{defn:Probabilistic-causation}) ordering of: 
         $sex\to age \to restecg\to thal\to exang\to cp\to chol\to thalach\to trestbps\to fbs\to ca\to oldpeak$}
         \label{fig:double-swap-cleveran-heart-data}
     \end{subfigure}
     
        \caption{The experimental results in terms of the statistical distance measure between the distribution of original model prediction and the  single feature swapping function in Figure~\ref{fig:single-feature-cleveran-heart-data}, and double features swapping functions in Figure~\ref{fig:double-swap-cleveran-heart-data} for the Cleveland Heart dataset, with the swap percentage of ($10\%, 30\%, 50\%$, and $70\%$)}
        \label{fig:swap-cleveran-heart-data}
         
\end{figure}

Figure~\ref{fig:swap-cleveran-heart-data} shows the  results on the Cleveland Heart dataset of our swapping functions comparing the direct impact of each feature on the model prediction and the Total Natural Impact of each feature on the model prediction. 

In Figure~\ref{fig:single-feature-cleveran-heart-data}, we show the results of the single feature swapping function on Cleveland Heart dataset to demonstrate the Controlled Direct Impact of each of the features on predicting the presence or absence of heart diseases in the patient, using the swap percentage $(10\%, 30\%, 50\%, 70\%)$. 
Similar to the Student performance dataset reported earlier, as shown in Figure~\ref{fig:single-feature-cleveran-heart-data}, the statistical distance between the model prediction before and after swapping each features increases with the swap proportion. These remains  consistent for all the four divergence measures. The ordering of the distance measures in Figure~\ref{fig:single-feature-cleveran-heart-data} are close similar, whereby the feature `cp', `ca', and `thal' shows the maximal values of statistical distances, while the feature `fbs' and `restecg' demonstrate the minimal values. These results indicate that the features cp, ca, and thal have the highest Controlled Direct Impact on the model prediction of the presence of patient's heart diseases. 

Figure~\ref{fig:double-swap-cleveran-heart-data} report the Total Natural Impact of each features of the Cleveland Heart dataset, estimated using our proposed double features swapping functions, with the temporal priority ordering: $sex\to age \to restecg\to thal\to exang\to cp\to chol\to thalach\to trestbps\to fbs\to ca\to oldpeak$. The  double features swapping estimate the Total Natural Impact of each feature through the mediating variables. The reported results in Figure~\ref{fig:double-swap-cleveran-heart-data} indicates the total sum of the pairwise swapping of each feature and its respective mediating variables; refer to Table~\ref{tab:double-swap-cleveran-heart-data}. According to Figure~\ref{fig:double-swap-cleveran-heart-data}, the features that consistently show the higher Total Natural Impact on the model predictions across the four divergence measures are in the order of maximal: `thal', `age', `ca', `thalach', `sex', and `exang'. Some of these features, e.g., `sex', and `age', originally were not captured as highly impacting the model prediction when the single feature swapping function was used, implying that the natural relation between the `sex' features and other features (also called mediating variable) that causes mediation in the model prediction was not captured by single feature swapping.

\begin{table}[h]
\Large 
\caption{Summary of the natural effect estimated using the double features swapping on $50\%$ of the \textbf{Cleveland Heart dataset}. The mediating variables are shown as column header in the column (from `age' to chol). We used the temporal priority ordering: $sex\to age \to restecg\to thal\to exang\to cp\to chol\to thalach\to trestbps\to fbs\to ca\to oldpeak$. 
The maximum row value is highlighted in highlighted in \textbf{\textit{\large bold face}}.
}.  
     \label{tab:double-swap-cleveran-heart-data}
     
    \begin{subtable}[h]{\textwidth}
        \centering
        \Large
        \resizebox{\textwidth}{!}{\begin{tabular}{m{2.5cm} l r  r r r r r r r r r r}
        \rowcolor{gray!15}
        \textbf{Measure}&\textbf{Feature}&\textbf{age}&\textbf{restecg}&\textbf{thal}&\textbf{exang}&\textbf{cp}&\textbf{chol}&\textbf{thalach}&\textbf{trestbps}&\textbf{fbs}&\textbf{ca}&\textbf{oldpeak}\\ \midrule
        \midrule

\multirow{8}{*}{\rotatebox[origin=c]{90}{\parbox[c]{3cm}{\centering \textbf{Hellinger distance}}}}&sex&0.16&0.26&\textbf{0.36}&0.24&0.34&0.27&0.3&0.29&0.11&\textbf{\textit{0.39}}&0.21\\
&age&-&-&\textbf{\textit{0.34}}&0.21&0.19&0.21&0.23&0.25&0.05&0.24&0.15\\
&restecg&0.11&-&\textbf{\textit{0.28}}&0.16&0.14&-&-&0.27&0.09&0.13&0.18\\
&thal&-&-&-&0.33&0.32&-&\textbf{\textit{0.35}}&0.34&0.27&0.3&0.31\\

&exang&-&-&-&-&\textbf{\textit{0.33}}&-&-&0.28&0.15&0.3&0.24\\
&cp&-&-&-&-&-&-&-&\textbf{\textit{0.36}}&0.19&0.27&0.18\\
&chol&-&0.09&-&0.08&0.06&-&-&-&-&\textbf{0.11}&-\\
&thalach&-&\textbf{\textit{0.27}}&-&0.23&0.24&-&-&-&-&-&0.22\\
\bottomrule

       \end{tabular}}
    \end{subtable}
\hfill
\begin{subtable}[h]{\textwidth}
       \Large
        \resizebox{\textwidth}{!}{\begin{tabular}{m{2.5cm} l r  r r r r r r r r r r}
        
        \rowcolor{gray!15}
        \textbf{Measure}&\textbf{Feature}&\textbf{age}&\textbf{restecg}&\textbf{thal}&\textbf{exang}&\textbf{cp}&\textbf{chol}&\textbf{thalach}&\textbf{trestbps}&\textbf{fbs}&\textbf{ca}&\textbf{oldpeak}\\ \midrule
        \midrule

\multirow{8}{*}{\rotatebox[origin=c]{90}{\parbox[c]{3cm}{\centering \textbf{Jensen-Shannon divergence}}}}&sex&0.06&0.1&\textbf{0.13}&0.1&0.12&0.09&0.1&0.1&0.04&\textbf{\textit{0.14}}&0.08\\
&age&-&-&\textbf{0.13}&0.08&0.07&0.08&0.08&0.09&0.02&0.09&0.06\\
&restecg&0.04&-&\textbf{\textit{0.11}}&0.06&0.05&-&-&0.09&0.03&0.05&0.07\\
&thal&-&-&-&\textbf{\textit{0.13}}&0.12&-&\textbf{0.13}&\textbf{\textit{0.13}}&0.1&0.12&0.11\\
&exang&-&-&-&-&\textbf{\textit{0.12}}&-&-&0.1&0.06&0.1&0.09\\
&cp&-&-&-&-&-&-&-&\textbf{\textit{0.14}}&0.07&0.1&0.07\\
&chol&-&\textbf{\textit{0.04}}&-&0.03&0.02&-&-&-&-&\textbf{0.04}&-\\
&thalach&-&\textbf{\textit{0.1}}&-&0.08&0.09&-&-&-&-&-&0.08\\
\bottomrule

       \end{tabular}}
    \end{subtable}
\hfill
\begin{subtable}[h]{\textwidth}
       \Large
        \resizebox{\textwidth}{!}{\begin{tabular}{m{2.5cm} l r  r r r r r r r r r r}
        
        \rowcolor{gray!15}
        \textbf{Measure}&\textbf{Feature}&\textbf{age}&\textbf{restecg}&\textbf{thal}&\textbf{exang}&\textbf{cp}&\textbf{chol}&\textbf{thalach}&\textbf{trestbps}&\textbf{fbs}&\textbf{ca}&\textbf{oldpeak}\\ \midrule
        \midrule

\multirow{8}{*}{\rotatebox[origin=c]{90}{\parbox[c]{3cm}{\centering \textbf{Total variation distance}}}}&sex&0.13&0.29&0.36&0.21&\textbf{0.41}&0.33&0.31&0.33&0.11&\textbf{\textit{0.43}}&0.21\\
&age&-&-&\textbf{\textit{0.59}}&0.19&0.23&0.25&0.24&0.29&0.04&0.21&0.13\\
&restecg&0.08&-&\textbf{\textit{0.26}}&0.13&0.13&-&-&\textbf{0.33}&0.07&0.11&0.16\\
&thal&-&-&-&\textbf{\textit{0.39}}&0.34&-&0.41&0.37&0.3&0.29&0.36\\
&exang&-&-&-&-&\textbf{\textit{0.42}}&-&-&0.33&0.17&0.33&0.23\\
&cp&-&-&-&-&-&-&-&\textbf{\textit{0.41}}&0.22&0.26&0.19\\
&chol&-&0.08&-&0.08&0.04&-&-&-&-&\textbf{\textit{0.11}}&-\\
&thalach&-&\textbf{\textit{0.26}}&-&0.2&0.25&-&-&-&-&-&0.22\\
\bottomrule

       \end{tabular}}
    \end{subtable}
\hfill
\begin{subtable}[h]{\textwidth}
       \Large
        \resizebox{\textwidth}{!}{\begin{tabular}{m{2.5cm} l r  r r r r r r r r r r}
        
        \rowcolor{gray!15}
        \textbf{Measure}&\textbf{Feature}&\textbf{age}&\textbf{restecg}&\textbf{thal}&\textbf{exang}&\textbf{cp}&\textbf{chol}&\textbf{thalach}&\textbf{trestbps}&\textbf{fbs}&\textbf{ca}&\textbf{oldpeak}\\ \midrule
        \midrule

\multirow{8}{*}{\rotatebox[origin=c]{90}{\parbox[c]{3cm}{\centering \textbf{Wasserstein distance}}}}&sex&0&0.02&\textbf{0.03}&0.01&\textbf{\textit{0.04}}&\textbf{0.03}&0.02&\textbf{0.03}&0&\textbf{0.03}&0.01\\
&age&-&-&\textbf{\textit{0.06}}&0.01&0.01&0.01&0.01&0.01&0&0.02&0\\
&restecg&0&-&0.02&0&0.01&-&-&\textbf{\textit{0.03}}&0&0.01&0.01\\
&thal&-&-&-&0.02&0.02&-&\textbf{0.03}&\textbf{0.03}&0.02&0.02&0.02\\
&exang&-&-&-&-&\textbf{\textit{0.04}}&-&-&0.03&0.01&0.03&0.02\\
&cp&-&-&-&-&-&-&-&\textbf{\textit{0.03}}&0.02&0.02&0.01\\
&chol&-&0&-&0&0&-&-&-&-&0.01&-\\
&thalach&-&0.02&-&0.01&0.02&-&-&-&-&-&0.01\\

\bottomrule

       \end{tabular}}
    \end{subtable}
    
\end{table}

The detailed summary of the features and the respective mediating variables are shown in Table~\ref{tab:double-swap-cleveran-heart-data}, for $50\%$ swap percentage. The variables in the column header names starting from `age' to  `oldpeak' are the mediating variables for the features listed in column `Feature'. The bold and italic font face are the maximal value along the rows and the bold face (without italic) are the second highest values. According to Table~\ref{tab:double-swap-cleveran-heart-data}, the variables that frequently causes the higher values of mediation between the  prediction of the presence of heart diseases and the given features are the variables: `thal' `ca', `cp', `restecg', and `restbps'. Notably, the mediating variables between `sex' feature and model prediction are frequently mediated by the variables `ca' (number of major vessels), `cp' (chest pain type) and `thal'.

\begin{tcolorbox}
As shown from our results, the features such as chest pain type (cp), number of major vessels (ca), and `thal'  have the most effect on the model prediction of the presence of heart disease when swapping values of each feature (keeping other features unchanged). On the other hand, the features such as resting electrocardiographic results (restecg), and fasting blood sugar (fbs) frequently show minimal impact on predicting the patients' heart disease. 

The features that are observed to cause the maximal mediation between the model prediction and other features are: `thal', `ca', and `cp'. 

Moreover, the existence of mediating variables causes the features such as `age', `sex', `ca', and `thal' to introduce more bias to the machine learning model trained on the Cleveland Heart dataset.
\end{tcolorbox}

\subsubsection{COMPAS Recidivism Dataset}

\begin{figure}
     \centering
     \begin{subfigure}{\textwidth}
         \centering
         \includegraphics[width=\textwidth]{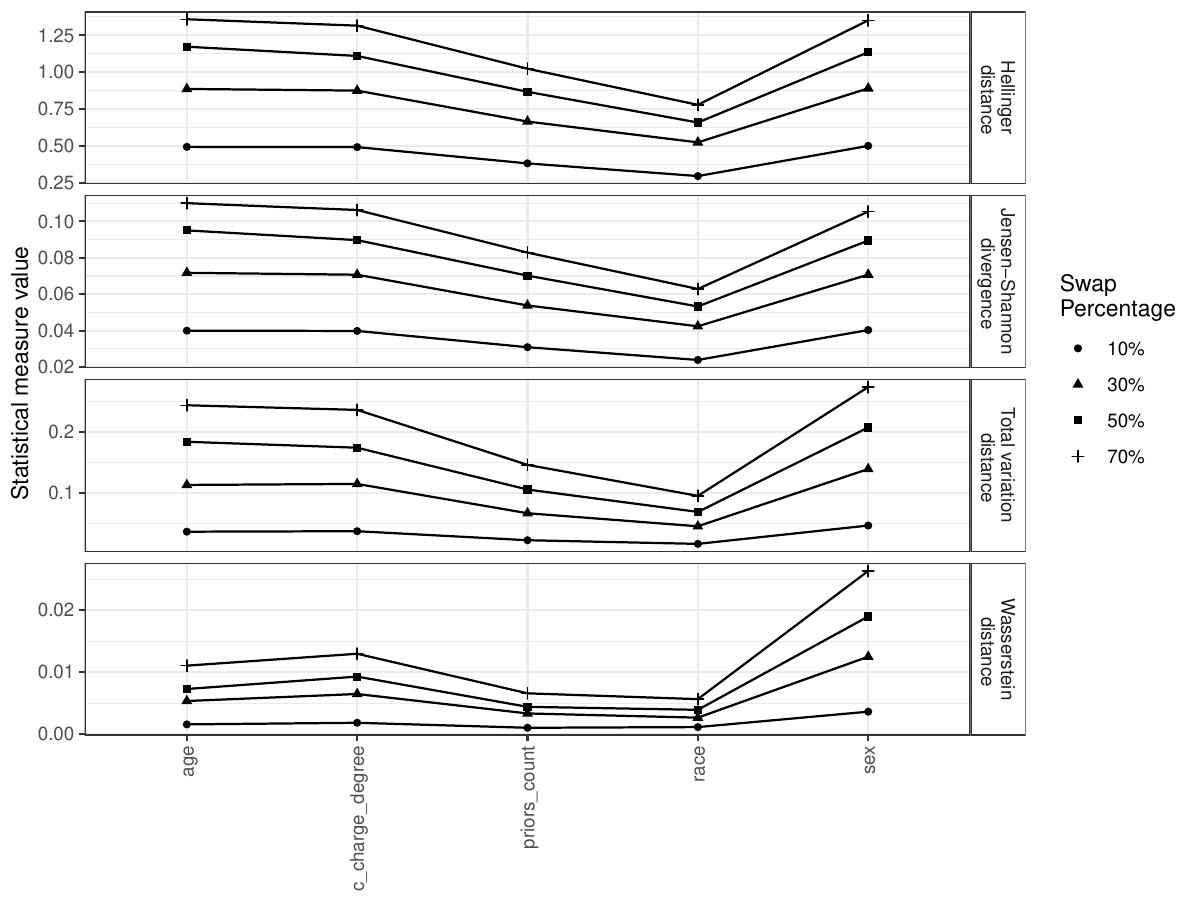}
         \caption{The results of swapping values of single feature swapping, demonstrating the Controlled Direct Impact of the feature on the model prediction}
         \label{fig:single-feature-compas-data}
     \end{subfigure}
     
     \hfill
     \begin{subfigure}{\textwidth}
         \centering
         \includegraphics[width=\textwidth]{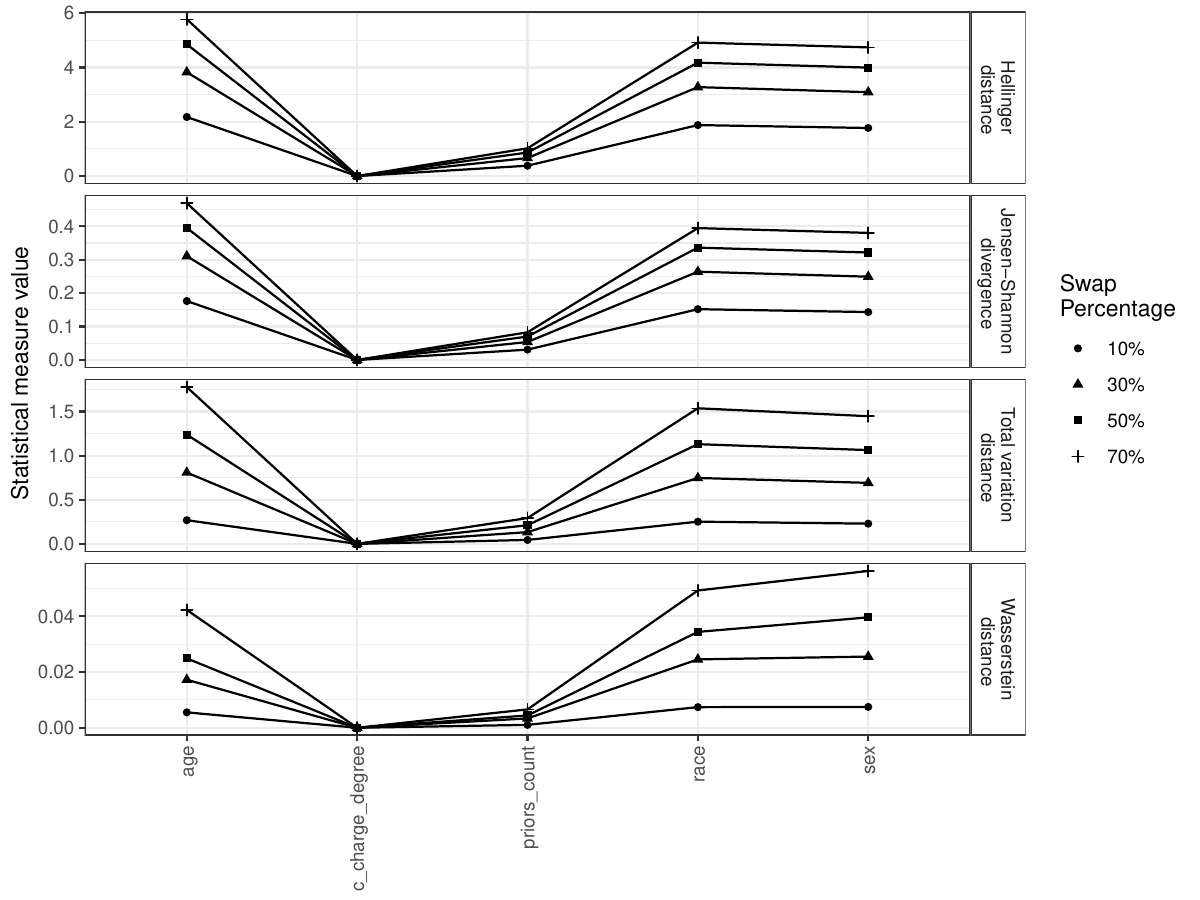}
         \caption{The results of the double features swapping function to show the Total Natural Impact of each feature on the model prediction. We used the temporal priority (refer to Definition~\ref{defn:Probabilistic-causation}) ordering of: 
         $race\to sex \to age\to c\_charge\_degree\to priors\_count$}
         \label{fig:double-swap-compas-data}
     \end{subfigure}
     
        \caption{The experimental results in terms of the statistical distance measure between the distribution of original model prediction and the  single feature swapping function in Figure~\ref{fig:single-feature-compas-data}, and double features swapping function in Figure~\ref{fig:double-swap-compas-data} for the Cleveland Heart dataset, with the swap percentage of ($10\%, 30\%, 50\%$, and $70\%$)}
        \label{fig:swap-compas-data}
\end{figure}

This Subsection report the results of experimenting with our proposed swapping functions on the COMPAS Recidivism dataset. Like the other dataset used in this study, we fit a logistic regression model on the modified COMPAS Recidivism dataset. Prior to fitting the model, a number of preprocessing was done including removing the duplicated and correlated features (e.g., `decile\_score'), dropping features such as `id', `name', `first', `last', `compas\_screening\_date',`dob'. Changing symbolics to numerics. The final features used to train and test the model include sex, race, priors\_count, c\_charge\_degree, and age. The goal variable includes predicting whether the criminal is likely to offend in the future or not (recidivism/crime risk). Figure~\ref{fig:swap-compas-data} report the experimental results of the proposed swapping function on the COMPAS Recidivism dataset comparing the Controlled Direct Impact Figure~\ref{fig:single-feature-compas-data} and the Total Natural Impact (Figure~\ref{fig:double-swap-compas-data}) of each feature on the model prediction. 

Figure~\ref{fig:single-feature-compas-data} demonstrates the Controlled Direct Impact of the features on the model prediction when the swap percentage are: $(10\%, 30\%, 50\%, 70\%)$. 
As seen in Figure~\ref{fig:single-feature-compas-data}, the statistical distance increases with the swap percentage for all the four divergence measures. Also, the ordering of the distance measures in Figure~\ref{fig:single-feature-compas-data} is closely similar across all the four divergence measures, where the features: `age', `sex', and `c\_charge\_degree' shows the maximal values of statistical distance measure. On the other hand, the feature `race' shows the minimal values of statistical distance for a single feature swapping function.  
\begin{table}[h]
\caption{Summary of the statistical distance measure for the $50\%$ swap percentage of the double features swapping on \textbf{COMPAS Recidivism dataset}, demonstrating the impact of each feature on the first columns on the model prediction through the mediating variables (i.e., first rows), with temporal priority ordering: $race\to sex \to priors\_count\to c\_charge\_degree$. \\
The \textbf{\textit{\large bold face+italic}}, highlights the maximal value}. 
\label{tab:double-swap-compas-data}.

        \centering
        \resizebox{\textwidth}{!}{\begin{tabular}{m{1.5cm} l  r r r r }
         \rowcolor{gray!15}
        \textbf{Measure}&\textbf{Feature}&\textbf{sex}&\textbf{age}&\textbf{priors\_count}&\textbf{c\_charge-degree}\\ \midrule
\multirow{4}{*}{\rotatebox[origin=c]{90}{\parbox[c]{1.3cm}{\centering \textbf{Hellinger distance}}}}&race&0.66&\textbf{\textit{1.52}}&1.14&0.87\\
&sex&-&\textbf{\textit{2}}&1.13&0.87\\
&age&-&-&\textbf{\textit{2.58}}&2.27\\
&priors\_count&-&-&-&0.87\\
\bottomrule

         \rowcolor{gray!15}
        \textbf{Measure}&\textbf{Feature}&\textbf{sex}&\textbf{age}&\textbf{priors\_count}&\textbf{c\_charge-degree}\\
        \midrule
\multirow{4}{*}{\rotatebox[origin=c]{90}{\parbox[c]{1.3cm}{\centering \textbf{Jensen-Shannon divergence}}}}&race&0.05&\textbf{\textit{0.12}}&0.09&0.07\\
&sex&-&\textbf{\textit{0.16}}&0.09&0.07\\
&age&-&-&\textbf{\textit{0.21}}&0.19\\
&priors\_count&-&-&-&0.07\\
\bottomrule

         \rowcolor{gray!15}
        \textbf{Measure}&\textbf{Feature}&\textbf{sex}&\textbf{age}&\textbf{priors\_count}&\textbf{c\_charge-degree}\\
        \midrule

\multirow{4}{*}{\rotatebox[origin=c]{90}{\parbox[c]{1.3cm}{\centering \textbf{Total variation distance}}}}&race&0.07&\textbf{\textit{0.23}}&0.17&0.11\\
&sex&-&\textbf{\textit{0.32}}&0.17&0.11\\
&age&-&-&\textbf{\textit{0.43}}&0.32\\
&priors\_count&-&-&-&0.11\\
\bottomrule

         \rowcolor{gray!15}
        \textbf{Measure}&\textbf{Feature}&\textbf{sex}&\textbf{age}&\textbf{priors\_count}&\textbf{c\_charge-degree}\\
        \midrule

\multirow{4}{*}{\rotatebox[origin=c]{90}{\parbox[c]{1.3cm}{\centering \textbf{Wassers-tein distance}}}}&race&0&\textbf{\textit{0.02}}&0.01&0\\
&sex&-&\textbf{\textit{0.03}}&0.01&0\\
&age&-&-&\textbf{\textit{0.02}}&0.01\\
&priors\_count&-&-&-&0\\

\bottomrule

       \end{tabular}}
    
\end{table}

Figure~\ref{fig:double-swap-compas-data} shows the Total Natural Impact of the features and mediating variables estimated using the double features swapping function on COMPAS dataset, with the temporal priority ordering of: $race\to sex \to age\to priors\_count\to c\_charge\_degree$. As can be seen in Figure~\ref{fig:double-swap-compas-data}, the feature `age' remains maximal, followed by the `sex' and `race', while the feature `c\_charge\_degree' show the minimal statistical distance. The breakdown of the double features swapping function is shown in Table~\ref{tab:double-swap-compas-data}, for the $50\%$ swap percentage. According to Table~\ref{tab:double-swap-compas-data}, the `age' feature indicates the maximal causes of mediation between the model prediction and other features.

\begin{tcolorbox}
The `age' and `sex' features has a maximal statistical distance compared to other features, implying that they exhibit the highest Controlled Direct Impact and the Total Natural Impact on the model prediction. Moreover, the `age' feature plays a significant role as the mediating variable in the COMPAS dataset, causing features such as race and sex to be more biased to the model prediction.
\end{tcolorbox}

     

\subsubsection{Bank Dataset}

\begin{figure}
     \centering
     \begin{subfigure}{\textwidth}
         \centering
         \includegraphics[width=\textwidth]{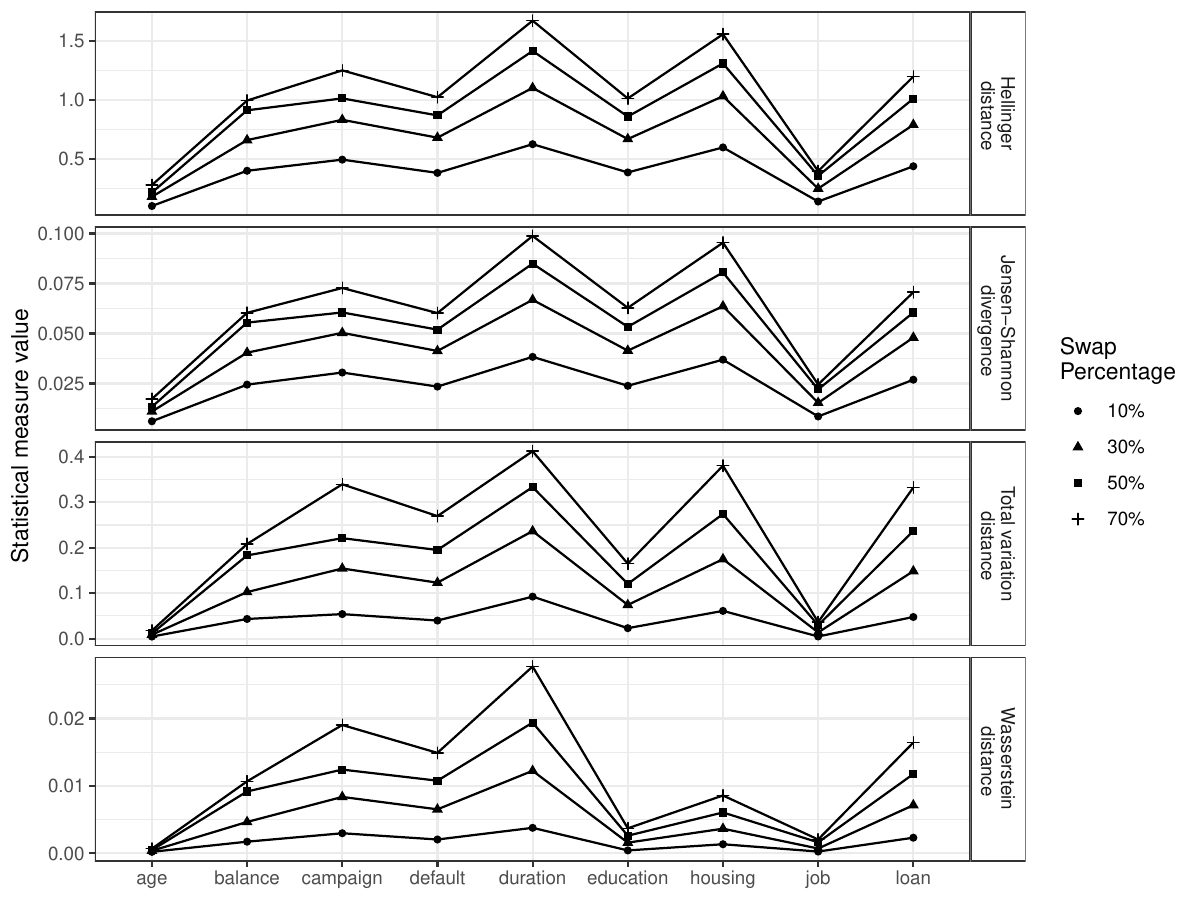}
          \caption{The evaluation results of swapping values of each feature, demonstrating the Controlled Direct Impact of the feature on the model prediction}
         \label{fig:single-feature-bank-data}
     \end{subfigure}
     \hfill
     
     \begin{subfigure}{\textwidth}
         \centering
         \includegraphics[width=\textwidth]{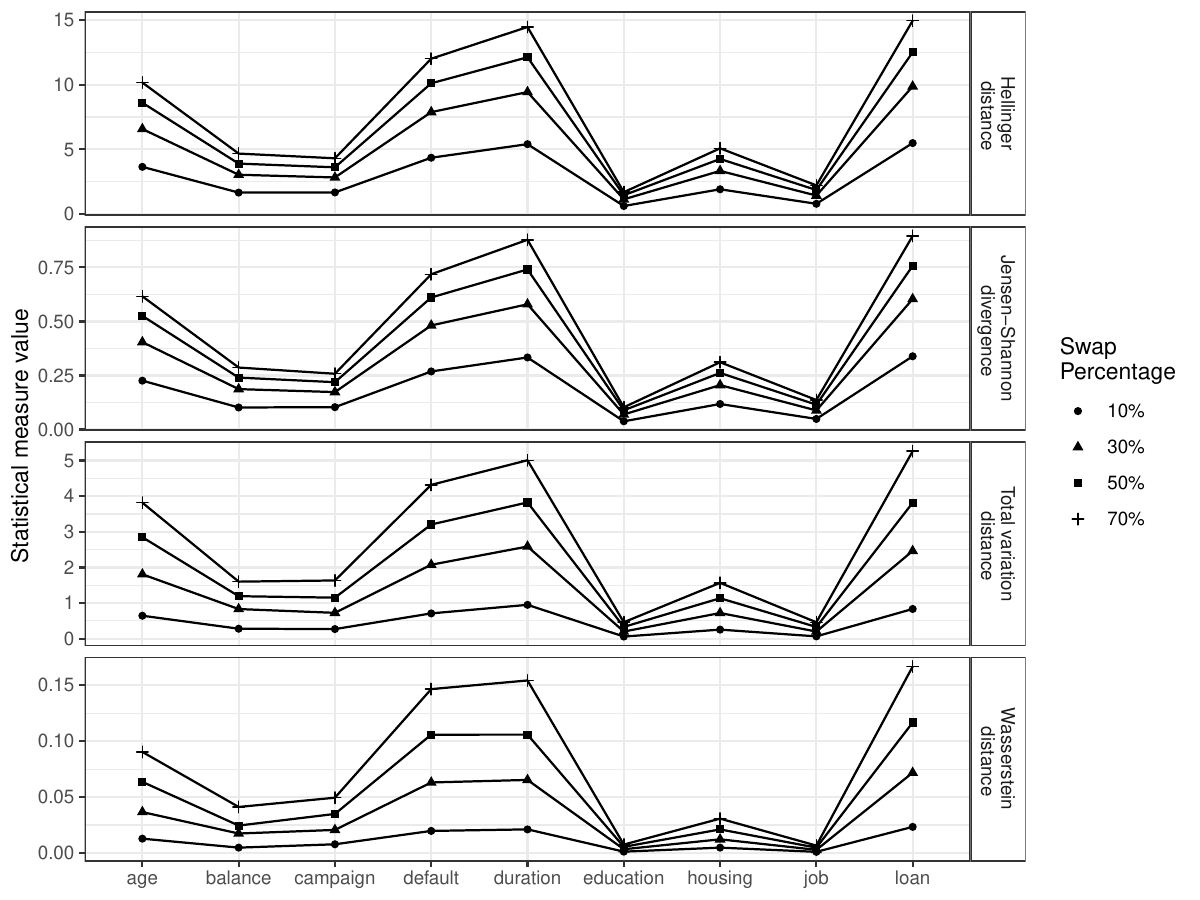}
         \caption{The experimental results of double features swapping function on the bank dataset estimating the Total Natural Impact of each feature on the model prediction. We set the temporal priority (refer to Definition~\ref{defn:Probabilistic-causation}) ordering of: 
         $age \to education\to job\to loan\to balance\to housing\to duration\to campaign\to default$}
         \label{fig:double-swap-bank-data}
     \end{subfigure}
     
        \caption{The experimental results in terms of the statistical distance measure between the distribution of original model prediction and the  single feature swapping function in Figure~\ref{fig:single-feature-bank-data}, and double features swapping function in Figure~\ref{fig:double-swap-bank-data} for the Bank dataset, with the swap percentage of ($10\%, 30\%, 50\%$, and $70\%$)}
        \label{fig:swap-bank-data}
\end{figure}

In the final part of our experiment, we empirically evaluated our proposed swapping functions using the bank dataset and reported the results in this Section. This dataset contains information about the marketing campaigns of a banking institution in Portugal, in which the goal includes predicting if the client would subscribe to a term deposit. The features of the dataset used for training the model include: `age',  `education', `job' (type of job), `loan' (has personal loan?), `balance' (average yearly balance),  `housing' (has housing loan?), `duration' (last contact duration), `campaign' (contacts performed during this campaign), and `default' (has credit in default?). Figure~\ref{fig:swap-bank-data} shows the experimental results of swapping values of the features in the bank dataset, to understand the impact of each features on the model prediction.

In the first part of Figure~\ref{fig:single-feature-bank-data}, we show the results of the single feature swapping function on the bank dataset, estimating the Controlled Direct Impact of each feature on the model prediction. According to Figure~\ref{fig:single-feature-bank-data}, the ordering of the statistical distances is close similar.  Overall, the features that show the higher statistical distances for single feature swapping are: `duration', `loan' and `housing' (except when Wasserstein distance function is used), indicating their high impact on the model prediction. On the other hand, the features `age', `job' and `education' have minimum values of statistical distance, showing how less they impact the model outcomes. 

Next, in Figure~\ref{fig:double-swap-bank-data} we visualize the Total Natural Impact of each feature of the Bank dataset when the temporal priority ordering was set as $housing \to job\to education\to balance\to campaign\to age\to default\to loan\to duration$; refer to Table~\ref{tab:double-swap-bank-data} for the pairwise summary of each feature and the mediating variables. According to Figure~\ref{fig:double-swap-bank-data}, the features that consistently show the higher statistical distance for the double features swapping evaluated across the four divergence measures are in the order of maximal: `loan', `duration', `default', and `age'. Moreover, the `loan' and `duration' features remained consistent as the ones that highly impacted the model prediction for both single and double feature swapping function. From these results, we can say that the model prediction strongly depends on the features `loan' (a binary feature indicating if client has personal loan) and `duration' (last contact duration, numeric) for the model trained on the bank dataset. On the other hand, `age' feature introduces more bias to the model due to the mediating effect.

\begin{table}[h]
\caption{Summary of the statistical distance measure for the $50\%$ swap percentage of the double features swapping on \textbf{Bank dataset}, demonstrating the impact of each feature on the first columns on the model prediction through the mediating variables (i.e., first rows), with temporal priority ordering: $housing \to job\to education\to balance\to campaign\to age\to default\to loan\to duration$. \\
The values are computed by keeping the value of the features (first column) fixed at some level, while altering the variables in the first rows. The maximum row value is highlighted in  \textbf{\textit{\large bold + italic face}}.
}.  
     \label{tab:double-swap-bank-data}
     
        \centering
        \resizebox{\textwidth}{!}{\begin{tabular}{m{1.2cm} l   r r r r r r r r r}
         \rowcolor{gray!15}
        \textbf{Measure}&\textbf{Feature}&\textbf{housing}&\textbf{job}&\textbf{education}&\textbf{balance}&\textbf{campaign}&\textbf{age}&\textbf{default}&\textbf{loan}&\textbf{duration}\\ \midrule
        \midrule

\multirow{7}{*}{\rotatebox[origin=c]{90}{\parbox[c]{2cm}{\centering \textbf{Hellinger distance}}}}&housing&-&1.57&1.71&-&-&\textbf{\textit{2.47}}&1.91&1.88&1.82\\
&job&-&-&0.7&-&\textbf{\textit{1.33}}&0.75&1.06&1.15&-\\
&education&-&-&-&-&\textbf{\textit{2.09}}&-&1.54&1.55&1.46\\
&balance&1.89&\textbf{\textit{2.28}}&-&-&-&-&1.69&2&-\\
&campaign&1.55&-&-&1.65&-&\textbf{2.06}&-&-&\textbf{\textit{2.22}}\\
&age&-&-&0.56&-&-&-&1.02&\textbf{\textit{1.04}}&-\\
&default&-&-&-&-&\textbf{\textit{2.06}}&-&-&-&1.31\\
&loan&-&-&-&-&1.4&-&\textbf{\textit{1.81}}&-&-\\
&duration&-&-&-&-&-&-&-&\textbf{\textit{2.51}}&-\\
\bottomrule

         \rowcolor{gray!15}
        \textbf{Measure}&\textbf{Feature}&\textbf{housing}&\textbf{job}&\textbf{education}&\textbf{balance}&\textbf{campaign}&\textbf{age}&\textbf{default}&\textbf{loan}&\textbf{duration}\\ \midrule\midrule
        \midrule
\multirow{7}{*}{\rotatebox[origin=c]{90}{\parbox[c]{2cm}{\centering \textbf{Jensen-Shannon divergence}}}}&housing&-&0.1&0.11&-&-&\textbf{\textit{0.15}}&0.11&0.11&0.11\\
&job&-&-&0.04&-&\textbf{\textit{0.08}}&0.04&0.06&0.067&-\\
&education&-&-&-&-&\textbf{\textit{0.13}}&-&0.09&0.09&0.09\\
&balance&0.11&0.14&-&-&-&-&0.1&\textbf{\textit{0.12}}&-\\
&campaign&0.1&-&-&0.1&-&\textbf{0.13}&-&-&\textbf{\textit{0.14}}\\
&age&-&-&0.03&-&-&-&\textbf{\textit{0.06}}&\textbf{\textit{0.06}}&-\\
&default&-&-&-&-&\textbf{\textit{0.13}}&-&-&-&0.08\\
&loan&-&-&-&-&0.08&-&\textbf{\textit{0.11}}&-&-\\
&duration&-&-&-&-&-&-&-&0.15&-\\
\bottomrule

         \rowcolor{gray!15}
        \textbf{Measure}&\textbf{Feature}&\textbf{housing}&\textbf{job}&\textbf{education}&\textbf{balance}&\textbf{campaign}&\textbf{age}&\textbf{default}&\textbf{loan}&\textbf{duration}\\ \midrule
        \midrule
\multirow{7}{*}{\rotatebox[origin=c]{90}{\parbox[c]{2cm}{\centering \textbf{Total variation distance}}}}&housing&-&0.27&0.33&-&-&\textbf{\textit{0.5}}&0.32&\textbf{0.38}&0.33\\
&job&-&-&0.06&-&0.27&0.06&0.12&0.23&-\\
&education&-&-&-&-&\textbf{\textit{0.44}}&-&0.22&0.29&0.25\\
&balance&\textbf{\textit{0.44}}&\textbf{\textit{0.44}}&-&-&-&-&0.33&0.43&-\\
&campaign&0.29&-&-&0.28&-&0.41&-&-&0.46\\
&age&-&-&0.03&-&-&-&\textbf{\textit{0.13}}&0.2&-\\
&default&-&-&-&-&\textbf{\textit{0.49}}&-&-&-&0.2\\
&loan&-&-&-&-&0.28&-&\textbf{\textit{0.41}}&-&-\\
&duration&-&-&-&-&-&-&-&0.6&-\\
\bottomrule

         \rowcolor{gray!15}
        \textbf{Measure}&\textbf{Feature}&\textbf{housing}&\textbf{job}&\textbf{education}&\textbf{balance}&\textbf{campaign}&\textbf{age}&\textbf{default}&\textbf{loan}&\textbf{duration}\\ \midrule
        \midrule
\multirow{7}{*}{\rotatebox[origin=c]{90}{\parbox[c]{2cm}{\centering \textbf{Wasserstein distance}}}}&housing&-&0.01&0.01&-&-&\textbf{\textit{0.02}}&0&0.01&0.01\\
&job&-&-&0&-&0.01&0&0&0.01&-\\
&education&-&-&-&-&\textbf{\textit{0.02}}&-&0&0.01&0.01\\
&balance&\textbf{\textit{0.03}}&0.02&-&-&-&-&0.01&0.02&-\\
&campaign&0.01&-&-&\textbf{0.02}&-&0.01&-&-&\textbf{0.02}\\
&age&-&-&0&-&-&-&0&0.01&-\\
&default&-&-&-&-&\textbf{\textit{0.03}}&-&-&-&0.02\\
&loan&-&-&-&-&0.02&-&0.02&-&-\\
&duration&-&-&-&-&-&-&-&0.03&-\\

\bottomrule

       \end{tabular}}
\end{table}

The summary details of each feature and the respective mediating variables are shown in Table~\ref{tab:double-swap-bank-data}, for $50\%$ swap percentage. The variable names in the column header, starting from `housing' to  `duration' are the mediating variables for the features listed in column `Feature'. The bold and italic font face are the maximal value along the rows, and the bold face (without italic) is the second highest value. According to Table~\ref{tab:double-swap-bank-data}, the `campaign', `duration', `loan', `age' and `default' are the features in the Bank dataset that frequently causes the higher values of mediation measures between the model prediction and other features. For example, the housing loan (i.e., `housing') feature is highly mediated by the `campaign' and `age' variable, and the minimal mediation is by the `job' variable.

\begin{tcolorbox}
As discussed above, the features such as `housing', `duration', and `loan'  show the maximal Controlled Direct Impact on the model prediction compared to features `job' and `education'. 

Moreover, the `loan', `duration', `default', and `age' features demonstrate a higher Total Natural Impact, due to mediating variables. 
\end{tcolorbox}
\subsection{\textbf{RQ2: Given these identified biased features, can we assert whether or not they are important to the model?}}

So far we have assessed how much each feature induces bias to the model decision. 
In this section, we aim to examine whether the features that we identified to be the potential source of bias to the machine learning model are also important feature to the model (i.e., feature importance). By feature importance, we only try to understand what features contribute to the predictive performance (component of model transparency in the FAT (Fairness, Accountability, and Transparency) model~\cite{shin2019role,bhattacharya2022applied}), while bias assessment we do not only assess the impact on the model outcomes but also tries to understand if the outcomes are without bias to some  sub-groups (a more fine-grained one). As part of our contributions, we wanted to show that treating the two concepts differently is essential to allow the ML practitioners and the domain experts to build a more responsible machine learning/ AI. 
We used the SHAP (SHapley Additive exPlanations) value to assess the feature's importance. SHAP value a state-of-the-art explainable method to help us explain the output of the machine learning model; refer to  Section~\ref{subsec:shap-value}. 
First, we compute the SHAP values for each instance of the sampled dataset. The similar instance of data used as inputs for our data swapping function was used in this step (i.e., $10\%, 30\%, 50\%, \text{ and } 70\%$ of the test set). Then, the importance of the feature is computed by taking the average of the absolute values of the SHAP values for all instances.

In the following, we present our results in the same order of the presentation for \textbf{RQ1} where each Subsection corresponds to the dataset used in the experiment, described in Section~\ref{subsec:RQ1} (i.e., Student Dataset, Cleveland Heart Dataset, COMPAS Dataset, and Bank Dataset).

\subsubsection{Feature importance for Student Dataset}

\begin{figure}[ht!]
\center
\includegraphics[width=\linewidth]{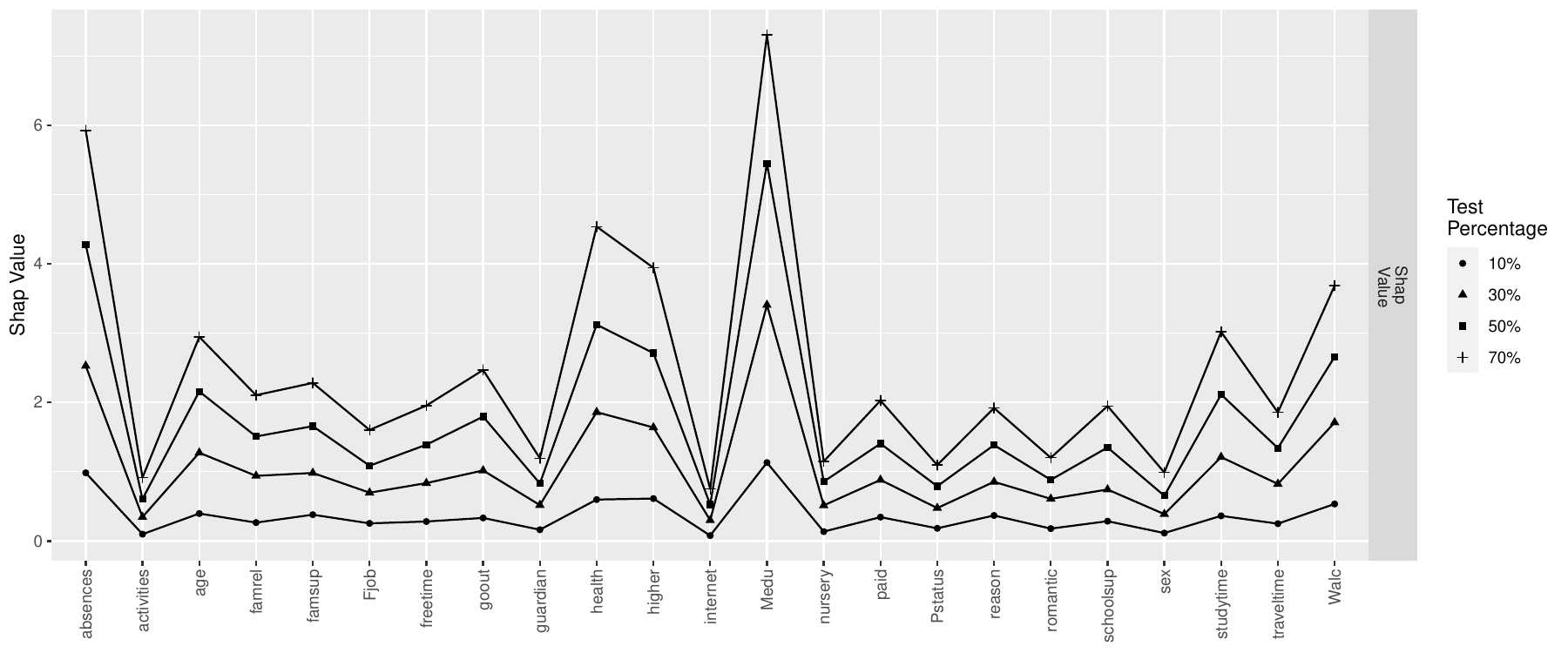}
\caption{The feature importance for the Student dataset, computed by running SHAP on the similar data points used as input by the swapping functions.}
\label{fig:shap-value-student-data}
\end{figure}
Figure~\ref{fig:shap-value-student-data} shows the average absolute  SHAP values for each feature in the Student Dataset, demonstrating the importance of a given feature in the model prediction. The higher SHAP value of a feature is an indication of the higher importance of that feature on the model outcome. As can be seen in Figure~\ref{fig:shap-value-student-data}, the SHAP values of each features increases with the number of data points used, similar to the results of swapping function, discussed in Section~\ref{subsec:RQ1}.

From Figure~\ref{fig:shap-value-student-data}, we can see that `Medu' consistently shows the highest SHAP values when tested on the different percentage of the test dataset. Furthermore, the ordering is generally consistent for the test percentages $30\%, 50\%, \text{and } 70\%$ and closely similar for $10\%$, ordering from maximum is: `Medu' , absences, health, higher, and Walc. On the other hand, the features with the minimal SHAP values are in the order: `activities', `guardian', `nursery' and `sex'

If we compare Figure~\ref{fig:shap-value-student-data} with Figure~\ref{fig:swap-student-data}, i.e., the results of SHAP values with the results of our swapping function, we can clearly see that the ordering of the SHAP values and the statistical distance are closely similar for the single feature swapping function shown in Figure~\ref{fig:single-swap-student-data}. To better understand the relationship between the SHAP values and our swapping functions, we introduced the concept of feature ranking stability in the following Subsection.

\paragraph{\textbf{Feature Ranking Stability:}}

We compute the stability of the rankings of features returned by SHAP and those returned by our swapping function to understand the consistency in the ranking order of the features from the two results. Specifically, we used Spearmans’ coefficient, following the idea from~\cite{kalousis2007stability}, to compare how the results of a ranking order are similar when different distance measures are used as follows:

\begin{equation}\label{eqn:stability}
    \text{(stability)} = 1 - \sum_{i} \frac{(r_{\Phi_i}-r_{\Im_i})^2 }{m\cdot (m^2-1)}
\end{equation}

In Equation (\ref{eqn:stability}), $m$ is the number of features, and $r_{\Phi}, r_{\Im}$ are the rankings by SHAP value and a given divergence measure, respectively. For example, SHAP Values and Hellinger distance, or SHAP Values and Jensen-divergence distance. $r_{\Phi_i}$ and $r_{\Im_i}$ also denote the ranks of feature $i$ in rankings $r_{\Phi}$ and $r_{\Im}$, respectively. Specifically, we computed the ranking stability between the SHAP value and the results of the single and double feature swapping function, reported in Figure~\ref{fig:swap-student-data} (i.e., Figures~\ref{fig:single-swap-student-data} and \ref{fig:double-swap-student-data}).

\begin{table}[h]
 \caption{Feature rankings Stability measure between the ranking for SHAP value and our swapping functions for \textbf{Student dataset}}\label{tab:stanility-student-ranking}
\centering
        \resizebox{\textwidth}{!}{\begin{tabular}{ l | c c}

 \rowcolor{gray!15} 
        \textbf{Distance Measure}&\textbf{SHAP vs Single}&\textbf{SHAP vs Double}\\ \midrule

Hellinger
 distance&0.897&0.862\\
Jensen-Shannon
 divergence&0.899&0.859\\
Total variation
 distance&0.903&0.866\\
Wasserstein
 distance&0.902&0.874\\
  \bottomrule
       \end{tabular}}
\end{table}


        \begin{center}
        \begin{longtable}{l| l r r c g}
        
        \caption{Comparing the Ranking of Feature by SHAP values vs. Ranking by double features swapping, for the Student dataset. $r_{\Im}$ is the ranking score based on the result of double feature swapping function, \textbf{$r_{\Phi}$} is the ranking based on SHAP value. The feature that is a potential source of bias (smaller $r_{\Im}$), yet is detected as less important to the model (large $r_{\Phi}$) is highlighted in \textbf{bold face}.}\label{logtab:compare-rank-student}. 
        \\
\hline
         \rowcolor{gray!15}
        \textbf{Distance Measure}&\textbf{Feature}&\textbf{$r_{\Im}$}&\textbf{$r_{\Phi}$}&$|r_{\Phi}-r_{\Im}|$&\textbf{Label}\\ \midrule
        \midrule

\multirow{20}{*}{\rotatebox[origin=c]{90}{\parbox[c]{5cm}{\centering \textbf{Hellinger distance}}}}&\textbf{sex}&2&21&19& $M_\Im L_\Phi$\\
&age&4&6&2& \\
&\textit{health}&3&3&0& $M_\Im M_\Phi$\\
&Pstatus&19&20&1& \\
&nursery&17&18&1& \\
&\textit{Medu}&1&1&0& $M_\Im M_\Phi$\\
&Fjob&15&16&1& \\
&schoolsup&18&14&4& \\
&\textit{absences}&22&2&20& $L_\Im M_\Phi$\\
&\textbf{activities}&7&22&15& $M_\Im L_\Phi$\\
&higher&21&4&17& \\
&traveltime&20&15&5& \\
&paid&12&11&1& \\
&guardian&23&19&4& \\
&Walc&16&5&11& \\
&freetime&9&12&3& \\
&famsup&5&9&4& \\
 
 \bottomrule
 
 \rowcolor{gray!15}
        \textbf{Distance Measure}&\textbf{Feature}&\textbf{$r_{\Im}$}&\textbf{$r_{\Phi}$}&$|r_{\Phi}-r_{\Im}|$& \textbf{Label}\\ \midrule
        \midrule
\multirow{20}{*}{\rotatebox[origin=c]{90}{\parbox[c]{5cm}{\centering \textbf{Jensen-Shannon
 divergence}}}}&\textbf{sex}&2&21&19& $M_\Im L_\Phi$\\
&age&4&6&2& \\
&\textit{health}&3&3&0& $M_\Im M_\Phi$\\
&Pstatus&18&20&2& \\
&nursery&17&18&1& \\
&\textit{Medu}&1&1&0& $M_\Im M_\Phi$\\
&Fjob&15&16&1& \\
&schoolsup&19&14&5& \\
&\textit{absences}&22&2&20& $L_\Im M_\Phi$\\
&\textbf{activities}&6&22&16& $M_\Im L_\Phi$\\
&higher&21&4&17& \\
&traveltime&20&15&5& \\
&paid&12&11&1& \\
&guardian&23&19&4& \\
&Walc&16&5&11& \\
&freetime&9&12&3& \\
&famsup&5&9&4& \\
 \bottomrule
 
 \rowcolor{gray!15}
        \textbf{Distance Measure}&\textbf{Feature}&\textbf{$r_{\Im}$}&\textbf{$r_{\Phi}$}&$|r_{\Phi}-r_{\Im}|$&\textbf{Label}\\ \midrule
        \midrule
\multirow{20}{*}{\rotatebox[origin=c]{90}{\parbox[c]{5cm}{\centering \textbf{Total variation
 distance}}}} &\textbf{sex}&3&21&18& $M_\Im L_\Phi$\\
&age&4&6&2& \\
&\textit{health}&2&3&1& $M_\Im M_\Phi$\\
&Pstatus&19&20&1& \\
&nursery&18&18&0& \\
&\textit{Medu}&1&1&0& $M_\Im M_\Phi$\\
&Fjob&16&16&0& \\
&schoolsup&17&14&3& \\
&\textit{absences}&22&2&20& $L_\Im M_\Phi$\\
&\textbf{activities}&5&22&17& $M_\Im L_\Phi$\\
&higher&20&4&16& \\
&traveltime&21&15&6& \\
&paid&11&11&0& \\
&guardian&23&19&4& \\
&Walc&15&5&10& \\
&freetime&8&12&4& \\
&famsup&6&9&3& \\
 
 \bottomrule
 
 \rowcolor{gray!15}
        \textbf{Distance Measure}&\textbf{Feature}&\textbf{$r_{\Im}$}&\textbf{$r_{\Phi}$}&$|r_{\Phi}-r_{\Im}|$&\textbf{Label}\\ \midrule
        \midrule
\multirow{20}{*}{\rotatebox[origin=c]{90}{\parbox[c]{5cm}{\centering \textbf{Wasserstein
 distance}}}} &\textbf{sex}&3&21&18& $M_\Im L_\Phi$\\
&age&4&6&2& \\
&\textit{health}&2&3&1& $M_\Im M_\Phi$\\
&Pstatus&18&20&2& \\
&nursery&19&18&1& \\
&\textit{Medu}&1&1&0& $M_\Im M_\Phi$\\
&Fjob&16&16&0& \\
&schoolsup&17&14&3& \\
&\textit{absences}&22&2&20& $L_\Im M_\Phi$\\
&\textbf{activities}&7&22&15& $M_\Im L_\Phi$\\
&higher&20&4&16& \\
&traveltime&21&15&6& \\
&paid&8&11&3& \\
&guardian&23&19&4& \\
&Walc&13&5&8& \\
&freetime&9&12&3& \\
&famsup&6&9&3& \\

\bottomrule

       \end{longtable}
       \end{center}

In Table~\ref{tab:stanility-student-ranking}, we report the feature ranking stability for the Student Dataset. The column `SHAP vs. Single' corresponds to the stability measure between the ranking generated from the distance measure after running single feature swapping function and the SHAP values ranking. On the other hand, the column named  `SHAP vs. Double' in Table~\ref{tab:stanility-student-ranking} is between the SHAP value and the Double features swapping function. Intuitively, A higher stability coefficient (close to 1) indicates that the ranking of the features was close to each other, and the rankings are consistent between the two measures. 

Looking at the stability measures in Table~\ref{tab:stanility-student-ranking}, we can say that the Controlled Direct Impact of the feature can also be used to explain the important features of the model. This is, however, not true when the mediating variables are considered when estimating the feature that is a potential source of bias. Table~\ref{logtab:compare-rank-student} shows the ranking of the features based on the results of double feature swapping function (i.e., $r_{\Im}$), and ranking of SHAP values (i.e., column $r_{\Phi}$), the magnitude of the difference (i.e., $|r_{\Phi}-r_{\Im}|$). The lower value of  $r_{\Im}$ indicates that the respective feature is estimated to be more likely the respective feature introduces bias to the model. For the feature importance, the lower the value of $r_{\Phi}$), the more important the feature is to the model. In column `label', we mark the features more bias inducing, and less important as $M_\Im M_\Phi$, more bias yet also more important as $M_\Im M_\Phi$, and less bias and more important as $L_\Im M_\Phi$. (i.e., the first letter indicating the bias-inducing capability and second letter for the importance on model predictive performance)

\begin{tcolorbox}

The features `sex', and `activities' are less important according to its SHAP value. Yet, they 
are also potential source of bias. Such kinds of features could be removed or constrained (given the domain context) from the Student dataset to improve the trained model's fairness. On the other hand, the feature `Medu' and `health' are both important and more potential source of bias, hence, require efficient bias mitigation to reduce the bias while not degrading the predictive performance. 
\end{tcolorbox}

\subsubsection{Feature importance for Cleveland Heart Dataset}

\begin{figure}[ht!]
\center
\includegraphics[width=\linewidth]{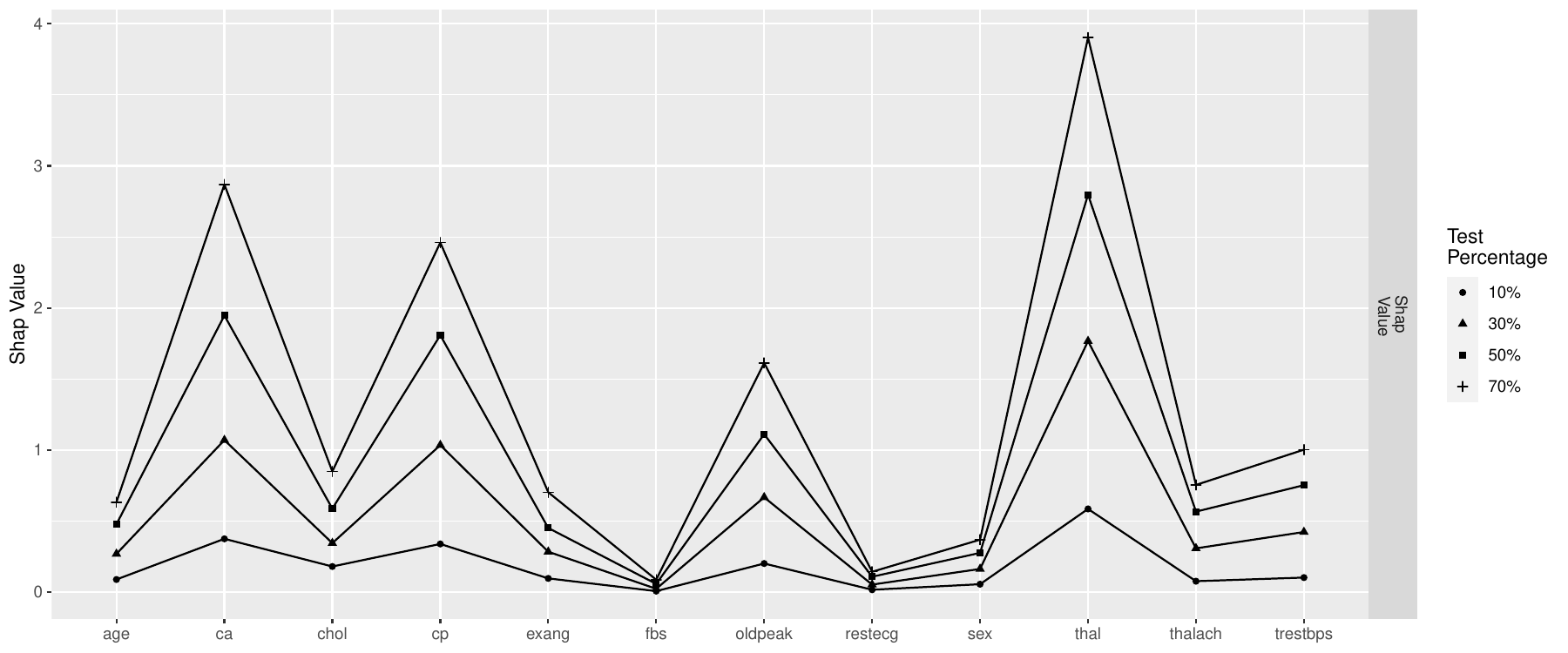}
\caption{The feature importance for the Cleveland Heart dataset, computed by running SHAP on the similar data points used as input by the swapping functions.}
\label{fig:shap-value-cleveran-data}
\end{figure}

Figure~\ref{fig:shap-value-cleveran-data} shows the summary of the average absolute SHAP values indicating the importance of each feature in the Cleveland Heart Dataset. The feature with the large SHAP value show its more important to the predictive model performance. In Figure~\ref{fig:shap-value-cleveran-data}, we see that feature can be arranged in the order of importance as `thal', `ca', `cp', `oldpeak', and `trestbps'. The close ordering was also observed for the single feature swapping function reported in Figure~\ref{fig:single-feature-cleveran-heart-data}.

\begin{table}[h]
 \caption{Feature rankings Stability measure between the ranking for SHAP value and our swapping functions for \textbf{Cleveran Heart dataset}}\label{tab:stanility-cleveran-ranking}
\centering
        \resizebox{\textwidth}{!}{\begin{tabular}{ l | c c}

 \rowcolor{gray!15} 
        \textbf{Distance Measure}&\textbf{SHAP vs Single}&\textbf{SHAP vs Double}\\ \midrule

Hellinger
 distance&0.985&0.879\\
Jensen-Shannon
 divergence&0.988&0.865\\
Total variation
 distance&0.99&0.846\\
Wasserstein
 distance&0.986&0.883\\
 
  \bottomrule
       \end{tabular}}
\end{table}

\begin{center}
        \begin{longtable}{c| l r r c g}
        
        \caption{Comparing the Ranking of Feature by SHAP values vs. Ranking by double features swapping, for the bank dataset. $r_{\Im}$ is the ranking score based on the double feature swapping function, \textbf{$r_{\Phi}$} is the ranking based on SHAP value, for \textbf{Cleveran Heart dataset}}\label{logtab:stability-ranking-cleveran-data}
        \\ \hline
         \rowcolor{gray!15}
        \textbf{Distance Measure}&\textbf{Feature}&\textbf{$r_{\Im}$}&\textbf{$r_{\Phi}$}&$|r_{\Phi}-r_{\Im}|$&\textbf{Label}\\ \midrule
        \midrule

 \multirow{12}{*}{\rotatebox[origin=c]{90}{\parbox[c]{3cm}{\centering \textbf{Hellinger distance}}}}&sex&3&4&1& $M_\Im M_\Phi$\\
&\textbf{age}&2&9&7& $M_\Im L_\Phi$\\
&thalach&6&3&3& \\
&ca&2&1&1& $M_\Im M_\Phi$\\
&thal&1&2&1& $M_\Im M_\Phi$\\
&exang&4&7&3& \\
&cp&7&6&1& \\
&trestbps&11&11&0& \\
&\textbf{restecg}&5&8&3& $M_\Im L_\Phi$\\
&fbs&9&10&1& \\
&oldpeak&12&5&7& $L_\Im M_\Phi$\\
&chol&10&12&2& \\
\bottomrule

\rowcolor{gray!15}
        \textbf{Distance Measure}&\textbf{Feature}&\textbf{$r_{\Im}$}&\textbf{$r_{\Phi}$}&$|r_{\Phi}-r_{\Im}|$&\textbf{Label}\\ \midrule
        \midrule
 \multirow{12}{*}{\rotatebox[origin=c]{90}{\parbox[c]{3cm}{\centering \textbf{Jensen-Shannon
 divergence}}}}&sex&3&4&1& $M_\Im M_\Phi$\\
&\textbf{age}&2&9&7& $M_\Im L_\Phi$\\
&thalach&6&3&3& \\
&ca&2&1&1& $M_\Im M_\Phi$\\
&thal&1&2&1& $M_\Im M_\Phi$\\
&exang&4&7&3& \\
&cp&7&6&1& \\
&trestbps&11&11&0& \\
&\textbf{restecg}&5&8&3& $M_\Im L_\Phi$\\
&fbs&9&10&1& \\
&oldpeak&12&5&7&$L_\Im M_\Phi$ \\
&chol&10&12&2& \\
\bottomrule
\rowcolor{gray!15}
        \textbf{Distance Measure}&\textbf{Feature}&\textbf{$r_{\Im}$}&\textbf{$r_{\Phi}$}&$|r_{\Phi}-r_{\Im}|$&\textbf{Label}\\ \midrule
        \midrule
 \multirow{12}{*}{\rotatebox[origin=c]{90}{\parbox[c]{3cm}{\centering \textbf{Total variation
 distance}}}}&sex&3&4&1& $M_\Im M_\Phi$\\
&\textbf{age}&2&9&7& $M_\Im L_\Phi$\\
&thalach&6&3&3& \\
&ca&2&1&1& $M_\Im M_\Phi$\\
&thal&1&2&1& $M_\Im M_\Phi$\\
&exang&4&7&3& \\
&cp&7&6&1& \\
&trestbps&11&11&0& \\
&\textbf{restecg}&5&8&3& $M_\Im L_\Phi$\\
&fbs&9&10&1& \\
&oldpeak&12&5&7& $L_\Im M_\Phi$\\
&chol&10&12&2& \\
\bottomrule
\rowcolor{gray!15}
        \textbf{Distance Measure}&\textbf{Feature}&\textbf{$r_{\Im}$}&\textbf{$r_{\Phi}$}&$|r_{\Phi}-r_{\Im}|$&\textbf{Label}\\ \midrule
        \midrule
\multirow{12}{*}{\rotatebox[origin=c]{90}{\parbox[c]{3cm}{\centering \textbf{Wasserstein
 distance}}}}&sex&3&4&1& $M_\Im M_\Phi$\\
&\textbf{age}&2&9&7& $M_\Im L_\Phi$\\
&thalach&7&3&4& \\
&ca&2&1&1& $M_\Im M_\Phi$\\
&thal&1&2&1& $M_\Im M_\Phi$\\
&exang&4&7&3& \\
&cp&8&6&2& \\
&trestbps&11&11&0& \\
&\textbf{restecg}&5&8&3& $M_\Im L_\Phi$\\
&fbs&9&10&1& \\
&oldpeak&12&5&7&$L_\Im M_\Phi$ \\
&chol&10&12&2& \\

\bottomrule

       \end{longtable}
       \end{center}
\paragraph{\textbf{Feature Ranking Stability:}} Table~\ref{tab:stanility-cleveran-ranking} report the feature ranking stability between the SHAP value and feature swapping functions using Equation~\ref{eqn:stability}. According to Table~\ref{tab:stanility-cleveran-ranking}, the ordering of the ranking of the features is closely similar between the SHAP value and the single feature swapping function. The closest similarity in the ordering was observed between the SHAP value and the Total variation distance measure for single figure swapping. For instance, the features detected to be more important to the model are similar detected to impose the controlled impact on the model prediction. 

When comparing the double features swapping functions with SHAP values, we observe that the ranking of some of the features is not close, implying a feature that might be important could also be biased, or vice-versa.
In Table~\ref{logtab:stability-ranking-cleveran-data} reports the ranking of the features by the double feature swapping function (i.e., $r_{\Im}$), and ranking of SHAP values (i.e., column $r_{\Phi}$), and their difference, and labels. Feature (s) detected as less important and more biased are highlighted in \textbf{bold face}.

\begin{tcolorbox}
The feature `age', and `restecg' are less important according to its SHAP value. Yet, they are potential source of bias to the model. These features could be removed from the Cleverant Heart dataset in order to improve the model's fairness. On the other hand, features `ca', and `thal' are potential bias inducing, yet important to model performance.
\end{tcolorbox}

\subsubsection{Feature importance for COMPAS Recidivism Dataset}

\begin{figure}[ht!]
\center
\includegraphics[width=\linewidth]{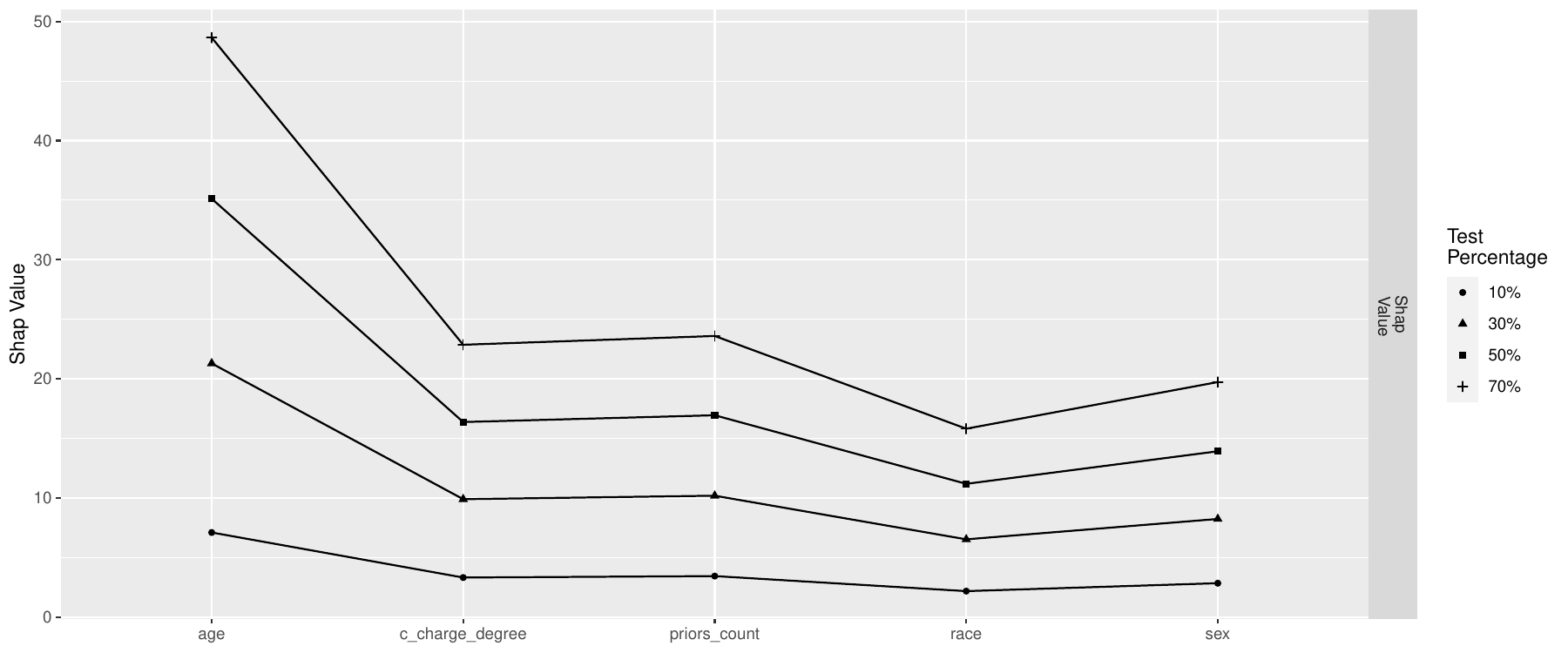}
\caption{The feature importance for the COMPAS Recidivism dataset, computed by running SHAP on the similar data points used as input by the swapping functions.}
\label{fig:shap-value-compas-data}
\end{figure}

In Figure~\ref{fig:shap-value-compas-data}, we reported the results of averaging the SHAP values for each features of COMPAS dataset.  The feature with the maximal SHAP value are more important. 
In Figure~\ref{fig:shap-value-compas-data}, the feature `age' is shown to be more important followed by the c\_charge\_degree, and `race' feature as less important.

\begin{table}[h]
 \caption{Feature rankings Stability measure between the ranking for SHAP value and our swapping functions for \textbf{COMPAS Recidivism dataset}}\label{tab:stanility-compas-ranking}
\centering
        \resizebox{\textwidth}{!}{\begin{tabular}{ l | c c}

 \rowcolor{gray!15} 
        \textbf{Distance Measure}&\textbf{SHAP vs Single}&\textbf{SHAP vs Double}\\ \midrule

Hellinger
 distance&0.933&0.983\\
Jensen-Shannon
 divergence&0.95&0.983\\
Total variation
 distance&0.883&0.967\\
Wasserstein
 distance&0.85&0.967\\
 
  \bottomrule
       \end{tabular}}
\end{table}

\begin{table}[h]
 \caption{Comparing the Ranking of Feature by SHAP values vs. Ranking by double features swapping, for the COMPAS Recidivism dataset. $r_{\Im}$ is the ranking score by the double feature swapping function, \textbf{$r_{\Phi}$} is the ranking based on the SHAP value.}\label{tab:stability-ranking-compas-data}
        \centering
        \begin{tabular}{c| l r r c g}

         \rowcolor{gray!15}
        \textbf{Distance Measure}&\textbf{Feature}&\textbf{$r_{\Im}$}&\textbf{$r_{\Phi}$}&$|r_{\Phi}-r_{\Im}|$&\textbf{Label}\\ \midrule
        \midrule

\multirow{5}{*}{\rotatebox[origin=c]{90}{\parbox[c]{1.5cm}{\centering \textbf{Hellinger distance}}}}&race&4&5&1& \\
&sex&5&4&1& \\
&age&1&1&0& $M_\Im M_\Phi$\\
&c\_charge\_degree&3&3&0& \\
&priors\_count&2&2&0& $M_\Im M_\Phi$\\
\bottomrule
 \rowcolor{gray!15}
        \textbf{Distance Measure}&\textbf{Feature}&\textbf{$r_{\Im}$}&\textbf{$r_{\Phi}$}&$|r_{\Phi}-r_{\Im}|$&\textbf{Label}\\ \midrule
        \midrule
\multirow{5}{*}{\rotatebox[origin=c]{90}{\parbox[c]{1.5cm}{\centering \textbf{Jensen-Shannon
 divergence}}}}&race&4&5&1& \\
&sex&5&4&1& \\
&age&1&1&0& $M_\Im M_\Phi$\\
&c\_charge\_degree&3&3&0& \\
&priors\_count&2&2&0& $M_\Im M_\Phi$\\
\bottomrule
 \rowcolor{gray!15}
        \textbf{Distance Measure}&\textbf{Feature}&\textbf{$r_{\Im}$}&\textbf{$r_{\Phi}$}&$|r_{\Phi}-r_{\Im}|$&\textbf{Label}\\ \midrule
        \midrule
\multirow{5}{*}{\rotatebox[origin=c]{90}{\parbox[c]{1.5cm}{\centering \textbf{Total variation
 distance}}}}&race&4&5&1& \\
&sex&5&4&1& \\
&age&1&1&0& $M_\Im M_\Phi$\\
&c\_charge\_degree&2&3&1& \\
&priors\_count&3&2&1& \\
\bottomrule
 \rowcolor{gray!15}
        \textbf{Distance Measure}&\textbf{Feature}&\textbf{$r_{\Im}$}&\textbf{$r_{\Phi}$}&$|r_{\Phi}-r_{\Im}|$&\textbf{Label}\\ \midrule
        \midrule
\multirow{5}{*}{\rotatebox[origin=c]{90}{\parbox[c]{1.5cm}{\centering \textbf{Wasserstein
 distance}}}}&race&4&5&1& \\
&sex&5&4&1& \\
&age&1&1&0& $M_\Im M_\Phi$\\
&c\_charge\_degree&2&3&1& \\
&priors\_count&3&2&1& \\

\bottomrule

       \end{tabular}
       
       \end{table}
Comparing the SHAP values with the swapping functions, we compute the feature ranking stability using Equation~(\ref{eqn:stability}) to examine the consistency in the two ranking orders. The results of Equation~(\ref{eqn:stability}) is shown in Table~\ref{tab:stanility-compas-ranking}. The divergence measures closely agree with the SHAP value ranking and Hellinger distance, Jensen-Shannon divergence, and Total variation distance, when measured for both single and double feature swapping function results. This results indicate the features detected as bias inducing are also most important to the model predictive performance double. 

Table~\ref{tab:stability-ranking-compas-data} details the ranking order of SHAP value (column $r_{\Phi}$) and the ranking order for double feature swapping function (i.e., $r_{\Im}$), and the absolute difference as $|r_{\Phi}-r_{\Im}|$. The results shows a close rankings between the double feature swapping and SHAP values, hence, the choice of what features are less important feature (s) and yet are more biased are not highlighted and the intepretation is left to the domain expert. For instance, according to Table~\ref{tab:stability-ranking-compas-data}, `race' consistently shows to be more bias-inducing than `sex' feature and yet less important between three of the  four distance measures vs. SHAP values.
\begin{tcolorbox}
The features in COMPAS data are either both potential sources of bias and also more important to the model prediction, or vice-versa. An interpreted by the domain expert is essential to help in proposing an efficient bias mitigation method to remove bias in the highly biased features (e.g., `age'), while preserving the model predictive performance. 
\end{tcolorbox}

\subsubsection{Feature importance for Bank Dataset}
Finally, in this Subsection, we examined the feature's importance and relation to the bank dataset's potential bias.
\begin{figure}[ht!]
\center
\includegraphics[width=\linewidth]{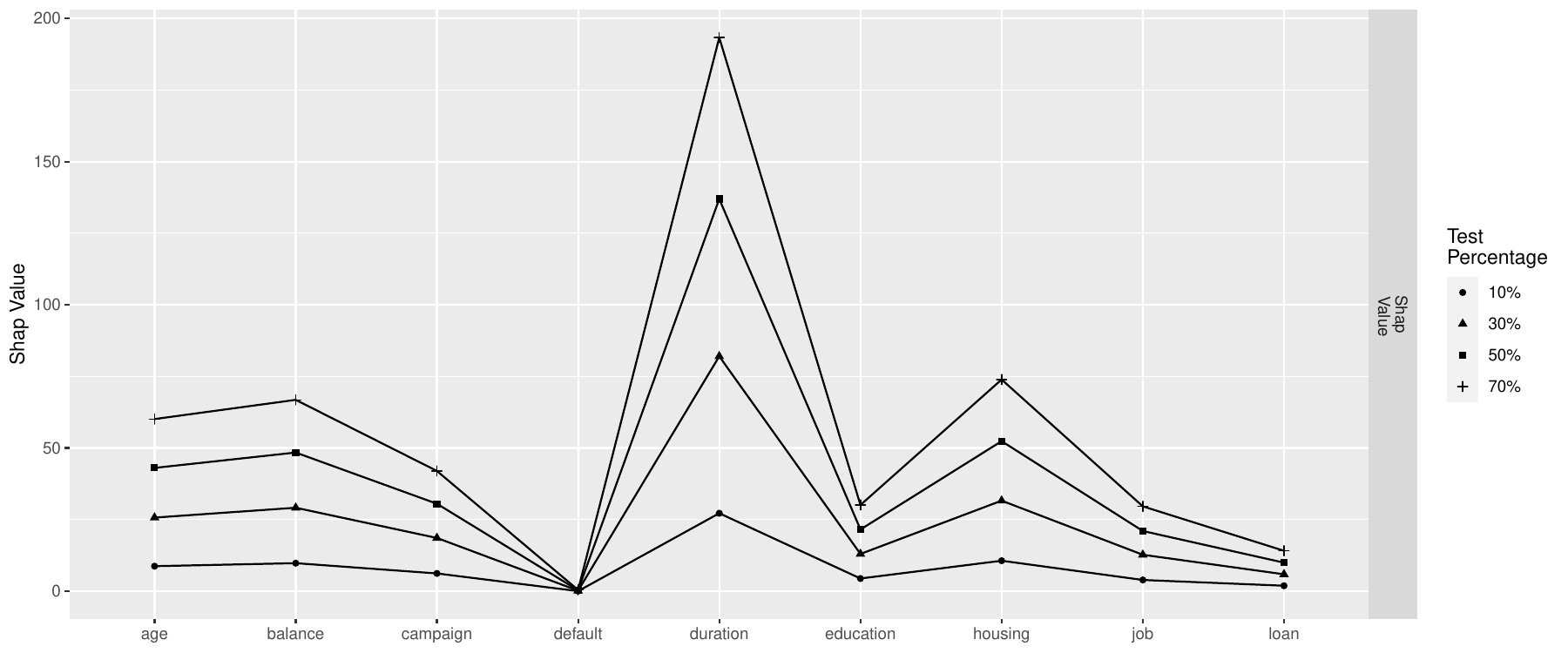}
\caption{The feature importance for the Bank dataset, computed by running SHAP on the similar data points used as input by the swapping functions.}
\label{fig:methodology}
\end{figure}
Figure~\ref{fig:shap-value-compas-data} report the average absolute SHAP values for the features in Bank dataset. The features with higher SHAP values are more important to the model prediction. From  Figure~\ref{fig:shap-value-compas-data}, we can see that `duration' consistently shows the highest SHAP value; hence is a more important feature.

\begin{table}[h]
 \caption{Feature rankings Stability measure between the ranking for SHAP value and our swapping functions for \textbf{Bank dataset}}\label{tab:stability-bank-data}
\centering
        \resizebox{\textwidth}{!}{\begin{tabular}{ l | c c}

 \rowcolor{gray!15} 
        \textbf{Distance Measure}&\textbf{SHAP vs Single}&\textbf{SHAP vs Double}\\ \midrule

Hellinger
 distance&0.917&0.836\\
Jensen-Shannon
 divergence&0.925&0.836\\
Total variation
 distance&0.892&0.847\\
Wasserstein
 distance&0.853&0.822\\
 
  \bottomrule
       \end{tabular}}
\end{table}


        \begin{center}
        \begin{longtable}{c| l r r c g }
        
        \caption{Comparing the Ranking of Feature by SHAP values vs. Ranking by double features swapping, for the bank dataset. $r_{\Im}$ is the ranking score by the double feature swapping function, \textbf{$r_{\Phi}$} is the ranking of the SHAP value}\label{longtab:stability-ranking-bank-data}.
        \\ \hline
         \rowcolor{gray!15}
        \textbf{Distance Measure}&\textbf{Feature}&\textbf{$r_{\Im}$}&\textbf{$r_{\Phi}$}&$|r_{\Phi}-r_{\Im}|$& \textbf{Label}\\ \midrule
        \midrule


 \multirow{9}{*}{\rotatebox[origin=c]{90}{\parbox[c]{3cm}{\centering \textbf{Hellinger distance}}}} &age&4&4&0& \\
&education&9&6&3& \\
&job&8&7&1& \\
&\textbf{loan}&1&8&7& $M_\Im L_\Phi$\\
&balance&6&3&3& \\
&\textit{housing}&5&2&3& $L_\Im M_\Phi$\\
&\textit{duration}&2&1&1& $M_\Im M_\Phi$\\
&campaign&7&5&2& \\
&\textbf{default}&3&9&6& $M_\Im L_\Phi$\\
\bottomrule

\rowcolor{gray!15}
        \textbf{Distance Measure}&\textbf{Feature}&\textbf{$r_{\Im}$}&\textbf{$r_{\Phi}$}&$|r_{\Phi}-r_{\Im}|$&  \textbf{Label}\\ \midrule
        \midrule

 \multirow{9}{*}{\rotatebox[origin=c]{90}{\parbox[c]{3cm}{\centering \textbf{Jensen-Shannon
 divergence}}}} &age&4&4&0& \\
&education&9&6&3& \\
&job&8&7&1& \\
&\textbf{loan}&1&8&7& $M_\Im L_\Phi$\\
&balance&6&3&3& \\
&\textit{housing}&5&2&3& $L_\Im M_\Phi$\\
&\textit{duration}&2&1&1&$M_\Im M_\Phi$ \\
&campaign&7&5&2& \\
&\textbf{default}&3&9&6& $M_\Im L_\Phi$\\

\bottomrule

\rowcolor{gray!15}
        \textbf{Distance Measure}&\textbf{Feature}&\textbf{$r_{\Im}$}&\textbf{$r_{\Phi}$}&$|r_{\Phi}-r_{\Im}|$& \textbf{Label}\\ \midrule
        \midrule

\multirow{9}{*}{\rotatebox[origin=c]{90}{\parbox[c]{3cm}{\centering \textbf{Total variation
 distance}}}} &age&4&4&0& \\
&education&8&6&2& \\
&job&9&7&2& \\
&\textbf{loan}&2&8&6& $M_\Im L_\Phi$\\
&balance&5&3&2& \\
&\textit{housing}&7&2&5& $L_\Im M_\Phi$\\
&\textit{duration}&1&1&0& $M_\Im M_\Phi$\\
&campaign&6&5&1& \\
&\textbf{default}&3&9&6& $M_\Im L_\Phi$\\
\bottomrule
\rowcolor{gray!15}
        \textbf{Distance Measure}&\textbf{Feature}&\textbf{$r_{\Im}$}&\textbf{$r_{\Phi}$}&$|r_{\Phi}-r_{\Im}|$ &\textbf{Label}\\ \midrule
        \midrule
\multirow{9}{*}{\rotatebox[origin=c]{90}{\parbox[c]{3cm}{\centering \textbf{Wasserstein
 distance}}}} &age&4&4&0& \\
&education&8&6&2& \\
&job&9&7&2& \\
&\textbf{loan}&1&8&7& $M_\Im L_\Phi$\\
&balance&6&3&3& \\
&\textit{housing}&7&2&5& $L_\Im M_\Phi$\\
&\textit{duration}&2&1&1& $M_\Im M_\Phi$\\
&campaign&5&5&0& \\
&\textbf{default}&3&9&6& $M_\Im L_\Phi$\\

\bottomrule

       \end{longtable}
       \end{center}

Next, we compare the relation in the ranking between the feature importance and bias measures using the swapping functions, and report the feature ranking stability (using Equation~(\ref{eqn:stability})) in Table~\ref{tab:stability-bank-data}. Like the Student and Cleveran-Heart datasets, the results show higher stability between the SHAP values and single feature swapping function than double feature swapping, especially for the Hellinger
 distance and Jensen-Shannon divergence, for single feature swapping. We detailed the summary ranking of the feature importance, and bias detected using double features swapping in Table~\ref{longtab:stability-ranking-bank-data}. The results show that features like default, and loan are more biased and have minimal importance to the model prediction performance.

\begin{tcolorbox}
The features such as `loan', and `default' are more biased yet they are less important to the model prediction performance. These features could be removed from the dataset to improve the model's fairness. On the other hand, feature `housing', and `balance' are more important and less biased. `duration' and `age' features are both important and also bias inducing, therefore require a better bias mitigation technique. The `default' is the least important to the model, and neither is a potential source of bias to the model.
\end{tcolorbox}




\section{Evaluation}\label{sec:evaluation}

This section aims to validate our conclusion from the previous Section, by demonstrating that, taking care of the identified biased-inducing and important features identified, can potentially improve the model fairness as well as the model predictive performance. 
To this end, we evaluated the features identified using our proposed framework following different combinations of scenarios between the bias-inducing and important features. The question we are asking here is which of the features if removed can potentially reduce the bias affecting the final decision of the model while maintaining/ improving the model performance. The scenarios evaluated include removing the most bias-inducing feature ($M_\Im$) yet also more important to model prediction ($M_\Phi$), i.e., the combination is labelled as $M_\Im M_\Phi$, the most biased yet less important feature ($M_\Im L_\Phi$), and finally, the less biased yet more important ($L_\Im M_\Phi$). We have also evaluated the scenarios when the mitigation techniques are used. We chose three relevant bias mitigation techniques proposed in the literature, including FairSMOTE~\cite{chakraborty2021bias}, Reweighing~\cite{kamiran2012data} and Linear-regression based Training Data Debugging (LTDD)~\cite{LRTDD:2022}, to verify how the model performance before and after removing the features.  All these mitigation techniques aim
at reducing bias in training data, e.g., data imbalance (FairSMOTE), statistical dependencies, i.e., sensitive and non-sensitive features (LRTDD), or classes (Reweighing). 

We used a scoring function to score the best scenarios basing on the combination of performance metrics and fairness metrics of the model. The scoring function uses the additive ranking approach~\cite{tofallis2014add} by adding all model performance measures (i.e., Accuracy, Precision, Recall, F1-score) and subtracting the fairness metrics, defined as follows:
\begin{equation}\label{eqn:tscore}
    \text{T-Score} = \mathbb{E} \left[\sum v_{i}- \sum u_{i} 
    \right]
\end{equation}

The scoring function above is based on the idea that the performance metrics (the first parameter $v_i$) where the more the better are added up, while the fairness measure (second parameter ($u_i$)) which is the smaller the value the better are subtracted. 
The highest value of $\text{T-Score}$ indicate a best choice and the corresponding scenario is recommended among the rest. Moreover, we used the Wilcoxon rank sum test~\cite{de2011stochastic} to compare the distributions between the different scenarios before and after the mitigation techniques are used to verify the statistically significant. Specifically, we assessed the significance of the two distributions using the standard
criterion of $\rho-$value $< 0.05$. 

Next, we used Cliff’s delta $\delta$~\cite{romano2006exploring} to quantify the amount of difference between the two sets of results (i.e., effect size), and categorize the effect size into three levels (Win, Tied, and Loss), as follows. 
The value of $|\delta| < 0.147$ indicate a negligible difference between the two sets; a $|\delta| > 0.147 \text{ and } < 0.330$, shows a relatively small difference; the $|\delta| >  0.330$ indicate a medium to the large difference between the two sets of results.
The comparison include comparing the model performance and fairness measures of default scenarios (before removing the feature) vs. after removing the feature. Specifically, we used ``W/T/L''~\cite{liu2018connecting,LRTDD:2022}  to label the comparison of two sets of results. Where ``W'' means the current scenario wins, whenever the corresponding $\rho$-value $< 0.05$, and $\delta > 0.147$.  ``L'' indicating loses, whenever the $\rho$-value $< 0.05$, and $\sigma < -0.147$. Otherwise, we used ``T'', to mark the two sets of results as tied.

\subsection{Student Dataset}\label{subsec:eval-student}

\begin{table}[h]
\Large 
\caption{
Comparing the model performance and fairness, before and after removing the potential biased inducing features (PBF), for \textbf{Student dataset}}
     \label{tab:student-eval-before}
      \begin{subtable}[h]{\textwidth}

        \caption{Model performance and fairness when no mitigation techniques are used}
       \label{tab:student-eval-before-1}
       
        \centering
        \resizebox{\textwidth}{!}{\begin{tabular}{l c| r | r r rr|rrr r r |g c}
        

        \textbf{Feature}&\textbf{Label}&\textbf{PBF}&\textbf{ACC}&\textbf{PRE}&\textbf{Recall}&\textbf{F1}&\textbf{F-alarm}&\textbf{AOD}&\textbf{SPD}&\textbf{DIR}&\textbf{FPR$_D$}&\textbf{T-Score}&\textbf{RANK}\\

        \midrule
        \midrule
        
        \multirow{5}{*}{\textbf{age}}&Default&-&68.41&60.66&50.02&54.03&0.198&0.1206&0.2985&1.0036&0.0801&231.42& 2\\
        \cline{2-13}
        &\multirow{2}{*}{$M_\Im M_\Phi$}&Medu&65.63&56.94&45.42&49.66&0.2092&0.1287&0.2705&0.9755&0.0853&215.98& \multirow{2}{*}{4}\\
        &&health&67.55&59.52&47.56&51.83&0.1946&\cellcolor{blue!25}0.1105&0.2893&0.8523&0.0606&224.95& \\
        \cline{2-13}
        &\multirow{2}{*}{\textbf{$M_\Im L_\Phi$}}&\textbf{activities}&\cellcolor{blue!25}68.98&\cellcolor{blue!25}61.58&\cellcolor{blue!25}51.67&\cellcolor{blue!25}55.35&0.196&0.1324&0.2952&1.01&\cellcolor{blue!25}0.0761&\textbf{235.87}&\multirow{2}{*}{1} \\
        &&sex&68.79&61.52&49.96&54.33&\cellcolor{blue!25}0.1909&0.1301&\cellcolor{blue!25}0.254&1.1574&0.0855&232.78& \\
         \cline{2-13}
        &$L_\Im M_\Phi$&absences&68.12&60.72&49.89&53.74&0.1979&0.105&0.2584&\cellcolor{blue!25}0.6626&0.1049&231.14& 3\\
        \hline

        \multirow{5}{*}{\textbf{health}}&Default&-&68.41&60.66&50.02&54.03&0.198&0.0134&0.1904&0.5456&0.1626&232.01& 2\\
        \cline{2-13}

        &$M_\Im M_\Phi$&Medu&65.63&56.94&45.42&49.66&0.2092&0.0346&0.2427&0.4595&0.1945&216.51&4 \\
        \cline{2-13}
        &\multirow{2}{*}{\textbf{$M_\Im L_\Phi$}}&\textbf{activities}&\cellcolor{blue!25}68.98&\cellcolor{blue!25}61.58&\cellcolor{blue!25}51.67&\cellcolor{blue!25}55.35&0.196&0.0267&0.1902&0.5525&0.1462&\textbf{236.47}& \multirow{2}{*}{1}\\
        &&sex&68.79&61.52&49.96&54.33&\cellcolor{blue!25}0.1909&0.0117&\cellcolor{blue!25}0.1868&0.5435&0.1526&233.51& \\
        
        \cline{2-13}
        &$L_\Im M_\Phi$&absences&68.12&60.72&49.89&53.74&0.1979&\cellcolor{blue!25}0.0097&0.2042&0.5176&0.1761&231.36& 3\\
        
        \hline
        
        \multirow{4}{*}{\textbf{famsup}}&Default&-&68.41&60.66&50.02&54.03&0.198&0.021&0.0789&0.8509&0.0858&231.89& 2\\
        \cline{2-13}

        &\multirow{2}{*}{$M_\Im M_\Phi$}&Medu&65.63&56.94&45.42&49.66&0.2092&0.0191&0.0885&0.8941&\cellcolor{blue!25}0.0744&216.36& \multirow{2}{*}{4}\\
        &&health&67.55&59.52&47.56&51.83&0.1946&\cellcolor{blue!25}0.0096&0.0769&\cellcolor{blue!25}0.8355&0.085&225.26& \\

        \cline{2-13}
        &\multirow{2}{*}{\textbf{$M_\Im L_\Phi$}}&\textbf{activities}&\cellcolor{blue!25}68.98&\cellcolor{blue!25}61.58&\cellcolor{blue!25}51.67&\cellcolor{blue!25}55.35&0.196&0.0269&\cellcolor{blue!25}0.074&0.8615&0.0886&\textbf{236.33}& \multirow{2}{*}{1}\\
        &&sex&68.79&61.52&49.96&54.33&\cellcolor{blue!25}0.1909&0.0295&0.0835&0.8574&0.0907&233.35& \\
        
        \cline{2-13}
        &$L_\Im M_\Phi$&absences&68.12&60.72&49.89&53.74&0.1979&0.0336&\cellcolor{blue!25}0.074&0.8713&0.083&231.21& 3\\

        \hline
        
        \multirow{4}{*}{\textbf{schoolsup}}&Default&-&68.41&60.66&50.02&54.03&0.198&0.1905&0.3136&\cellcolor{blue!25}0.1893&\cellcolor{blue!25}0.1566&232.07& 2\\
        \cline{2-13}

        &\multirow{2}{*}{$M_\Im M_\Phi$}&Medu&65.63&56.94&45.42&49.66&0.2092&\cellcolor{blue!25}0.15&\cellcolor{blue!25}0.2924&3.2041&0.1862&213.61& \multirow{2}{*}{4}\\
        &&health&67.55&59.52&47.56&51.83&0.1946&0.1616&0.3087&0.3147&0.1881&225.29& \\
        \cline{2-13}

        &\multirow{2}{*}{\textbf{$M_\Im L_\Phi$}}&\textbf{activities}&\cellcolor{blue!25}68.98&\cellcolor{blue!25}61.58&\cellcolor{blue!25}51.67&\cellcolor{blue!25}55.35&0.196&0.1958&0.3252&0.2313&0.1634&\textbf{236.47}& \multirow{2}{*}{1}\\
        &&sex&68.79&61.52&49.96&54.33&\cellcolor{blue!25}0.1909&0.1899&0.3101&0.2244&0.1571&233.53& \\

        \cline{2-13}
        &$L_\Im M_\Phi$&absences&68.12&60.72&49.89&53.74&0.1979&0.1885&0.3127&0.1964&0.1591&231.42& 3\\

        \hline
        
        \multirow{4}{*}{\textbf{sex}}&Default&-&68.41&60.66&50.02&54.03&0.198&0.0044&\cellcolor{blue!25}0.1274&\cellcolor{blue!25}1.5162&0.0793&231.19& 2\\
        \cline{2-13}

        &\multirow{2}{*}{$M_\Im M_\Phi$}&Medu&65.63&56.94&45.42&49.66&0.2092&0.017&0.135&1.7515&0.1321&215.41& \multirow{2}{*}{4}\\
        &&health&67.55&59.52&47.56&51.83&\cellcolor{blue!25}0.1946&0.026&0.1354&1.6464&\cellcolor{blue!25}0.0637&224.39& \\
        \cline{2-13}

        &\textbf{$M_\Im L_\Phi$}&\textbf{activities}&\cellcolor{blue!25}68.98&\cellcolor{blue!25}61.58&\cellcolor{blue!25}51.67&\cellcolor{blue!25}55.35&0.196&0.0121&0.134&1.5887&0.0763&\textbf{235.57}&1 \\

        \cline{2-13}
        &$L_\Im M_\Phi$&absences&68.12&60.72&49.89&53.74&0.1979&\cellcolor{blue!25}0.0015&0.1312&1.6123&0.0855&230.44& 3\\

        \hline
        \multirow{4}{*}{\textbf{studytime}}&Default&-&68.41&60.66&50.02&54.03&0.198&0.0116&0.2784&0.5522&0.2349&231.84&2 \\
        \cline{3-13}

        &\multirow{2}{*}{$M_\Im M_\Phi$}&Medu&65.63&56.94&45.42&49.66&0.2092&0.0581&0.3046&\cellcolor{blue!25}0.4746&0.2206&216.38&\multirow{2}{*}{4} \\
        &&health&67.55&59.52&47.56&51.83&0.1946&\cellcolor{blue!25}0.0002&0.2812&0.5026&0.2387&225.24& \\

        &\textbf{$M_\Im L_\Phi$}&\textbf{activities}&\cellcolor{blue!25}68.98&\cellcolor{blue!25}61.58&\cellcolor{blue!25}51.67&\cellcolor{blue!25}55.35&0.196&0.0146&\cellcolor{blue!25}0.279&0.5536&\cellcolor{blue!25}0.2065&\textbf{236.33}& 1\\
        &&sex&68.79&61.52&49.96&54.33&\cellcolor{blue!25}0.1909&0.008&0.2813&0.5433&0.2292&233.35& \\

        \cline{3-13}
        &$L_\Im M_\Phi$&absences&68.12&60.72&49.89&53.74&0.1979&0.0273&\cellcolor{blue!25}0.279&0.5517&0.2405&231.17& 3\\
\bottomrule

\end{tabular}}
\vspace{10pt}
\end{subtable}

    \hfill

 \begin{subtable}[h]{\textwidth}
   \small

        \caption{Model performance and fairness measure comparing the default scenario against other scenarios. 
        }
       \label{tab:student-eval-before-2}
       
       \centering
        \resizebox{0.95\textwidth}{!}{\begin{tabular}{l| c | c c c|cc c}
        

        \multirow{2}{*}{\textbf{Label}}&\multirow{2}{*}{\textbf{PBF}}&\multicolumn{3}{c|}{\textbf{Performance}}&\multicolumn{3}{c|}{\textbf{Fairness}}\\
        

        &&\textbf{Win}&\textbf{Tie}&\textbf{Loss}&\textbf{Win}&\textbf{Tie}&\textbf{Loss}\\

        \midrule
        \midrule
        
       \multirow{2}{*}{$M_\Im M_\Phi$}&Medu&0&22&0&0&12&10\\
       &health&0&22&0&0&11&11\\
        \hline
        \multirow{2}{*}{$M_\Im L_\Phi$}&activities&0&22&0&0&21&1\\
        
        &sex&0&22&0&0&14&8\\

        \hline
        $L_\Im M_\Phi$&absences&0&22&0&0&16&6\\
    
\bottomrule

\end{tabular}}

       \vspace{5pt}
    \end{subtable}

   \end{table}


\begin{table}[h]
\Large 
\caption{
Model performance and fairness before and after applying the bias mitigation techniques, on \textbf{Student dataset}
}  
     \label{tab:student-eval-mitigation}

        \begin{subtable}[h]{\textwidth}
        \caption{Model performance and fairness when bias mitigation technique is used. 
        }
       \label{tab:student-eval-mitigation-1}
        \centering
        \resizebox{0.95\textwidth}{!}{
    \begin{tabular}{l l c| r | r r rr|rrr r r |g c}
        

        \textbf{Feature}&\textbf{Algo}&\textbf{Label}&\textbf{PBF}&\textbf{ACC}&\textbf{PRE}&\textbf{Recall}&\textbf{F1}&\textbf{F-alarm}&\textbf{AOD}&\textbf{SPD}&\textbf{DIR}&\textbf{FPR$_D$}&\textbf{T-Score}&\textbf{RANK}\\

\midrule
\midrule

        \multirow{10}{*}{\textbf{\textbf{age}}}&\multirow{5}{*}{\textbf{FairSMote}}&\textbf{Default}&&\cellcolor{blue!25}63.02&\cellcolor{blue!25}49.37&\cellcolor{blue!25}54.35&\cellcolor{blue!25}51.19&\cellcolor{blue!25}0.3344&0.0874&0.365&\cellcolor{blue!25}1.4687&0.275&\textbf{215.40}& 1\\
        \cline{3-14}
        &&\multirow{2}{*}{$M_\Im M_\Phi$}&Medu&58.99&45.5&50.83&47.58&0.386&0.0651&0.2965&2.9444&0.2159&198.99& \multirow{2}{*}{4}\\
        &&&health&61.96&47.11&51.46&48.91&0.3427&0.111&0.3723&1.9087&0.2589&206.45&\\
        \cline{3-14}
        &&\multirow{2}{*}{$M_\Im L_\Phi$}&activities&61.39&46.16&49.61&47.6&0.3433&\cellcolor{blue!25}0.0335&\cellcolor{blue!25}0.2503&2.1503&\cellcolor{blue!25}0.1994&201.78& \multirow{2}{*}{3}\\
        &&&sex&62.35&48.54&53.74&50.78&0.3467&0.1513&0.3433&2.4977&0.2159&211.86&\\

        \cline{3-14}
        &&$L_\Im M_\Phi$&absences&62.25&48.82&53.8&50.69&0.344&0.1362&0.3528&2.2349&0.2366&212.26& 2\\

        \cline{2-14}
        \cline{2-15}


        &\multirow{5}{*}{\textbf{LRTDD}}&Default&-&66&56.66&41.76&47.71&0.1923&0.0587&0.2077&0.7189&0.016&210.94&3 \\
        \cline{3-14}
        &&\multirow{2}{*}{$M_\Im M_\Phi$}&Medu&63.81&54.15&37.98&43.85&0.1978&0.0693&0.1504&0.7851&0.0579&198.53& \multirow{2}{*}{4}\\
        &&&health&66.1&56.22&42.13&47.71&0.197&\cellcolor{blue!25}0.0411&0.2132&0.8237&0.0298&210.86&\\
        \cline{3-14}
        
        &&\multirow{2}{*}{\textbf{$M_\Im L_\Phi$}}&activities&66.1&56.84&42.08&48.04&0.1925&0.0593&0.2072&\cellcolor{blue!25}0.7107&0.0166&211.87& \multirow{2}{*}{1}\\

        &&&\textbf{sex}&66.49&56.55&\cellcolor{blue!25}45.14&\cellcolor{blue!25}49.69&0.208&0.0482&\cellcolor{blue!25}0.176&0.8977&\cellcolor{blue!25}0.0136&\textbf{216.53}&\\

        \cline{3-14}
        &&$L_\Im M_\Phi$&absences&\cellcolor{blue!25}67.07&\cellcolor{blue!25}58.42&42.67&48.71&\cellcolor{blue!25}0.1806&0.0739&0.1976&0.8831&0.0314&215.5& 2\\

        \hline
        \hline

        \multirow{10}{*}{\textbf{health}}&\multirow{5}{*}{\textbf{FairSMote}}&Default&-&63.39&49.71&61.48&54.69&0.3765&0.015&0.1122&\cellcolor{blue!25}0.8611&0.0754&227.83& 3\\
        \cline{3-14}

        &&$M_\Im M_\Phi$&Medu&62.34&49.27&61.3&54.24&0.3835&0.0503&0.1227&0.9251&0.0869&225.58& 4\\

        \cline{3-14}
        &&\multirow{2}{*}{\textbf{$M_\Im L_\Phi$}}&activities&64.16&50.35&60.47&54.67&0.3585&0.0115&0.1208&0.8666&0.0809&228.21& \multirow{2}{*}{1}\\
        &&&\textbf{sex}&\cellcolor{blue!25}65.22&\cellcolor{blue!25}52.00&\cellcolor{blue!25}62.26&\cellcolor{blue!25}56.26&\cellcolor{blue!25}0.3475&\cellcolor{blue!25}0.0084&\cellcolor{blue!25}0.0858&0.8819&\cellcolor{blue!25}0.0426&\textbf{234.37}&\\

        \cline{3-14}
        &&$L_\Im M_\Phi$&absences&63.88&50.67&61.71&55.21&0.3644&0.0383&0.1042&0.8758&0.0811&230.01& 2\\
        \cline{2-15}

        
       &\multirow{5}{*}{\textbf{LRTDD}}&Default&-&63.5&51.55&61.8&55.5&0.3546&0.0328&0.068&\cellcolor{blue!25}0.9184&0.0467&230.93& 2\\
        \cline{3-14}

        &&$M_\Im M_\Phi$&Medu&63.42&\cellcolor{blue!25}52.01&43.31&46.69&\cellcolor{blue!25}0.2427&0.0372&0.0758&0.9601&0.0465&204.07& 4\\
        \cline{3-14}

        &&\multirow{2}{*}{$M_\Im L_\Phi$}&activities&63.11&51.13&60.42&54.69&0.3532&\cellcolor{blue!25}0.0233&\cellcolor{blue!25}0.0649&0.9284&0.0378&227.94& \multirow{2}{*}{3}\\
        &&&sex&\cellcolor{blue!25}63.59&51.65&61.46&55.43&0.3517&0.0409&0.0651&0.9233&0.0528&230.7&\\
        
        \cline{3-14}
        &&\textbf{$L_\Im M_\Phi$}&\textbf{absences}&63.21&51.33&\cellcolor{blue!25}64.4&\cellcolor{blue!25}56.42&0.3742&0.0465&0.169&1.1936&\cellcolor{blue!25}0.0179&\textbf{233.56}& 1\\

        \hline
        \hline
        

        \multirow{8}{*}{\textbf{famsup}}&\multirow{5}{*}{\textbf{FairSMote}}&Default&-&65.41&\cellcolor{blue!25}52.22&62.82&56.64&0.3484&\cellcolor{blue!25}0.0085&0.0649&0.9944&\cellcolor{blue!25}0.0302&235.64& 2\\
        \cline{3-14}

        &&\multirow{2}{*}{$M_\Im M_\Phi$}&Medu&62.25&49.3&59.71&53.56&0.3772&0.0099&0.1098&0.9621&0.0347&223.33& \multirow{2}{*}{4}\\
        &&&health&66.18&52.87&62.19&57&0.3372&0.0098&\cellcolor{blue!25}0.0631&0.9732&0.0372&236.82&\\

        \cline{3-14}

        &&\multirow{2}{*}{$M_\Im L_\Phi$}&activities&65.32&51.99&59.21&54.99&\cellcolor{blue!25}0.3326&0.0198&0.0841&0.9423&0.0566&230.07& \multirow{2}{*}{3}\\
        &&&sex&64.83&51.73&60.46&55.27&0.3443&0.0136&0.0894&0.927&0.0616&230.85&\\
        
        \cline{3-14}
        &&\textbf{$L_\Im M_\Phi$}&\textbf{absences}&\cellcolor{blue!25}65.51&52.04&\cellcolor{blue!25}63.99&\cellcolor{blue!25}57.13&0.3569&0.0125&0.0657&\cellcolor{blue!25}0.9214&0.0455&\textbf{237.27}& 1\\

        \cline{2-15}

        &\multirow{5}{*}{\textbf{LRTDD}}&Default&-&63.88&51.1&56.64&53.19&0.3283&0.0163&0.0688&\cellcolor{blue!25}1.1674&0.0334&223.2& 2\\
        \cline{3-14}

        &&\multirow{2}{*}{$M_\Im M_\Phi$}&Medu&62.55&49.82&44.82&46.57&\cellcolor{blue!25}0.2693&0.024&0.0767&1.2715&\cellcolor{blue!25}0.0125&202.11& \multirow{2}{*}{4}\\
        &&&health&63.4&51.01&55.25&52.4&0.3242&0.019&\cellcolor{blue!25}0.0681&1.1683&0.0295&220.45&\\
        \cline{3-14}

        &&\multirow{2}{*}{$M_\Im L_\Phi$}&activities&63.88&\cellcolor{blue!25}51.59&55.8&52.96&0.3171&\cellcolor{blue!25}0.0101&0.0728&1.1979&0.0433&222.59& \multirow{2}{*}{1}\\
        &&&\textbf{sex}&\cellcolor{blue!25}63.98&51.21&\cellcolor{blue!25}60.66&\cellcolor{blue!25}54.89&0.3501&0.0244&0.1064&1.2496&0.0626&\textbf{228.95}&\\

        \cline{3-14}
        &&$L_\Im M_\Phi$&absences&63.11&50.59&55.61&52.26&0.3294&0.0119&0.0838&1.238&0.0713&219.84& 3\\

        \hline
        \hline
        

\multirow{8}{*}{\textbf{sex}}&\multirow{5}{*}{\textbf{FairSMOTE}}&Default&-&63.59&49.69&59.15&53.59&0.3618&0.0322&0.1496&1.1671&0.0457&224.26& 4\\
        \cline{3-14}

        &&\multirow{2}{*}{$M_\Im M_\Phi$}&Medu&61.96&49.28&61.54&54.31&0.3918&\cellcolor{blue!25}0.0069&\cellcolor{blue!25}0.0975&\cellcolor{blue!25}1.0859&0.0196&225.49& \multirow{2}{*}{1}\\

        &&&health&65.89&\cellcolor{blue!25}52.21&61.72&\cellcolor{blue!25}56.28&0.345&0.0283&0.1292&1.1006&\cellcolor{blue!25}0.0189&\textbf{234.48}&\\

        \cline{3-14}

        &&\textbf{$M_\Im L_\Phi$}&\textbf{activities}&\cellcolor{blue!25}64.93&51.1&\cellcolor{blue!25}62.33&55.73&0.3592&0.0427&0.1191&1.1587&0.0509&232.36& 2\\
        
        \cline{3-14}
        &&$L_\Im M_\Phi$&absences&64.83&51.65&61.11&55.58&\cellcolor{blue!25}0.3447&0.0405&0.1361&1.1991&0.0487&231.4&3 \\

        \cline{2-15}

        &\multirow{5}{*}{\textbf{LRTDD}}&Default&None&67.83&58.56&43.24&49.25&0.1796&0.0522&0.1084&1.0115&0.0515&217.48& 3\\
        \cline{3-14}

        &&\multirow{2}{*}{$M_\Im M_\Phi$}&Medu&64.85&54.46&34.94&42.17&\cellcolor{blue!25}0.1742&\cellcolor{blue!25}0.0101&0.1021&1.1757&\cellcolor{blue!25}0.0073&194.95&\multirow{2}{*}{4} \\

        &&&health&66.87&56.84&40.59&46.98&0.1794&0.0436&0.1064&1.0051&0.0476&209.9&\\
        \cline{3-14}

        &&\textbf{$M_\Im L_\Phi$}&\textbf{activities}&\cellcolor{blue!25}68.11&\cellcolor{blue!25}58.79&44.15&\cellcolor{blue!25}49.97&0.1814&0.0427&\cellcolor{blue!25}0.1004&\cellcolor{blue!25}0.9654&0.0554&\textbf{219.67}& 1\\

        \cline{3-14}
        &&$L_\Im M_\Phi$&absences&67.44&57.36&\cellcolor{blue!25}44.97&\cellcolor{blue!25}49.97&0.1951&0.0417&0.1189&1.0184&0.0427&218.32& 2\\

        \hline

        \bottomrule

\end{tabular}}


          \vspace{10pt}
    \end{subtable}

\hfill

    \begin{subtable}[h]{\textwidth}

        \caption{Comparison of the model performance and fairness before and after applying bias mitigation techniques on the  scenario}
       \label{tab:student-eval-mitigation-2}
       
       \centering
        \resizebox{0.95\textwidth}{!}{\begin{tabular}{l c| c | c c cc|ccc cc |g}
        

        \multirow{2}{*}{\textbf{Algo}}&\multirow{2}{*}{\textbf{Label}}&\multirow{2}{*}{\textbf{PBF}}&\textbf{ACC}&\textbf{PRE}&\textbf{Recall}&\textbf{F1}&\textbf{F-alarm}&\textbf{AOD}&\textbf{SPD}&\textbf{DIR}&\textbf{FPR$_D$}&\textbf{Total}\\

        &&&\textbf{W/T/L}&\textbf{W/T/L}&\textbf{W/T/L}&\textbf{W/T/L}&\textbf{W/T/L}&\textbf{W/T/L}&\textbf{W/T/L}&\textbf{W/T/L}&\textbf{W/T/L}&\textbf{W/T/L}\\ 

        \midrule
        \midrule
        
        \multirow{5}{*}{\textbf{FairSMOTE}}&Default&-&0/23/0&0/23/0&20/3/0&0/23/0&0/23/0&6/17/0&13/10/0&9/14/0&10/13/0&58/149/0\\
        \cline{2-12}
        &\multirow{2}{*}{$M_\Im M_\Phi$}&Medu&0/22/0&0/22/0&22/0/0&13/9/0&0/22/0&3/19/0&11/11/0&7/15/0&11/11/0&67/131/0\\
        &&health&0/22/0&0/22/0&22/0/0&1/21/0&0/22/0&4/18/0&10/12/0&10/12/0&8/14/0&55/143/0\\

        \cline{2-12}
        &\multirow{2}{*}{$M_\Im L_\Phi$}&activities&0/22/0&0/22/0&20/2/0&0/22/0&0/22/0&6/16/0&13/9/0&10/12/0&11/11/0&60/138/0\\
        &&sex&0/22/0&0/22/0&20/2/0&0/22/0&0/22/0&8/14/0&11/11/0&9/13/0&7/15/0&55/143/0\\
        
        \cline{2-12}
        &$L_\Im M_\Phi$&absences&0/22/0&0/22/0&21/1/0&0/22/0&0/22/0&6/16/0&12/10/0&8/14/0&7/15/0&54/144/0\\
    
\bottomrule


        
    

 \multirow{5}{*}{\textbf{LRTDD}}&Default&-&0/23/0&0/23/0&19/4/0&0/23/0&0/23/0&4/19/0&14/9/0&9/14/0&11/12/0&57/150/0\\
        \cline{2-12}
        &\multirow{2}{*}{$M_\Im M_\Phi$}&Medu&0/22/0&0/22/0&5/17/0&0/22/0&2/20/0&2/20/0&14/8/0&7/15/0&14/8/0&44/154/0\\
        &&health&0/22/0&0/22/0&17/5/0&4/18/0&0/22/0&3/19/0&12/10/0&10/12/0&11/11/0&57/141/0\\

        \cline{2-12}
        &\multirow{2}{*}{$M_\Im L_\Phi$}&activities&0/22/0&0/22/0&12/10/0&0/22/0&0/22/0&5/17/0&14/8/0&10/12/0&10/12/0&51/147/0\\
        &&sex&0/22/0&0/22/0&19/3/0&4/18/0&0/22/0&6/16/0&12/10/0&9/13/0&7/15/0&57/141/0\\
        
        \cline{2-12}
        &$L_\Im M_\Phi$&absences&0/22/0&0/22/0&14/8/0&5/17/0&0/22/0&6/16/0&11/11/0&10/12/0&7/15/0&53/145/0\\
    
\bottomrule

\end{tabular}}

       \vspace{5pt}
    \end{subtable}

   \end{table}
Table~\ref{tab:student-eval-before-1} shows evaluation results on the Student dataset evaluated on five different features, chosen as follows:  i.e., $M_\Im M_\Phi = \{Medu, health\}$ (i.e., the two features are both bias inducing and more important to model predictive performance, see Table~\ref{tab:stanility-student-ranking}, column `Label'), $M_\Im L_\Phi =\{activities, sex\}$ (more bias, yet less important), $L_\Im M_\Phi=\{absences\}$ (less biased, yet more important). The features removed in each of the scenarios are shown in column `PBF' (a.k.a, Potential Bias Feature), and the model performance and fairness are measured against the different features in the dataset shown in column `Feature'. For instance, the accuracy and fairness measures such as False-alarm, Average Odd Difference (AOD), Statistical parity difference (SPD), of the model on feature `age', after removing the feature that is most bias-inducing, yet also important to the model (i.e., `Medu'), among others. The row labelled as `Default' correspond to the model performance on the original data. The highlighted cells with darker backgrounds in Table~\ref{tab:student-eval-before} denote treatments that are performing better than the rest in the current runs. Also, column $T-score$ indicate the corresponding ranking of each scenario, computed using Equation (\ref{eqn:tscore}), the best score and the corresponding scenario are marked with \textbf{bolt-face}.

The result shows that removing the feature `activity' results in a reduced bias and higher model performance compared to the rest of the scenarios. This result indicates that removing such more bias-inducing features, yet less important feature, like `activities', and `sex', potentially improves the model fairness, without negatively impacting the model predicting performance. 
Therefore, a simple mitigation techniques could be by removing such features from the feature list, given the domain interpretation. On the other hand, our evaluation also shows that removing the feature `Medu' or `health' consistently indicate a negative impact on the model's predictive performance and model fairness. This is an indication that features `Medu' or `health' are more important to the model but are also more bias-inducing to the model. Thus, removing such features is not the best option, instead a better mitigation technique should be employed for such features. The feature `absences' is more important but less biased and according to Table~\ref{fig:shap-value-student-data}, removing it has negligible improvement in the fairness of the model decision. 

In Table~\ref{tab:student-eval-before-2} we compare the the values of the fairness and model predictive performance for the original Student dataset (default) and those after removing the potential bias inducing features. We count the number of win, tie, and loss cases compared
with the default scenarios and summarize the results for the model performance and the fairness. For instance, the win indicates that the performance metrics of the current scenario (when the feature is removed) are higher than the model performance on the original dataset (default). The win in the Fairness column, indicates the bias metrics are less in the current scenario than the default case, hence wins. According to Table~\ref{tab:student-eval-before-2}, the high number of losses for the model fairness was observed when the features `Medu' or `health' is removed from the dataset. Both `Medu' and `health' are categorized as most biased inducing, yet also most important, hence removing them highly affects the model decision negatively. On the other hand, close similar (i.e., could be slightly higher/ lower or same) performance and model fairness were observed when the feature `activities' is removed from the dataset. Combining the results from  Table~\ref{tab:student-eval-before-1}, and Table~\ref{tab:student-eval-before-2}, we can see that the model performs better after removing the features which are more bias-inducing yet less important to the model. The results are not statistically significant but are much better option than the rest of the scenarios. We believe that better bias mitigation techniques should be proposed given the awareness (breakdown) of such biased-inducing (and important) features in the dataset. 

Next, as part of our evaluation, we compare the performance and the fairness of the model when bias mitigation techniques are used on the original data and the modified data (i.e., after removing the potential bias feature). We report in Table~\ref{tab:student-eval-mitigation} the results of our evaluation for different scenarios when the three bias mitigation technique are used. The bias mitigation techniques considered in this study are FairSMOTE, LRTDD, and Reweighing. Table~\ref{tab:student-eval-mitigation-1} shows the performance and fairness metrics of the models after employing bias mitigation techniques on the original data (i.e., row `Default'), and the other scenarios. According to the result in Table~\ref{tab:student-eval-mitigation-1}, in many of the cases, the model performance and fairness improves after removing the most bias, yet less important feature of the data (i.e., category $M_\Im M_\Phi$). 
Note that, the higher number of wins (W) would indicate that the model performs better in the current scenario, only after applying the bias mitigation techniques, but does not tell if the model is indeed better than other scenarios; without referring to Table~\ref{tab:student-eval-before-1} and Table~\ref{tab:student-eval-mitigation-1}. For instance, according to Table~\ref{tab:student-eval-mitigation-2} the high number of wins is $67$ corresponding to when feature `Medu' is removed, indicating that the metrics `Recall', F1-scores, AOD, and FPR$_D$ are better after applying FairSMOTE algorithm on the modified data compared to the model performance and fairness of the model trained on the dataset without using the mitigation technique. Based on this results, we can say that, the model performance and fairness significantly improves after removing the most bias, yet also important features `only if' the mitigation technique is used.

\begin{tcolorbox}
Our evaluation result indicates that removing more biased, yet less important features, such as `activities', and `sex' can improve the model fairness, with little to no (or even improve) the predictive performance of the model trained on Student dataset. On the other hand, removing the features that are more biased and yet also more important improves the model prediction `only if' the mitigation technique is used. Therefore, a better bias mitigation technique is necessary (instead of simply removing such features). 
\end{tcolorbox}

\subsection{Cleveland Heart Dataset}\label{subsec:eval-cleveran-heart}

\begin{table}[h]
\Large 
\caption{
Comparing the model performance before and after removing the potentially biased features (PBF)  for \textbf{Cleveran Heart dataset}}
      \begin{subtable}[h]{\textwidth}

        \caption{Comparing the model performance on the original model (i.e., no mitigation techniques are used) before and after removing the potentially biased features (PBF)}
       \label{tab:cleveran-heart-eval-before-1}
       
        \centering
        \resizebox{\textwidth}{!}{\begin{tabular}{l c| r | r r rr|rrr r r |g c}
        

        \textbf{Feature}&\textbf{Label}&\textbf{PBF}&\textbf{ACC}&\textbf{PRE}&\textbf{Recall}&\textbf{F1}&\textbf{F-alarm}&\textbf{AOD}&\textbf{SPD}&\textbf{DIR}&\textbf{FPR$_D$}&\textbf{T-Score}&\textbf{RANK}\\

        \midrule
        \midrule
        
        \multirow{5}{*}{\textbf{age}}&\textbf{Default}&-&\cellcolor{blue!25}84.17&85.32&\cellcolor{blue!25}80.16&\cellcolor{blue!25}81.96&\cellcolor{blue!25}0.1236&0.0227&\cellcolor{blue!25}0.3098&\cellcolor{blue!25}2.235&0.1459&\textbf{328.77}&1\\
        \cline{2-13}
        &\multirow{2}{*}{$M_\Im M_\Phi$}&ca&81.14&80.95&76.86&78.32&0.1536&0.0624&0.3679&2.5647&0.2637&313.86&\multirow{2}{*}{4}\\
        
        & &thal&81.45&84.97&75.05&78.82&0.1241&0.0289&0.3196&2.449&\cellcolor{blue!25}0.1303&317.24& \\
        \cline{2-13}
        &$M_\Im L_\Phi$&restecg&83.83&85.29&79.5&81.65&0.1242&\cellcolor{blue!25}0.0154&0.3293&2.368&0.1647&327.27&2\\
        \cline{2-13}
        &$L_\Im M_\Phi$&oldpeak&83.48&\cellcolor{blue!25}85.35&78.77&81.21&\cellcolor{blue!25}0.1236&0.0276&0.3228&2.4844&0.1647&325.69&3\\
        \hline

        \multirow{5}{*}{\textbf{exang}}&Default&-&84.17&85.32&\cellcolor{blue!25}80.16&81.96&0.1236&0.0413&0.5658&0.3101&0.2322&330.34&2\\
        \cline{2-13}

        &\multirow{2}{*}{$M_\Im M_\Phi$}&ca&81.14&80.95&76.86&78.32&0.1536&0.1012&\cellcolor{blue!25}0.5324&0.3371&\cellcolor{blue!25}0.157&315.99&\multirow{2}{*}{4}\\
        & &thal&81.45&84.97&75.05&78.82&0.1241&\cellcolor{blue!25}0.0434&0.5604&\cellcolor{blue!25}0.2905&0.2515&319.02& \\
        \cline{2-13}
        
        &\multirow{2}{*}{\textbf{$M_\Im L_\Phi$}}&restecg&83.83&85.29&79.5&81.65&0.1242&0.0511&0.5725&0.3003&0.2327&328.99&\multirow{2}{*}{1}\\
        & &\textbf{age}&\cellcolor{blue!25}84.51&\cellcolor{blue!25}85.98&\cellcolor{blue!25}80.16&\cellcolor{blue!25}82.31&\cellcolor{blue!25}0.1179&0.0586&0.5577&0.3158&0.1977&\textbf{331.71}& \\
        
        \cline{2-13}
        &$L_\Im M_\Phi$&oldpeak&83.48&85.35&78.77&81.21&0.1236&0.0699&0.5447&0.3169&0.175&327.06&3\\
        \hline
        
        \multirow{4}{*}{\textbf{sex}}&Default&-&84.17&85.32&\cellcolor{blue!25}80.16&81.96&0.1236&0.0138&0.3338&2.6009&0.1539&328.39&2\\
        \cline{2-13}

        &\multirow{2}{*}{$M_\Im M_\Phi$}&ca&81.14&80.95&76.86&78.32&0.1536&0.3306&0.0271&\cellcolor{blue!25}2.47&0.1847&314.1&\multirow{2}{*}{4}\\
        & &thal&81.45&84.97&75.05&78.82&0.1241&0.0448&0.3523&3.0634&0.1591&316.54& \\
        \cline{2-13}
        &\multirow{2}{*}{\textbf{$M_\Im L_\Phi$}}&restecg&83.83&85.29&79.5&81.65&0.1242&0.3136&0.0076&3.1493&0.1337&326.54&\multirow{2}{*}{1}\\
        & &\textbf{age}&\cellcolor{blue!25}84.51&\cellcolor{blue!25}85.98&\cellcolor{blue!25}80.16&\cellcolor{blue!25}82.31&\cellcolor{blue!25}0.1179&\cellcolor{blue!25}0.0093&0.329&2.5796&0.1448&\textbf{329.78}& \\
        
        \cline{2-13}
        &$L_\Im M_\Phi$&oldpeak&83.48&85.35&78.77&81.21&0.1236&0.3229&\cellcolor{blue!25}0.0045&3.1281&\cellcolor{blue!25}0.1289&326.54&3\\

        \hline
        
        \multirow{4}{*}{\textbf{thal}}&Default&-&84.17&85.32&\cellcolor{blue!25}80.16&81.96&0.1236&0.5852&\cellcolor{blue!25}0.0045&0.2639&0.3051&330.33&2\\
        \cline{2-13}

        &$M_\Im M_\Phi$&ca&81.14&80.95&76.86&78.32&0.1536&0.6015&0.1058&0.2664&0.5236&315.62&4\\
        
        \cline{2-13}
        &\multirow{2}{*}{\textbf{$M_\Im L_\Phi$}}&restecg&83.83&85.29&79.5&81.65&0.1242&0.5669&0.0148&0.2736&\cellcolor{blue!25}0.2734&329.01&\multirow{2}{*}{1}\\

        & &\textbf{age}&\cellcolor{blue!25}84.51&\cellcolor{blue!25}85.98&\cellcolor{blue!25}80.16&\cellcolor{blue!25}82.31&\cellcolor{blue!25}0.1179&\cellcolor{blue!25}0.0402&0.5925&0.2589&0.3944&\textbf{331.55}& \\

        \cline{2-13}
        &$L_\Im M_\Phi$&oldpeak&83.48&85.35&78.77&81.21&0.1236&0.5845&0.0082&\cellcolor{blue!25}0.2586&0.3051&327.53&3\\
\bottomrule
\end{tabular}}
\vspace{10pt}
\end{subtable}

    \hfill

 \begin{subtable}[h]{\textwidth}
   \tiny

        \caption{Model performance and fairness when no mitigation techniques are used}
       \label{tab:cleveran-heart-eval-before-2}
       
       \centering
        \resizebox{0.95\textwidth}{!}{\begin{tabular}{l| c | c c c|cc c}
        

        \multirow{2}{*}{\textbf{Label}}&\multirow{2}{*}{\textbf{PBF}}&\multicolumn{3}{c|}{\textbf{Performance}}&\multicolumn{3}{c|}{\textbf{Fairness}}\\
        

        &&\textbf{Win}&\textbf{Tie}&\textbf{Loss}&\textbf{Win}&\textbf{Tie}&\textbf{Loss}\\

        \midrule
        \midrule
        
       \multirow{2}{*}{$M_\Im M_\Phi$}&ca&0&11&0&0&4&7\\
       &thal&0&11&0&0&7&4\\
        \hline
        \multirow{2}{*}{$M_\Im L_\Phi$}&restecg&0&11&0&0&9&2\\
        
        &age&0&11&0&0&10&1\\

        \hline
        $L_\Im M_\Phi$&oldpeak&0&11&0&0&10&1\\
    
\bottomrule

\end{tabular}}

       \vspace{5pt}
    \end{subtable}

\end{table}

\begin{table}[h]
\Large
\caption{
Comparing the model performance and fairness, on \textbf{Cleveran Heart dataset}, for different scenarios, with bias mitigation techniques used}
     \label{tab:cleveran-heart-eval-mitigation}
     
        \begin{subtable}[h]{\textwidth}
        \caption{Model performance and fairness when the mitigation techniques are used}
       \label{tab:cleveran-heart-eval-mitigation-1}
        \centering
        \resizebox{\textwidth}{!}{
    \begin{tabular}{l l c m{2cm} | r r rr|rrr r r |g c}
        

        \textbf{Feature}&\textbf{Algo}&\textbf{Cases}&\textbf{PBF}&\textbf{ACC}&\textbf{PRE}&\textbf{Recall}&\textbf{F1}&\textbf{F-alarm}&\textbf{AOD}&\textbf{SPD}&\textbf{DIR}&\textbf{FPR$_D$}&\textbf{T-Score}&\textbf{RANK}\\

\midrule
\midrule


        


        \multirow{10}{*}{\textbf{\textbf{Age}}}&\multirow{5}{*}{\textbf{FairSMote}}&Default&-&71.32&67.48&\cellcolor{blue!25}89.55&75.23&0.4171&0.049&0.2257&1.6535&0.1269&298.14& 2\\
        \cline{3-14}
        &&$M_\Im M_\Phi$&ca&67.62&64.33&85.97&71.62&0.4521&0.106&\cellcolor{blue!25}0.2036&1.6567&0.1908&289.39& 4\\
        \cline{3-14}
        &&$M_\Im L_\Phi$$^{**}$&\textbf{restecg}&\cellcolor{blue!25}72.33&\cellcolor{blue!25}68.93&88.83&\cellcolor{blue!25}75.75&\cellcolor{blue!25}0.3913&0.1025&0.2246&\cellcolor{blue!25}1.6103&\cellcolor{blue!25}0.1781&\textbf{302.18}& 1\\
        \cline{3-14}
        &&$L_\Im M_\Phi$&oldpeak&67.61&65.27&85.06&71.72&0.4524&\cellcolor{blue!25}0.0728&0.2255&1.7419&0.1831&293.07& 3\\
        \cline{2-14}
        \cline{2-15}


        &\multirow{5}{*}{\textbf{Reweighing}}&Default&-&71.4&65.19&\cellcolor{blue!25}93.98&\cellcolor{blue!25}75.64&0.4588&0.1245&0.3258&1.8604&0.3146&\textbf{303.13}& 1\\
        \cline{3-14}
        &&$M_\Im M_\Phi$&ca&69.37&63.14&93.89&74.18&0.4958&0.1281&0.3166&1.9276&0.3268&297.38& 3\\
        \cline{3-14}
        
        &&$M_\Im L_\Phi$&restecg&\cellcolor{blue!25}72.05&\cellcolor{blue!25}66.6&90.45&75.22&\cellcolor{blue!25}0.4149&0.0742&0.3204&1.9364&0.2566&301.31& 2\\
        \cline{3-14}
        &&$L_\Im M_\Phi$&oldpeak&66.68&60.64&93.77&72.51&0.5479&\cellcolor{blue!25}0.0708&\cellcolor{blue!25}0.2439&\cellcolor{blue!25}1.6992&\cellcolor{blue!25}0.2028&290.83& 4\\
        \cline{2-14}
        \cline{2-15}


        &\multirow{5}{*}{\textbf{LRTDD}}&Default$^{**}$&-&71.99&\cellcolor{blue!25}70.16&85.62&74.87&\cellcolor{blue!25}0.3745&0.1581&\cellcolor{blue!25}0.1739&\cellcolor{blue!25}1.2971&\cellcolor{blue!25}0.1719&\textbf{300.46}& 1\\
        \cline{3-14}
        &&$M_\Im M_\Phi$&ca&67.62&64.33&85.97&71.62&0.4521&0.106&0.2036&1.6567&0.1908&287.03& 4\\
        \cline{3-14}
        
        &&$M_\Im L_\Phi$&restecg&\cellcolor{blue!25}72.33&68.93&\cellcolor{blue!25}88.83&\cellcolor{blue!25}75.75&0.3913&0.1025&0.2246&1.6103&0.1781&295.84& 3\\
        \cline{3-14}
        &&$L_\Im M_\Phi$&oldpeak&67.61&65.27&85.06&71.72&0.4524&\cellcolor{blue!25}0.0728&0.2255&1.7419&0.1831&300.07&2 \\
        \hline
        \hline

        \multirow{10}{*}{\textbf{exang}}&\multirow{5}{*}{\textbf{FairSMote}}&Default&-&67.3&63.73&86.97&71.71&0.4637&\cellcolor{blue!25}0.0089&0.2991&0.6524&0.1007&288.19& 2\\
        \cline{3-14}

        &&$M_\Im M_\Phi$&ca&65.97&62.47&85.33&70.28&0.4763&0.0103&0.2637&0.6853&\cellcolor{blue!25}0.0534&279.77&4 \\
        \cline{3-14}
        
        &&$M_\Im L_\Phi$$^{**}$&\textbf{restecg}&\cellcolor{blue!25}68.97&\cellcolor{blue!25}65.81&85.44&\cellcolor{blue!25}72.48&\cellcolor{blue!25}0.422&0.0185&0.311&\cellcolor{blue!25}0.624&0.1084&\textbf{289.85}& 1\\
        
        \cline{3-14}
        &&$L_\Im M_\Phi$&oldpeak&67.63&65.09&84.99&71.62&0.4455&0.0797&\cellcolor{blue!25}0.249&0.6836&0.0544&286.32& 3\\
        \cline{2-15}

        
       &\multirow{5}{*}{\textbf{Reweighing}}&Default&-&71.4&65.19&\cellcolor{blue!25}93.98&\cellcolor{blue!25}75.64&0.4588&0.1182&0.3968&0.5842&0.311&\textbf{304.34}& \textbf{1} \\
        \cline{3-14}

        &&$M_\Im M_\Phi$&ca&69.37&63.14&93.89&74.18&0.4958&0.088&0.3874&0.5985&0.2715&298.74& 3\\
        \cline{3-14}
        
        &&$M_\Im L_\Phi$$^{**}$&restecg&\cellcolor{blue!25}72.05&\cellcolor{blue!25}66.6&90.45&75.22&\cellcolor{blue!25}0.4149&\cellcolor{blue!25}0.0752&0.4452&\cellcolor{blue!25}0.5279&0.3052&302.55& 2\\
        
        \cline{3-14}
        &&$L_\Im M_\Phi$&oldpeak&66.68&60.64&93.77&72.51&0.5479&0.0844&\cellcolor{blue!25}0.3571&0.6371&\cellcolor{blue!25}0.2657&291.71& 4\\
        \cline{2-14}
        \cline{2-15}

        
       &\multirow{5}{*}{\textbf{LRTDD}}&Default&-&72.07&65.05&88.63&74.31&0.4085&\cellcolor{blue!25}0.0009&0.3649&0.5917&0.1327&298.56& 2\\
        \cline{3-14}

        &&$M_\Im M_\Phi$&ca&65.97&62.47&85.33&70.28&0.4763&0.0103&0.2637&0.6853&\cellcolor{blue!25}0.0534&293.57& 4\\
        \cline{3-14}
        
        &&$M_\Im L_\Phi$$^{**}$&\textbf{restecg}&68.97&\cellcolor{blue!25}65.81&85.44&72.48&0.422&0.0185&0.311&0.624&0.1084&\textbf{302.44}& 1\\
        
        \cline{3-14}
        &&$L_\Im M_\Phi$&oldpeak&67.63&65.09&84.99&71.62&0.4455&0.0797&\cellcolor{blue!25}0.249&0.6836&0.0544&295.41& 3\\
        \hline
        \hline
        

        \multirow{8}{*}{\textbf{sex}}&\multirow{5}{*}{\textbf{FairSMote}}&Default&-&71.29&69.38&85.62&74.43&\cellcolor{blue!25}0.3792&0.0159&0.2888&2.8579&0.1602&292.64& 2\\
        \cline{3-14}

        &&$M_\Im M_\Phi$&ca&67.97&64.76&85.41&71.57&0.4429&\cellcolor{blue!25}0.0071&\cellcolor{blue!25}0.2351&\cellcolor{blue!25}1.3971&0.1068&286.05&4 \\
        \cline{3-14}
        &&$M_\Im L_\Phi$&\textbf{restecg}&\cellcolor{blue!25}71.33&69&\cellcolor{blue!25}85.92&74.25&0.3806&0.0164&0.2441&2.3804&\cellcolor{blue!25}0.1047&\textbf{294.01}& 1\\
        
        \cline{3-14}
        &&$L_\Im M_\Phi$&oldpeak&68.95&66.36&85.31&72.52&0.4241&0.0549&0.2504&2.0997&0.1501&290.74& 3\\

        \cline{2-15}
        \cline{2-15}

        &\multirow{5}{*}{\textbf{Reweighing}}&\textbf{Default}&-&71.4&65.19&\cellcolor{blue!25}93.98&\cellcolor{blue!25}75.64&0.4588&\cellcolor{blue!25}0.0451&0.3993&2.8954&\cellcolor{blue!25}0.2949&\textbf{302.12}& 1\\
        \cline{3-14}

        &&$M_\Im M_\Phi$&ca&69.37&63.14&93.89&74.18&0.4958&0.0537&0.4187&\cellcolor{blue!25}1.6674&0.3643&297.58&3 \\
        \cline{3-14}
        &&$M_\Im L_\Phi$&restecg&\cellcolor{blue!25}72.05&\cellcolor{blue!25}66.6&90.45&75.22&\cellcolor{blue!25}0.4149&0.0939&0.4481&3.269&0.3921&299.7& 2\\
        
        \cline{3-14}
        &&$L_\Im M_\Phi$&oldpeak&66.68&60.64&93.77&72.51&0.5479&0.0982&\cellcolor{blue!25}0.3983&2.6492&0.3728&289.53& 4\\
        \cline{2-15}
        \cline{2-15}
        &\multirow{5}{*}{\textbf{LRTDD}}&Default&-&67.31&62.66&89.85&72.23&0.493&0.0436&\cellcolor{blue!25}0.2084&1.4674&\cellcolor{blue!25}0.0088&289.83& 2\\
        \cline{3-14}

        &&$M_\Im M_\Phi$&ca&67.97&64.76&85.41&71.57&0.4429&\cellcolor{blue!25}0.0071&0.2351&\cellcolor{blue!25}1.3971&0.1068&278.44& 4\\
        \cline{3-14}
        &&$M_\Im L_\Phi$$^{**}$&\textbf{restecg}&\cellcolor{blue!25}71.33&\cellcolor{blue!25}69&85.92&\cellcolor{blue!25}74.25&\cellcolor{blue!25}0.3806&0.0164&0.2441&2.3804&0.1047&\textbf{294.25}& 1\\
        
        \cline{3-14}
        &&$L_\Im M_\Phi$&oldpeak&68.95&66.36&85.31&72.52&0.4241&0.0549&0.2504&2.0997&0.1501&282.86& 3\\

       \bottomrule

\end{tabular}}

\vspace{10pt}
\end{subtable}

\hfill

    \begin{subtable}[h]{\textwidth}
    \Large

        \caption{Summary results comparing the model performance and fairness before and after applying bias mitigation techniques on the current scenario}
       \label{tab:cleveran-heart-eval-mitigation-2}
       
        \centering
        \resizebox{0.95\textwidth}{!}{\begin{tabular}{l c| c | c c cc|ccc cc |g}
        

        \multirow{2}{*}{\textbf{Feature}}&\multirow{2}{*}{\textbf{Label}}&\multirow{2}{*}{\textbf{PBF}}&\textbf{ACC}&\textbf{PRE}&\textbf{Recall}&\textbf{F1}&\textbf{F-alarm}&\textbf{AOD}&\textbf{SPD}&\textbf{DIR}&\textbf{FPR$_D$}&\textbf{Total}\\

        &&&\textbf{W/T/L}&\textbf{W/T/L}&\textbf{W/T/L}&\textbf{W/T/L}&\textbf{W/T/L}&\textbf{W/T/L}&\textbf{W/T/L}&\textbf{W/T/L}&\textbf{W/T/L}&\textbf{W/T/L}\\

        \midrule
        \midrule
        
        \multirow{5}{*}{\textbf{FairSMOTE}}&Default&-&0/12/0&0/12/0&10/2/0&0/12/0&0/12/0&4/8/0&8/4/0&2/10/0&1/11/0&25/83/0\\
        \cline{2-12}
        &\multirow{2}{*}{$M_\Im M_\Phi$}&ca&0/11/0&0/11/0&9/2/0&0/11/0&0/11/0&6/5/0&7/4/0&4/7/0&3/8/0&29/70/0\\
        &&thal&0/11/0&0/11/0&9/2/0&0/11/0&0/11/0&4/7/0&5/6/0&3/8/0&0/11/0&21/78/0\\

        \cline{2-12}
        &$M_\Im L_\Phi$&restecg&0/11/0&0/11/0&8/3/0&0/11/0&0/11/0&4/7/0&9/2/0&2/9/0&1/10/0&24/75/0\\
        
        \cline{2-12}
        &$M_\Im L_\Phi$&oldpeak&0/11/0&0/11/0&8/3/0&0/11/0&0/11/0&3/8/0&9/2/0&3/8/0&1/10/0&24/75/0\\
    
\bottomrule

 \multirow{5}{*}{\textbf{Reweighing}}&Default&-&0/12/0&0/12/0&12/0/0&0/12/0&0/12/0&4/8/0&8/4/0&2/10/0&0/12/0&26/82/0\\
        \cline{2-12}
        &\multirow{2}{*}{$M_\Im M_\Phi$}&ca&0/11/0&0/11/0&11/0/0&0/11/0&0/11/0&3/8/0&5/6/0&3/8/0&2/9/0&24/75/0\\
        &&thal&0/11/0&0/11/0&11/0/0&0/11/0&0/11/0&3/8/0&5/6/0&2/9/0&0/11/0&21/78/0\\

        \cline{2-12}
        &$M_\Im L_\Phi$&restecg&0/11/0&0/11/0&11/0/0&0/11/0&0/11/0&4/7/0&7/4/0&1/10/0&0/11/0&23/76/0\\
        \cline{2-12}
        &$M_\Im L_\Phi$&oldpeak&0/11/0&0/11/0&11/0/0&0/11/0&0/11/0&3/8/0&6/5/0&3/8/0&0/11/0&23/76/0\\
    
\bottomrule

 \multirow{5}{*}{\textbf{LRTDD}}&Default&-&0/12/0&0/12/0&9/3/0&0/12/0&0/12/0&7/5/0&8/4/0&2/10/0&1/11/0&27/81/0\\
        \cline{2-12}
        &\multirow{2}{*}{$M_\Im M_\Phi$}&ca&0/11/0&0/11/0&8/3/0&0/11/0&0/11/0&6/5/0&7/4/0&2/9/0&4/7/0&27/72/0\\
        &&thal&0/11/0&0/11/0&8/3/0&0/11/0&0/11/0&5/6/0&6/5/0&2/9/0&0/11/0&21/78/0\\

        \cline{2-12}
        &$M_\Im L_\Phi$&restecg&0/11/0&0/11/0&8/3/0&0/11/0&0/11/0&5/6/0&8/3/0&3/8/0&1/10/0&25/74/0\\
        
        \cline{2-12}
        &$M_\Im L_\Phi$&oldpeak&0/11/0&0/11/0&9/2/0&0/11/0&0/11/0&2/9/0&7/4/0&3/8/0&0/11/0&21/78/0\\

        \bottomrule

\end{tabular}}

\vspace{5pt}
\end{subtable}

\end{table}

This Section details the  evaluation results on the Cleveland Heart dataset, chosen as follows: $M_\Im M_\Phi = \{ca, thal\}$ (i.e., the two features are both bias inducing and more important to model predictive performance, $M_\Im L_\Phi = \{restcg, age\}$ (i.e., more bias, yet less important), and $L_\Im M_\Phi = \{oldpeak\}$ (for less bias, yet more important), refer to Table~\ref{tab:stanility-cleveran-ranking}, column `Label').
The evaluation results are detailed in Table~\ref{tab:cleveran-heart-eval-before-1}. The result in Table~\ref{tab:cleveran-heart-eval-before-1} shows that removing the feature `restecg'  reduces the model bias and slightly improves the model performance compared to the rest of the scenarios. 

In Table~\ref{tab:cleveran-heart-eval-before-2} we compare the significance in the model fairness and model predictive performance on the original dataset (default) vs. the scenario when the potential bias inducing features is removed. Specifically, we reported in Table~\ref{tab:cleveran-heart-eval-before-2} the total count of win, tie, and loss cases compared
with the default scenarios for the model performance and the fairness. The results in Table~\ref{tab:student-eval-before-2} shows that removing the more bias-inducing yet also more important features is not a good option, among the rest of the evaluation scenarios. This is because  there is a high number of losses in the model fairness when compared to the model performance when the model is trained on the original dataset. On the other hand, removing the more biased, yet less important features does not result in many losses, but instead shows a close similar and/ or a better performance and model fairness, specifically when the feature `restecg' is removed. This evaluation result is consistent with the conclusion made on \textbf{Student dataset}, in Section~\ref{subsec:eval-student}. 

Next, in Table~\ref{tab:cleveran-heart-eval-mitigation-1} and Table~\ref{tab:cleveran-heart-eval-mitigation-2} we compare the performance of the model when the mitigation techniques are used on the dataset after removing the potently biased features indicated in column `PBF'. Table~\ref{tab:cleveran-heart-eval-mitigation-1} indicate that model performance and fairness metrics are better for the case where the feature which is more bias-inducing yet less important is removed. Specifically when the feature `restecg' is removed, highlighted in \textbf{bold face font}. Also, we compare the significance of the difference in model performance and fairness before and after employing the bias mitigation technique in Table~\ref{tab:cleveran-heart-eval-mitigation-2}. The results show a higher number of win counts corresponding to the scenario when the more bias-inducing, yet also more important feature  `ca' is removed, indicating that the model performance increase after applying the mitigation technique, unlike the first case (where no mitigation was used). 

\begin{tcolorbox}
Our result indicates a non-significant improvement in the model fairness and performance when we removed `restecg' or `age' (the more bias-inducing, yet less important features) in the dataset. Removing the features `ca' or `thal' which are more bias-inducing, yet important features shows a negative impact on model fairness and performance,--- significant improvement is observed only after applying the mitigation technique.  A better bias mitigation technique can be proposed. 
\end{tcolorbox}

\subsection{COMPAS Dataset}

\begin{table}[h]
\Large 
\caption{
Comparing the model performance and fairness measure before and after removing the potentially biased features (PBF)  for \textbf{COMPAS dataset}}
     \label{tab:cleveran-heart-eval-before}
      \begin{subtable}[h]{\textwidth}

        \caption{The model performance and fairness on the original data}
      \label{tab:compas-eval-before-1}
       
        \centering
        \resizebox{0.95\textwidth}{!}{\begin{tabular}{l c| r | r r rr|rrr r r |g c}
        

        \textbf{PBF}&\textbf{Cases}&\textbf{Feature}&\textbf{ACC}&\textbf{PRE}&\textbf{Recall}&\textbf{F1}&\textbf{F-alarm}&\textbf{AOD}&\textbf{SPD}&\textbf{DIR}&\textbf{FPR$_D$}&\textbf{T-Score}&\textbf{RANK}\\

        \midrule
        \midrule
        
        \multirow{5}{*}{\textbf{Age}}&\textbf{Default}&-&\cellcolor{blue!25}60.35&\cellcolor{blue!25}62.55&\cellcolor{blue!25}69.63&\cellcolor{blue!25}65.81&0.5082&0.0332&0.4332&0.489&0.0126&\textbf{256.87}& 1\\
        \cline{2-13}
        &&race&60.05&62.46&68.67&65.34&\cellcolor{blue!25}0.5034&0.0321&\cellcolor{blue!25}0.4403&\cellcolor{blue!25}0.4819&\cellcolor{blue!25}0.021&255.04& 2\\
        \cline{2-13}
        &&sex&59.54&61.85&69.08&65.21&0.5199&\cellcolor{blue!25}0.0095&0.4439&0.4975&0.0096&254.2& 3\\
        \hline

        \multirow{5}{*}{\textbf{c\_charge\_degree}}&Default&-&\cellcolor{blue!25}60.35&\cellcolor{blue!25}62.55&69.63&65.81&\cellcolor{blue!25}0.5082&0.0475&\cellcolor{blue!25}0.4074&0.5345&0.4457&256.4& 2\\
        \cline{2-13}

        &\textbf{MM}&\textbf{age}&57.46&58.69&\cellcolor{blue!25}77.1&\cellcolor{blue!25}66.41&0.6629&\cellcolor{blue!25}0.0495&0.4283&0.5717&0.4787&\textbf{257.47}& 1\\
        \cline{2-13}
        &&race&60.05&62.46&68.67&65.34&0.5034&0.0524&0.4103&0.5287&0.4545&254.57& 3\\
        
        \cline{2-13}
        &&sex&59.54&61.85&69.08&65.21&0.5199&0.0699&0.4533&\cellcolor{blue!25}0.5001&\cellcolor{blue!25}0.5198&253.61& 4\\
        \hline
        
        \multirow{4}{*}{\textbf{priors\_count}}&Default&-&\cellcolor{blue!25}60.35&\cellcolor{blue!25}62.55&69.63&65.81&0.5082&0.0031&\cellcolor{blue!25}0.185&0.7461&0.1722&256.73& 2\\
        \cline{2-13}

       &\textbf{MM}&\textbf{age}&57.46&58.69&\cellcolor{blue!25}77.1&\cellcolor{blue!25}66.41&0.6629&0.0626&0.2721&\cellcolor{blue!25}0.7239&\cellcolor{blue!25}0.3368&\textbf{257.6}& 1\\
       \cline{2-13}
        &&race&60.05&62.46&68.67&65.34&\cellcolor{blue!25}0.5034&\cellcolor{blue!25}0.0067&\cellcolor{blue!25}0.185&0.7434&0.1692&254.91& 3\\
        
        \cline{2-13}
        &&sex&59.54&61.85&69.08&65.21&0.5199&0.0104&0.1964&0.7326&0.1763&254.04& 4\\

        \hline
        
        \multirow{4}{*}{\textbf{sex}}&Default&-&\cellcolor{blue!25}60.35&\cellcolor{blue!25}62.55&69.63&65.81&0.5082&0.0792&0.4051&0.5685&0.4854&256.3& 2\\
        \cline{2-13}

        &\textbf{MM}&\textbf{age}&57.46&58.69&\cellcolor{blue!25}77.1&\cellcolor{blue!25}66.41&0.6629&0.0508&\cellcolor{blue!25}0.3442&0.6558&0.3978&\textbf{257.55}& 1\\
        
        \cline{2-13}
        &&race&60.05&62.46&68.67&65.34&\cellcolor{blue!25}0.5034&\cellcolor{blue!25}0.0855&0.4147&\cellcolor{blue!25}0.5586&\cellcolor{blue!25}0.5038&254.45& 3\\

         \hline
        
        \multirow{4}{*}{\textbf{race}}&Default&&\cellcolor{blue!25}60.35&\cellcolor{blue!25}62.55&69.63&65.81&\cellcolor{blue!25}0.5082&0.0339&\cellcolor{blue!25}0.2584&0.6706&0.2824&256.59& 2\\
        \cline{2-13}

        &\textbf{MM}&\textbf{age}&57.46&58.69&\cellcolor{blue!25}77.1&\cellcolor{blue!25}66.41&0.6629&\cellcolor{blue!25}0.0589&0.4213&\cellcolor{blue!25}0.5787&\cellcolor{blue!25}0.4831&\textbf{257.45}&1 \\
        
        \cline{2-13}
        &&sex&59.54&61.85&69.08&65.21&0.5199&0.038&0.3134&0.6182&0.3443&253.84& 3\\

       \end{tabular}}
\vspace{10pt}
\end{subtable}

    \hfill

 \begin{subtable}[h]{\textwidth}
   \small

        \caption{Model performance and fairness when no mitigation techniques are used}
       \label{tab:compas-eval-before-2}
       
       \centering
        \resizebox{0.95\textwidth}{!}{
        \begin{tabular}{l| c | c c c|cc c}
        

        \multirow{2}{*}{\textbf{Label}}&\multirow{2}{*}{\textbf{PBF}}&\multicolumn{3}{c|}{\textbf{Performance}}&\multicolumn{3}{c}{\textbf{Fairness}}\\
        

        &&\textbf{Win}&\textbf{Tie}&\textbf{Loss}&\textbf{Win}&\textbf{Tie}&\textbf{Loss}\\

        \midrule
        \midrule
        
       $M_\Im M_\Phi$&age&0&4&0&0&3&1\\
        \hline
        \multirow{4}{*}{Undefined}&race&0&4&0&0&4&0\\
        
        &sex&0&4&0&0&2&2\\
        &priors\_count&0&4&0&0&2&2\\
        &c\_charge\_degree&0&4&0&0&2&2\\

\bottomrule

\end{tabular}}

       \vspace{5pt}
    \end{subtable}
    
    \end{table}

This Section details the evaluation of the model performance on the COMPAS Dataset. For the COMPAS Dataset, since it was hard to categorize most of features based on the ranking of our framework and SHAP values (except for the `age' feature, which we categorised as being more bias-inducing and yet more important feature, $M_\Im M_\Phi$), we compare the model performance when each of the features is removed, independently, as part of the evaluation. Table~\ref{tab:compas-eval-before-1} shows the results of our evaluation when a given feature is removed. Our results show 
in the majority of the cases ($3/5$), the model trained after removing the `age' feature performs better than the rest of the cases. Specifically, we see that the Recall, F1-score, and the fairness measures improve when measured in the feature `prior\_count', `sex', and `race'. To better understand the variability in the model performance, we can compare the natural impact of these features when the mediating variable is `age', as reported earlier in Table~\ref{tab:double-swap-compas-data}. In Table~\ref{tab:double-swap-compas-data}, our results show that the `age' has a higher direct mediation effect on the features: `sex' and `race'. While we see the mediation effect between the `age' variable on the features `prior\_count', `c\_charge\_degree', we suspect that removing the mediating impact of feature `age' on race, and sex will subsequently result in the reduced bias in the rest of the features.

\begin{tcolorbox}
Likewise, for the COMPAS dataset, removing the potentially bias-inducing yet important feature of the model (in this case the `age') negatively impacts the resulting model decision. Instead, a significant improvement is observed only after applying bias mitigation.  A better bias mitigation can be proposed, instead of removing such features. 
\end{tcolorbox}

\subsection{Bank Dataset}

\begin{table}[h]
\Large 
\caption{
Comparing the model performance and fairness metrics when the potential biased features (PBF) is removed and no mitigation techniques are used, for the \textbf{Bank dataset}
}  
     \label{tab:bank-eval-before}
       
        \centering
        \resizebox{\textwidth}{!}{\begin{tabular}{l c| r | r r rr|rrr r r |g c}
        

        \textbf{Feature}&\textbf{Cases}&\textbf{PBF}&\textbf{ACC}&\textbf{PRE}&\textbf{Recall}&\textbf{F1}&\textbf{F-alarm}&\textbf{AOD}&\textbf{SPD}&\textbf{DIR}&\textbf{FPR$_D$}&\textbf{T-Score}&\textbf{RANK}\\

        \midrule
        \midrule
        
        \multirow{5}{*}{\textbf{age}}&\textbf{Default}&&\cellcolor{blue!25}68.52&50.89&\cellcolor{blue!25}31.1&\cellcolor{blue!25}38.12&0.1292&0.0008&0.1571&0.6668&0.0744&\textbf{187.60}& 1\\
        \cline{2-13}
        &$M_\Im M_\Phi$&duration&51.87&48.29&22.89&30.61&0.2454&0.037&0.1855&0.8383&0.113&152.24& 4\\
        \cline{2-13}
        &\multirow{2}{*}{$M_\Im L_\Phi$}&loan&67.74&50.9&30.25&37.37&0.1282&0.0051&\cellcolor{blue!25}0.1241&0.7286&\cellcolor{blue!25}0.0545&185.22& \multirow{2}{*}{2}\\
        &&default&68.15&49.24&29.47&36.42&0.1333&\cellcolor{blue!25}0&0.1588&\cellcolor{blue!25}0.6573&0.0769&182.25& \\
         \cline{2-13}
        &$L_\Im M_\Phi$&housing&66.35&\cellcolor{blue!25}51&28.24&35.35&\cellcolor{blue!25}0.1211&0.0223&0.1377&0.8038&0.0281&179.83& 3\\
        \hline

        \multirow{5}{*}{\textbf{loan}}&\textbf{Default}&&\cellcolor{blue!25}68.52&50.89&\cellcolor{blue!25}31.1&\cellcolor{blue!25}38.12&0.1292&0.0055&0.2065&2.8621&0.1015&\textbf{185.33}&1 \\
        \cline{2-13}

        &$M_\Im M_\Phi$&duration&51.87&48.29&22.89&30.61&0.2454&0.0223&0.4023&7.0547&0.1994&145.74& 4\\
        \cline{2-13}
        &$M_\Im L_\Phi$&default&68.15&49.24&29.47&36.42&0.1333&\cellcolor{blue!25}0.0022&0.2109&2.915&0.1066&179.91&2 \\
        
        \cline{2-13}
        &$L_\Im M_\Phi$&housing&66.35&\cellcolor{blue!25}51&28.24&35.35&\cellcolor{blue!25}0.1211&0.0057&\cellcolor{blue!25}0.1782&\cellcolor{blue!25}2.7645&\cellcolor{blue!25}0.0945&177.78& 3\\
        
        \hline
        
        \multirow{4}{*}{\textbf{job}}&\textbf{Default}&&\cellcolor{blue!25}68.52&50.89&\cellcolor{blue!25}31.1&\cellcolor{blue!25}38.12&0.1292&0.0037&\cellcolor{blue!25}0.125&\cellcolor{blue!25}1.5745&0.0668&\textbf{186.73}&1 \\
        \cline{2-13}

        &$M_\Im M_\Phi$&duration&51.87&48.29&22.89&30.61&0.2454&0.0068&0.2239&1.9314&0.1275&151.13&4 \\

        \cline{2-13}
        &\multirow{2}{*}{$M_\Im L_\Phi$}&loan&67.74&50.9&30.25&37.37&0.1282&\cellcolor{blue!25}0.0104&0.1277&1.5891&\cellcolor{blue!25}0.0611&184.34& \multirow{2}{*}{2}\\
        &&default&68.15&49.24&29.47&36.42&0.1333&0.0023&0.1285&1.6055&0.0702&181.34& \\
        
        \cline{2-13}
        &$L_\Im M_\Phi$&housing&66.35&\cellcolor{blue!25}51&28.24&35.35&\cellcolor{blue!25}0.1211&0.0057&0.1362&1.7924&0.0698&178.81& 3\\

        \hline
        
        \multirow{4}{*}{\textbf{campaign}}&\textbf{Default}&&\cellcolor{blue!25}68.52&50.89&\cellcolor{blue!25}31.1&\cellcolor{blue!25}38.12&0.1292&0.0801&0.4303&-&\cellcolor{blue!25}0.1486&\textbf{187.84}& 1\\
        \cline{2-13}

        &$M_\Im M_\Phi$&duration&51.87&48.29&22.89&30.61&0.2454&\cellcolor{blue!25}0.0056&0.5025&-&0.2766&152.63& 4\\
        \cline{2-13}

        &\multirow{2}{*}{$M_\Im L_\Phi$}&loan&67.74&50.9&30.25&37.37&0.1282&0.0746&0.4224&-&0.1499&185.48&\multirow{2}{*}{2} \\
        &&default&68.15&49.24&29.47&36.42&0.1333&0.0702&0.419&-&0.1533&182.5& \\

        \cline{2-13}
        &$L_\Im M_\Phi$&housing&66.35&\cellcolor{blue!25}51&28.24&35.35&\cellcolor{blue!25}0.1211&0.0643&\cellcolor{blue!25}0.4051&-&0.1499&180.2& 3\\

        \hline
        
        \multirow{4}{*}{\textbf{balance}}&\textbf{Default}&&\cellcolor{blue!25}68.52&50.89&\cellcolor{blue!25}31.1&\cellcolor{blue!25}38.12&0.1292&\cellcolor{blue!25}0.121&\cellcolor{blue!25}0.2419&0.129&\cellcolor{blue!25}0.0484&\textbf{187.96}& 1\\
        \cline{2-13}

        &$M_\Im M_\Phi$&duration&51.87&48.29&22.89&30.61&0.2454&0.125&0.45&0&0.15&152.69& 4\\
        \cline{2-13}

        &\multirow{2}{*}{$M_\Im L_\Phi$}&loan&67.74&50.9&30.25&37.37&0.1282&\cellcolor{blue!25}0.121&\cellcolor{blue!25}0.2419&0.129&\cellcolor{blue!25}0.0484&185.59&\multirow{2}{*}{2} \\

        &&default&68.15&49.24&29.47&36.42&0.1333&0.125&0.25&\cellcolor{blue!25}0.1&0.05&182.62& \\
        \cline{2-13}
        &$L_\Im M_\Phi$&housing&66.35&\cellcolor{blue!25}51&28.24&35.35&\cellcolor{blue!25}0.1211&\cellcolor{blue!25}0.121&\cellcolor{blue!25}0.2419&0.129&\cellcolor{blue!25}0.0484&180.28& 3\\

        \hline
        \multirow{4}{*}{\textbf{education}}&\textbf{Default}&&\cellcolor{blue!25}68.52&50.89&\cellcolor{blue!25}31.1&\cellcolor{blue!25}38.12&0.1292&\cellcolor{blue!25}0.0192&0.1483&0.6663&0.1001&\textbf{187.57}& 1\\
        \cline{3-13}

        &$M_\Im M_\Phi$&duration&51.87&48.29&22.89&30.61&0.2454&0.0372&0.3199&\cellcolor{blue!25}0.515&0.2068&152.34& 4\\
        
\cline{2-13}
        &\multirow{2}{*}{$M_\Im L_\Phi$}&loan&67.74&50.9&30.25&37.37&0.1282&0.0206&0.1493&0.6613&0.1028&185.2& \multirow{2}{*}{2}\\
        &&default&68.15&49.24&29.47&36.42&0.1333&0.0217&0.148&0.6614&0.1027&182.21& \\

        \cline{3-13}
        &$L_\Im M_\Phi$&housing&66.35&\cellcolor{blue!25}51&28.24&35.35&\cellcolor{blue!25}0.1211&0.0249&\cellcolor{blue!25}0.1337&0.6674&\cellcolor{blue!25}0.098&179.89& 3\\
\bottomrule

\end{tabular}}
\end{table}


\begin{table}[h]
\small 
\caption{
Comparing the model predictive performance and fairness before and after removing the potentially biased features (PBF), when no mitigation techniques are used,  on \textbf{Bank dataset}
}

 \begin{subtable}[h]{\textwidth}
   \small

        \caption{Model performance and fairness when no mitigation techniques are used}
       \label{tab:bank-eval-mitigation-1}
       
       \centering
        \resizebox{0.95\textwidth}{!}{\begin{tabular}{l| c | c c c|cc c}
        

        \multirow{2}{*}{\textbf{Label}}&\multirow{2}{*}{\textbf{PBF}}&\multicolumn{3}{c|}{\textbf{Performance}}&\multicolumn{3}{c}{\textbf{Fairness}}\\
        

        &&\textbf{Win}&\textbf{Tie}&\textbf{Loss}&\textbf{Win}&\textbf{Tie}&\textbf{Loss}\\

        \midrule
        \midrule
        
        $M_\Im M_\Phi$&duration&0&8&0&0&4&4\\

        \hline
        \multirow{2}{*}{$M_\Im L_\Phi$}&loan&0&8&0&0&8&0\\
        &default&0&8&0&0&8&0\\
        
        \hline
        $M_\Im L_\Phi$&housing&0&8&0&0&8&0\\
    
\bottomrule

\end{tabular}}

       \vspace{5pt}
    \end{subtable}

\hfill

    \begin{subtable}[h]{\textwidth}
   \Large

        \caption{Summary comparison of model performance and fairness before and after using the bias mitigation techniques, on the dataset}
       \label{tab:bank-eval-mitigation-2}
       
       \centering
        \resizebox{0.95\textwidth}{!}{\begin{tabular}{l c| c | c c cc|ccc cc |g}
        

        \multirow{2}{*}{\textbf{Feature}}&\multirow{2}{*}{\textbf{Cases}}&\multirow{2}{*}{\textbf{PBF}}&\textbf{ACC}&\textbf{PRE}&\textbf{Recall}&\textbf{F1}&\textbf{F-alarm}&\textbf{AOD}&\textbf{SPD}&\textbf{DIR}&\textbf{FPR$_D$}&\textbf{Total}\\

        &&&\textbf{W/T/L}&\textbf{W/T/L}&\textbf{W/T/L}&\textbf{W/T/L}&\textbf{W/T/L}&\textbf{W/T/L}&\textbf{W/T/L}&\textbf{W/T/L}&\textbf{W/T/L}&\textbf{W/T/L}\\

        \midrule
        \midrule
        
        \multirow{5}{*}{\textbf{FairSMOTE}}&Default&&0/9/0&0/9/0&0/9/0&0/9/0&0/9/0&1/8/0&4/5/0&3/6/0&2/7/0&10/71/0\\
        \cline{2-12}
        &$M_\Im M_\Phi$&duration&0/8/0&0/8/0&0/8/0&0/8/0&0/8/0&1/7/0&7/1/0&3/5/0&4/4/0&15/57/0\\

        \cline{2-12}
        &\multirow{2}{*}{$M_\Im L_\Phi$}&loan&0/8/0&0/8/0&0/8/0&0/8/0&0/8/0&0/8/0&4/4/0&2/6/0&0/8/0&6/66/0\\
        &&default&0/8/0&0/8/0&0/8/0&0/8/0&0/8/0&0/8/0&4/4/0&3/5/0&0/8/0&7/65/0\\
        
        \cline{2-12}
        &$M_\Im L_\Phi$&housing&0/8/0&0/8/0&0/8/0&0/8/0&0/8/0&3/5/0&3/5/0&2/6/0&2/6/0&10/62/0\\
    
\bottomrule

 \multirow{5}{*}{\textbf{LRTDD}}&Default&None&0/9/0&0/9/0&0/9/0&0/9/0&0/9/0&1/8/0&6/3/0&4/5/0&3/6/0&14/67/0\\
        \cline{2-12}
        &$M_\Im M_\Phi$&duration&0/8/0&0/8/0&0/8/0&0/8/0&0/8/0&1/7/0&6/2/0&3/5/0&4/4/0&14/58/0\\

        \cline{2-12}
        &\multirow{2}{*}{$M_\Im L_\Phi$}&loan&0/8/0&0/8/0&0/8/0&0/8/0&0/8/0&2/6/0&6/2/0&3/5/0&4/4/0&15/57/0\\
        &&default&0/8/0&0/8/0&0/8/0&0/8/0&0/8/0&1/7/0&6/2/0&3/5/0&3/5/0&13/59/0\\
        
        \cline{2-12}
        &$M_\Im L_\Phi$&housing&0/8/0&0/8/0&0/8/0&0/8/0&0/8/0&2/6/0&5/3/0&2/6/0&3/5/0&12/60/0\\
    
\bottomrule

\multirow{5}{*}{\textbf{Reweighing}}&Default&None&0/9/0&0/9/0&0/9/0&0/9/0&0/9/0&2/7/0&6/3/0&3/6/0&3/6/0&14/67/0\\
        \cline{2-12}
        &$M_\Im M_\Phi$&duration&0/8/0&0/8/0&0/8/0&0/8/0&0/8/0&1/7/0&4/4/0&3/5/0&3/5/0&11/61/0\\

        \cline{2-12}
        &\multirow{2}{*}{$M_\Im L_\Phi$}&loan&0/8/0&0/8/0&0/8/0&0/8/0&0/8/0&2/6/0&4/4/0&2/6/0&5/3/0&13/59/0\\
        &&default&0/8/0&0/8/0&0/8/0&0/8/0&0/8/0&2/6/0&5/3/0&3/5/0&3/5/0&13/59/0\\
        
        \cline{2-12}
        &$M_\Im L_\Phi$&housing&0/8/0&0/8/0&0/8/0&0/8/0&0/8/0&3/5/0&4/4/0&2/6/0&4/4/0&13/59/0\\
\bottomrule
\end{tabular}}

       \vspace{5pt}
    \end{subtable}

   \end{table}

Finally, in this Section we evaluated the potential biased features identified in the Bank dataset. Table~\ref{tab:bank-eval-before} shows evaluation results on the Bank dataset evaluated, given the different scenarios, as follows: $M_\Im M_\Phi = \{duration\}$ (i.e., the bias-inducing and yet important feature), $M_\Im L_\Phi =\{loan, default\}$ (more bias, yet less important), $L_\Im M_\Phi=\{housing\}$ (less biased, yet more important). Each result in the row reports the model performance and fairness metrics for different scenarios including the default, and after removing the features in column `PBF'. The ranking of each scenario in column `T-score'. 

The results in Table~\ref{tab:bank-eval-before} shows that the model performs better on the original data (row `Default'), and a negligible impact in model performance is observed when the feature `loan' or `default' is removed, i.e., the features categorised as more bias-inducing, yet less important feature ($M_\Im L_\Phi$). Looking at the `T-score' values (in Table~\ref{tab:bank-eval-before}) corresponding to the $M_\Im L_\Phi$ category, the results clearly show a minimal difference in the model performance and fairness measures compared to the rest of the scenarios. The feature `duration', which is categorised as both more important and more bias-inducing ($M_\Im M_\Phi$) has the overall worse performances and fairness scores, ranked as the least in all the cases. In Table~\ref{tab:bank-eval-mitigation-1} we compare the model performance metrics and fairness of the original dataset against the scenarios when the feature is removed, i.e., Default vs. $M_\Im M_\Phi$, Default vs. $M_\Im L_\Phi$, and Default vs. $L_\Im M_\Phi$. The result indicates a larger number of losses on fairness measure when the feature `duration' is removed (i.e., the feature which is more biased, yet also more important), implying the original dataset wins in terms of fairness compared to the bank dataset without feature  `duration'.

Table~\ref{tab:bank-eval-mitigation-2} compares the performance of the model when different mitigation techniques are used on the dataset after removing a given potential biased features, in column `PBF'. We can see that the model fairness and performance improves when the feature `duration' is removed prior to applying the LRTDD mitigation technique. On the other hand, the number of wins is high at $15$ when FairSMOTE is used. The variability in the performance of the model when different mitigation are used can potentially explain the fact that each of the mitigation techniques tries to address specific kinds of bias (e.g., data imbalance, statistical dependencies or simply finding a correlation/  association between features, i.e., sensitive and non-sensitive features). 

\begin{tcolorbox}
Significant improvement in the fairness and predictive performance of the model can be achieved if appropriate bias mitigation techniques are used--- on the potentially bias inducing yet important features are known upfront. Hence, treating the feature importance and bias inducing features as separate entities is essential when building a robust model in terms of fairness and predictive performance.
\end{tcolorbox}

\section{Discussion}\label{sec:discussion}
The Controlled Direct Impact can help us explain the impact of each feature on the model prediction closely similar to SHAP value. However, the conclusion drawn from such analysis should be supported by further analysis. For example, we have shown in this study that concluding that such features are biased by a simple perturbation of given feature values when other variables are kept unchanged can be misleading (such as~\cite{alelyani2021detection,chakraborty2021bias,perera2022search}).  A genuine conclusion should consider the effect of confounding. 

Indeed, if we assume causation in which some of the relevant features are influenced by a potential bias feature, then computing counterfactuals for this biased feature would require altering downstream features~\cite{freedman2005specifying,holland2003causation,pearl2000models,blank2004measuring,barocas2017fairness,sep-causation-probabilistic,suppes1970theory}. Changing the values of the potential bias feature alone will only correspond to a counterfactual if this feature does not have any mediator variables, which is unlikely~\cite{barocas2017fairness}.

Motivated by the plaintiff’s expert report~\cite{arcidiacono_2018,arcidiacono_2018-2} that claimed that race played a significant role in admissions decisions of the Harvard’s machine learning model~\footnote{Plaintiff’s expert report of Peter S. Arcidiacono, Professor of Economics at Duke University.}
\emph{``Consider the case of a male applicant who is an Asian-American, is not disadvantaged, and has other characteristics that result in a 25\% chance. The chance of the applicant was observed to increase to 36\% when only the applicant's race was changed to white— and keeping all his other characteristics the same. On the other hand, the applicant's chance of being admitted increased to 77\% when his race is changed to Hispanic while keeping his other characteristics constant. When the race was changed to African-American, leaving all other features constant increased the chance of his admission to 95\%.''}. 
A logistic regression model was fitted against the historical admissions decisions regarding features considered relevant for Harvard's admission decision to model Harvard's decision rule above. The plaintiff’s report above, however, can be translated into the technical claim that the model used for admissions decisions did not satisfy the conditional statistical parity. Formally, let $X$ be the set of applicant features and $S$ as the applicant’s reported race. If we use $\hat{Y}=\hat{y}$ to denote the  model admissions decision deemed relevant for admission, then we say:
\[
\frac{\rho(\hat{y} |X=x, S\neq s)}{\rho(\hat{y} |X=x,S=s)} \leq \tau
\]
 where $\rho(\hat{y}|X=x,S \neq s)$ denotes the conditional probability (evaluated over $X$) that the class outcome is $\hat{y} \in \hat{Y}$ given protected race group $S \neq s$ and $P(\hat{y}|X=x, S = s)$ is the conditional probability given non-protected group $S:=s$, and $\tau$ is the allowed ratio, such as the $80\%$ rule.
For this condition to be violated depends solely on which feature was considered relevant for the admission, which to a large extent will correspond to the defendant’s expert~\cite{arcidiacono2022legacy}.

In this study, we propose an approach for systematically identifying all bias-inducing features of a model to help support the decision-making of domain experts.  The proposed method considers both the case of using direct and indirect of each features on the model prediction using a novel technique of data swapping. We defined a single feature swapping function as a function that modifies the values of the single feature keeping other features unchanged, and the function's output is used to estimate the direct impact of the feature on the model prediction. On the other hand, we proposed the double features swapping function which switch the values of pairs consisting of the feature and all the mediating variables used to estimate the Total Natural Impact of the feature on the model prediction. 

We demonstrate the usefulness of our approach in the decision-making by the domain experts on how our framework can spearhead the development, testing, maintenance, and deployment of fair machine learning systems. Specifically, we show how our proposed framework can be used to rank all the features based on the level of bias-inducing capability. We showed that when the single feature swapping function is used, the resulting conclusion can be used to explain the feature importance with the result closely similar to the SHAP value, a state-of-the-art method to explain the impact of each feature on the model prediction based on the famous Shapley values from game-theory. Then, we showed how the double feature swapping functions that consider the mediating variables following the concept of probabilistic causal; the resulting outcome can be used to estimate the Total Natural Impact of the features on the model prediction. We demonstrated empirically that treating the concepts of feature importance and bias inducing features as separate is essential in making informed decisions on what constitute a 
most relevant or least relevant features 
as interpreted by the domain experts. The resulting insights will help the domain expert choose the features that improve predictive performance and fairer machine learning models.



\section{Related Works}\label{sec:related-works}
Our work combined the idea of bias assessment and explaining the feature importance. This Section discusses the literature related to the testing for bias and fairness and the model explainability relevant to our works.
\subsection{Bias and Fairness}

The analysis of bias has been approached from different directions, including legal reasoning, sociological theories, statistical methods, and economic models~\cite{custers1866discrimination,romei2014multidisciplinary,sunstein2018legal,lang2020race}. Some previous researchers such as~\cite{hajian2012methodology,hajian2011discrimination,kamiran2012data} have studied how to prevent bias in the data mining process. In this study, instead, we focus on the detection of bias in a dataset involving historical decisions, and the literature presented the most relevant literature .

Perera, Anjana, et al.~\cite{perera2022search} proposed a novel search-based fairness testing (SBFT) approach based on the concept of fairness degree to test for fairness and evaluate the fairness of regression-based ML systems. Their proposed SBFT approach works by computing the maximum difference in the predicted values by the machine learning system for all pairs of identical instances apart from the sensitive features to describe the worst-case behavior of the system. Chakraborty, Joymallya, et al~\cite{chakraborty2021bias} proposed a Fair-SMOTE algorithm to test for the root causes of bias in the prior decisions about the data selection and the labeling assigned to the data. Then, a mitigation technique is proposed to solve the data imbalance. The steps include dividing the data into subgroups based on class and protected features defined by the sensitive features and then generating synthetic data points for all the subgroups except the subset  with the maximum number of data points. The Fair-SMOTE algorithm uses the situation testing technique to test how the labeling can induce bias in the model by flipping the value of sensitive features for every data point and computing the propensity score to estimate the impact on the model prediction. Aggarwal, Aniya, et al.~\cite{aggarwal2019black} uses the combination of symbolic execution and local explainability for automatic generation of test case that help detect individual bias in machine learning models. Given the machine learning model, domain constraints, and protected attribute set, the aim is to generate test cases to maximize the successful test cases defined by the protected attribute leading to a bias behaviour for different combinations of protected attribute values. 
Li, Yanhui, et al.~\cite{LRTDD:2022} propose to analyze the association between non-sensitive features and sensitive features to identify the biased data points of the sensitive features, in the training set. Biased data points are then removed to make the privileged and non-privileged subgroups independent. Moreover, the approach requires applying the same revision of modification on the test set, and can only work with the logistic regression. 

Clearly the above approaches assume the prior knowledge of the sensitive features in the dataset under analysis. The sole idea of the above approach is to analyze and mitigate the bias defined by the protected subgroup of these sensitive features to estimate the direct discrimination.  Similarly other previous studies have mainly relied upon simple statistical analysis involving association or correlation measures~\cite{calders2010three,luong2011k,ruggieri2010data,majumder2021fair,peng2021fairmask}. However, such analyses can lead to incorrect conclusions because they largely ignore the effect of confounding variables;--- variables that can be used to determine both the outcome and the feature pairs. In other words, quantifying bias through such analyses can distort the causal effect belonging to the protected or unprotected features. Next, we present the relevant literature for indirect distribution analysis.

Mancuhan, Koray, and Chris Clifton.~\cite{mancuhan2014combating} propose Bayesian networks as a technique for modeling the probability distribution of a class to identify discrimination. The method includes discovering all the dependencies between attributes and using these dependencies to estimate the joint probability distribution. Ruggieri et al.~\cite{Pedreschi:2009} introduce a reference model, a form of rule inference for classification then analyze those rules to help discover the indirect discrimination of Automatic Decision Support Systems (DSS).  The notion of itemsets, association rules, and classification rules are used to encode the background knowledge about the feature correlation.  The itemset was used to represent the sensitive features e.g., $sex=female$, $age=older$, or $race=black$. The associated rule will combine the classification rule of these itemset $sex=female$, $age=older$, $car=own \to credit=no$, and potentially biased subgroup (i.e., the intersection of itemset consisting of only the protected feature value).

Our work deviates from the above literature in that: 1) the proposed technique is based on the concept of probabilistic causation instead of defining the correlation. 2) Our technique can be used to detect both direct and indirect bias as compared to the above methods that focus on the specific type of bias. 3) The approach applies to both classification and regression models. 4) Our work does not require prior knowledge of the sensitive features; instead, we identify all the features that are potentially inducing bias to the model, and the domain expert makes the decision.

\subsection{Explainability}

Our approach to identify potentially biased features is closely linked to techniques from the field of eXplainable AI (XAI), specifically `Explainability', which aim at characterizing the behavior of complex models in order to increase user trust and/or find bugs in the models. The main connection between our method and XAI is that, like us, they consists in perturbing the model input in some way and records the resulting effects on the output. For instance Individual Conditional Expectations (ICE) \cite{goldstein2015peeking} perturb the inputs
by forcing feature $j$ to take a specific value
$\tau$, scan over possible values of $-\infty<\tau<\infty$,
and visualize the resulting model outputs as
a line chart. Permutation Feature Importance (PFI) \cite{fisher2019all} provide
global importance scores for each input feature by permuting
the $j$th row of $\bm{X}$, get predictions on this new dataset and
records the decrease in performance. Local
methods like LIME \cite{ribeiro2016should} and
SHAP \cite{lundberg2017unified} perturb a
specific instance $X_i$ thousands of times
and estimate the model response on those
noisy inputs with a linear model. The weights
of this linear model are interpreted as the
``local'' importance of the features for the
specific model decision $f(X_i)$.
The main difference with our approach is the
mechanism for perturbing the input,
which is inspired by the counterfactual
approach to causal inference.

It is also important to note that our approach is not the same as the so-called Counterfactual
examples for XAI \cite{wachter2017counterfactual}. Indeed,
counterfactual examples aim at finding the smallest realistic
perturbation that can change the outcome of the model. In a sense, the perturbation is optimized to ensure that change model decision. 
Our approach, on the contrary, perturbs specific
features and records the resulting change in 
distributions as evidence for potential bias.
We do not attempt to find the optimal ways to
perturb each data instance so that the model output changes. When describing our approach, we use ``counterfactual'' as a reference to the
counterfactual approach to causal inference.


\section{Threat to Validity}\label{sec:threat}

\nd \textbf{Reliability Validity:}  concerns the possibility of replicating this study. Every conclusion obtained through empirical studies is threatened by potential bias from data sets. To mitigate this potential bias, We carried out the empirical study to evaluate the proposed approach on four different well-known datasets and a classification model; however, there may be a slight change in the conclusions derived if other datasets and models are used. Similarly, we evaluated our techniques on four difference divergence measures, and the results show high consistency. We will explore more evaluation criteria in the future. 
 
\nd \textbf{Internal Validity:} relates to threats in the study's structure concerning how well a study is conducted, e.g., analysis method, the selection bias. Where previous researchers focused mainly on the measures of the predicted class labels, such as propensity score and the classification measure. Our work, instead, uses distance measures to compute the statistical distance;--- as the statistical distance will capture even a small change that might not have changed the class label. In the future, we would like to explore how the results might change if both statistical distance and measures involving only the predicted class labels. 
Also, concerning the temporal priority ordering, we must note that the conclusion derived from our findings for the double swapping functions is strongly directly impacted by the temporal priority ordering of the cause and effect variables.  The feature rankings' outcome will likely change when different temporal ordering is used. In temporal priority ordering used in this study consist of both manually defined for some variables and positive statistical determined using automated approach. Also, we recommend the manual approach to capture the semantic characteristics of the feature, the choice of temporal priority ordering must be guided by the domain experts and stakeholders involved. 
 
\nd \textbf{External Validity:} is about the possibility to generalize our results. Our work is based on tabular data, which is common in many machine learning systems and discriminative measures. We will consider extending this work to include other data mining domains, such as image processing and text mining. 
Also, to maximize the stability of rankings what constitutes relevant features of the ML models, we removed the highly correlated features prior to the evaluation experiments of our approach. This is true because some specific model like logistic regression (used in this study) becomes unstable in the presence of correlation in the dataset. Such models, therefore, tend to assign similar weights to the highly correlated feature; and interns the weights of the features belonging to groups of correlated features decrease as the group sizes increase, leading to incorrect model interpretation and misleading feature ranking. However, this also implies that our feature ranking of bias-inducing and feature importance might change if the correlated features were included in the studied dataset. Despite this preprocessing step, it is worth pointing out that the similarly sets of features of the respective dataset has been used in the previous studies~\cite{chakraborty2021bias,zhang2020white,chakraborty2020fairway,LRTDD:2022,perera2022search} on bias assessment and mitigation in machine learning. Moreover, in the future we plan to perform a thorough empirical analysis to understand how our process behaves given different types of models and the presence of the correlation. 

For our evaluation of the proposed framework, we consider only a case where a single bias-inducing feature is removed at a given time, not considering the combination of features. We will consider evaluating our technique on a combination of bias-inducing features. Also, our current evaluation using the biased mitigation techniques does not focus on what techniques work better, instead, we only wanted to show how the model behaves when we apply the mitigation techniques after removing the different features. Our future work will address this concern by breaking down the type of biased and subsequently proposing the mitigation technique that matches the causes. For instance, by  addressing the bias mediating variables that are also important variables.

\section{Conclusion}\label{sec:conclusion}

In this study, we proposed an approach for systematical identification bias-inducing features of the machine learning model based on a feature swapping technique. Two different kinds of swapping functions are proposed. First, a single feature swapping function that modifies the values of the single feature keeping other features unchanged, and helps estimate the direct impact of the feature on the model prediction. Second, the double features swapping functions which switch the values of pairs consisting of the feature and all the mediating variables  and the resulting outputs are used to estimate the total natural impact of the feature on the model prediction. Four different distance measures (i.e., Hellinger distance, Jensen-Shannon divergence, total variation distance, and Wasserstein distance) are used to evaluate the impact of swapping the features using our proposed functions on the model prediction. 
 
We demonstrate by answering two main research questions how the domain experts can use our proposed approach to identify the potentially bias-inducing features. We showed with the help of the state-of-the-art model explainability SHAP tool that the potential bias-inducing features that are less important to the model can be removed (or a better mitigation technique is required) to improve the fairness of the machine learning model. Our study is the first step in helping domain experts make an informed decision by following a systematic identification of bias-inducing features. Notably, treating the feature importance and bias-inducing features as separate entities is essential to help  build a robust model that is fairer yet with better predictive performance. The domain experts can use our approach to visualize the cause of bias in the model, systematic features selection, prioritizing the fixes,  and therefore helping contribute to the standard procedure when developing, maintaining, and deploying fairer machine learning systems.


There are several exciting directions for future work. First, we would like to extend our framework to include the technique for mitigating or removing the bias induced to the model as a result of the identified bias-inducing features. Second, our framework can naturally be extended to support non-probabilistic classification models using the propensity score instead of the divergence measures of the model prediction. Thirdly, when applying the double feature swapping function, it will be interesting to consider the spatial relationship to eliminate any relationship that could be captured as counterfactual.  We also aim to evaluate the performance of our methods on the large models that also include non-tabular dataset. This study will help spearhead the standard procedure when automating the development, testing, deployment~\cite{openja2022studying,openja2022empirical}, and maintaining fairer machine learning workflow~\cite{majidi2022empirical}.

\begin{acknowledgements}
This work is supported by the DEEL Project CRDPJ 537462-18 funded by the National Science and Engineering Research Council of Canada (NSERC) and the Consortium for Research and Innovation in Aerospace in Québec (CRIAQ), together with its industrial partners Thales Canada inc, Bell Textron Canada Limited, CAE inc and Bombardier inc.\footnote{\url{https://deel.quebec}}
\end{acknowledgements}

\section*{Conflict of Interest}
The authors declared that they have no conflict of interest

\section*{Data Availability Statements (DAS)}
We made available our framework and all dataset used in this study, including the intermediate experimental results, in a public repository at~\footnote{\url{https://github.com/openjamoses/Bias-detection-dataswap}}


%
%


\bibliographystyle{spmpsci} 
\bibliography{main}   

\end{document}